\documentclass{article}
\pdfoutput=1

     \usepackage[final]{neurips_2023}

\usepackage[utf8]{inputenc} 
\usepackage[T1]{fontenc}    
\usepackage{hyperref}       
\usepackage{url}            
\usepackage{booktabs}       
\usepackage{amsfonts}       
\usepackage{nicefrac}       
\usepackage{microtype}      
\usepackage{xcolor}         

\usepackage[normalem]{ulem}
\useunder{\uline}{\ul}{}

\usepackage{amsmath}
\usepackage{multirow}
\usepackage{subcaption}
\usepackage{graphicx}
\usepackage{float}
\usepackage{tabularx}
\usepackage{wrapfig,lipsum,booktabs}
\usepackage{wrapfig}
\restylefloat{figure}

\usepackage{amsthm}

\usepackage{booktabs,etoolbox}

\usepackage{algorithm2e} 
\SetKwComment{Comment}{/* }{ */}
\SetKwInput{KwData}{Input}
\SetKwInput{KwResult}{Output}

\DeclareMathOperator*{\argmax}{argmax} 
\DeclareMathOperator*{\argmin}{argmin} 

\usepackage{listings}
\usepackage{colortbl}

\definecolor{codegreen}{rgb}{0,0.6,0}
\definecolor{codegray}{rgb}{0.5,0.5,0.5}
\definecolor{codepurple}{rgb}{0.58,0,0.82}
\definecolor{backcolour}{rgb}{0.95,0.95,0.92}
\definecolor{red}{RGB}{233,25,25}
\lstdefinestyle{mystyle}{
    backgroundcolor=\color{backcolour},   
    commentstyle=\color{codegreen},
    keywordstyle=\color{magenta},
    numberstyle=\tiny\color{codegray},
    stringstyle=\color{codepurple},
    basicstyle=\ttfamily\tiny,
    breakatwhitespace=false,         
    breaklines=true,                 
    captionpos=b,                    
    keepspaces=true,                 
    showspaces=false,                
    showstringspaces=false,
    showtabs=false,                  
    tabsize=2
}

\title{\textsc{Emma-X}: An EM-like Multilingual Pre-training Algorithm for Cross-lingual Representation Learning}

\author{%
  \hspace{-1mm}Ping Guo\textsuperscript{\ddag,\S}\enskip\enskip\enskip\enskip\enskip\enskip
  Xiangpeng Wei\textsuperscript{\dag,}\thanks{Corresponding Author.}\enskip\enskip\enskip\enskip\enskip\enskip
  Yue Hu\textsuperscript{\ddag,\S,*} \\
  \vspace{2mm}\textbf{Baosong Yang\textsuperscript{\dag}\enskip\enskip\enskip\enskip\enskip\enskip
  Dayiheng Liu\textsuperscript{\dag}\enskip\enskip\enskip\enskip\enskip\enskip
  Fei Huang\textsuperscript{\dag}\enskip\enskip\enskip\enskip\enskip\enskip
  Jun Xie\textsuperscript{\dag}} \\
	\textsuperscript{\ddag}Institute of Information Engineering, Chinese Academy of Sciences, Beijing, China\\
	\textsuperscript{\S}School of Cyber Security, University of Chinese Academy of Sciences, Beijing, China\\
   \textsuperscript{\dag}Machine Intelligence Technology Lab, Alibaba DAMO Academy, Hangzhou, China\\
	\hspace{-3mm}\texttt{\small{\{guoping, huyue\}@iie.ac.cn, pemywei@gmail.com}}\,\
}

\begin{document}

\maketitle

\begin{abstract}
 
Expressing universal semantics common to all languages is helpful in understanding the meanings of complex and culture-specific sentences. The research theme underlying this scenario focuses on learning universal representations across languages with the usage of massive parallel corpora. However, due to the sparsity and scarcity of parallel data, there is still a big challenge in learning authentic ``universals'' for any two languages. In this paper, we propose \textsc{Emma-X}: an \textbf{{EM}}-like \textbf{{M}}ultilingual pre-training \textbf{{A}}lgorithm, to learn \textbf{(X)C}ross-lingual universals with the aid of excessive multilingual non-parallel data. \textsc{Emma}-X unifies the cross-lingual representation learning task and an extra semantic relation prediction task within an EM framework. Both the extra semantic classifier and the cross-lingual sentence encoder approximate the semantic relation of two sentences, and supervise each other until convergence. To evaluate \textsc{Emma}-X, we conduct experiments on \textbf{\textsc{xrete}}, a newly introduced benchmark containing 12 widely studied cross-lingual tasks that fully depend on sentence-level representations. Results reveal that \textsc{Emma}-X achieves state-of-the-art performance. Further geometric analysis of the built representation space with three requirements demonstrates the superiority of \textsc{Emma}-X over advanced models~\footnote{Codes and datasets of the \textsc{\textsc{xrete}} benchmark: \url{https://github.com/guopingiie/EMMA-X}}. 
\end{abstract}

\section{Introduction}



Research on how to express universal semantics for natural languages (metaphorically as ``alphabet of human thoughts'' by \citet{leibniz1996leibniz}) has lasted a long time. Usually, these universal meanings underlying all human natural languages are referred to as irreducible semantic cores \citep{wierzbicka1999emotions}. These common cores across languages can serve as a bridge, to help better understand the exact meanings of complex sentences in different languages.

\vspace{-2pt}


In the context of computational linguistics, various works~\citep{huang-etal-2019-unicoder,conneau-etal-2020-unsupervised,chi-etal-2021-infoxlm,wei2021on,lee-etal-2022-toward,li2023dualalignment,chen2023alleviating} have led to great improvements on learning cross-lingual universal representations with the usage of parallel corpora, and verify that multilingual universality contributes a major performance on cross-lingual understanding. However, due to the sparsity and scarcity of parallel data, these advanced techniques face a big challenge in learning real universality among all languages. For instance, among the widely-available top 100 languages that theoretically can build 4950 language pairs, only about 200 language pairs have considerable parallel data~\citep{aharoni-etal-2019-massively,51503}. Recently, Large Language Models (\textsc{Llm}s) (e.g., PaLM~\citep{chowdhery2022palm}, OPT~\citep{zhang2022opt}, BLOOMZ~\citep{workshop2023bloom}, ChatGPT, etc.) have reached a milestone in the field of Natural Language Processing, for their promising capability at understanding and following complex natural language instructions in different languages. 
By modeling a wide variety of sentence samples in discrete sentence space, \textsc{Llm}s can capture some universal linguistic phenomena to gain cross-lingual transferability. This is consistent with our goal of building a universal basement that supports all languages. The difference lies in that we achieve it through learning universal continuous representations across different languages.


\vspace{-2pt}
Concretely, we propose \textsc{Emma-X} to tackle the above challenge from a continuous perspective. \textsc{Emma}-X can learn cross-lingual universal sentence representations with excessive non-parallel multilingual data by unifying two highly dependent tasks in an EM \citep{543975} framework: semantic relation classification and cross-lingual universal representation learning. For the former, we introduce a Gaussian Mixture Model~\citep{everitt1981finite} classifier (\textbf{GMM classifier}) to deal with the key challenge of forming positive sentence pairs for non-parallel multilingual corpora, by annotating the semantic relationship of sentence pairs in any two arbitrary languages on the fly. For the latter, we employ \textbf{a cross-lingual encoder} to learn universal sentence representations via contrastive learning, where positive pairs are chosen by GMM classifier. Further, we construct training signals according to the output of the cross-lingual encoder, to inversely supervise GMM classifier. From the perspective of EM algorithm, in E-step, both modules try to approximate the semantic relationship given a sentence pair sampled from two arbitrary languages. One module is supervised by the approximation of the other to build its own expectation. In M-step, two modules update their parameters by maximizing expectations, respectively. We give a theoretical justification about how these two tasks can be interpreted from an EM perspective (Section~\ref{sec:theoretical}).


\vspace{-2pt}
To incentivize the research of universal sentence representation learning, we form a Cross-lingual REpresentation Transfer Evaluation (\textbf{\textsc{xrete}}) benchmark, which includes 12 cross-lingual tasks covering more than 50 languages. \textsc{xrete} fully depends on sentence-level representations. Experimental results demonstrate that \textsc{Emma}-X significantly outperforms pre-trained language models~\citep{conneau-etal-2020-unsupervised,chi-etal-2021-infoxlm} by 32\% at most on \textsc{xrete}. We also perform an evaluation of ChatGPT on \textsc{xrete} to explore its multilingual performance. Detailed analysis also shows that \textsc{Emma}-X can mitigate the representation discrepancy between head and massive long-tail languages. We further conduct geometric analysis directly on representation space from three perspectives: \underline{Invariance} \citep{abend-rappoport-2017-state}, \underline{Canonical Form} \citep{10.1162/089120100750105975} and \underline{Isotropy} \citep{mu2018allbutthetop}, which provides a further understanding of the cross-lingual transferability of these models.

\section{Preliminaries}\label{sec:preliminary}

Cross-lingual representation learning aims at mapping sentences from different languages into a unified continuous space, where synonyms across different languages are pulled closer. Given a sentence $\mathbf{x}$, the representation is formulated as
\begin{equation}
\footnotesize
    \boldsymbol{\gamma^{(\mathbf{x})}} = f \big[ g \big( \mathcal{M}(\mathbf{x};\Theta_{\mathcal{M}}) \big) \big],
\end{equation}
where $\mathcal{M}(\cdot;\Theta_{\mathcal{M}})$ denotes the encoder network with a set of trainable parameters $\Theta_{\mathcal{M}}$, which is typically implemented as a transformer encoder architecture \citep{Vaswani2017Attention,devlin-etal-2019-bert,lee-etal-2022-toward,feng-etal-2022-language}. $f(\cdot)$ is L-2 normalization and $g(\cdot)$ is the aggregate function. We take the final hidden states of ``\texttt{[CLS]}'' token as the aggregate sentence representation.

To learn reasonable representations that can express universal semantics across different languages, various well-designed techniques have been applied to $\gamma^{(\mathbf{x})}$. A predominant one is to build contrastive learning (CTL) \citep{saunshi2019theoretical} objective with parallel corpora. The basic idea is to maximize the similarity between representations (i.e., $\boldsymbol{ \gamma^{(\mathbf{x})}}$ and $\boldsymbol{\gamma^{(\mathbf{y})}}$) of two semantically-equivalent sentences $(\mathbf{x},\mathbf{y})$, while keep randomly sampled irrelevant ones $\boldsymbol{\gamma^{(\mathbf{y}^{\prime})}}$ away. Formally, assume $\mathcal{B}$ to be a batch of multilingual parallel bitexts, the contrastive loss under InfoNCE~\citep{Oord2018ctl} formulation is
\begin{equation}
\begin{split}
\footnotesize
    \mathcal{L}_{\mathbf{CTL}}=-\log \frac{e^{{s(\gamma^{(\mathbf{x})}, \gamma^{(\mathbf{y})})}}}{e^{s(\gamma^{(\mathbf{x})}, \gamma^{(\mathbf{y})})} + \sum_{\mathbf{y}^{\prime} \in \mathcal{B}, \mathbf{y}^{\prime} \neq \mathbf{y}} e^{s(\gamma^{(\mathbf{x})}, \gamma^{(\mathbf{y}^{\prime})})}},
\end{split}
\label{eq:ctl}
\end{equation}
where $s(\cdot)$ is implemented as the cosine similarity $ s(\gamma^{(\mathbf{x})},\gamma^{(\mathbf{y})}) = \frac{\gamma^{(\mathbf{x})\top} \gamma^{(\mathbf{y})}}{\Vert \gamma^{(\mathbf{x})} \Vert \cdot \Vert \gamma^{(\mathbf{y})} \Vert}$, $\mathbf{y}$ and $\mathbf{y}^{\prime}$ are typically called positive and negative samples.


\begin{figure*}
    \centering
    \includegraphics[scale=0.65]{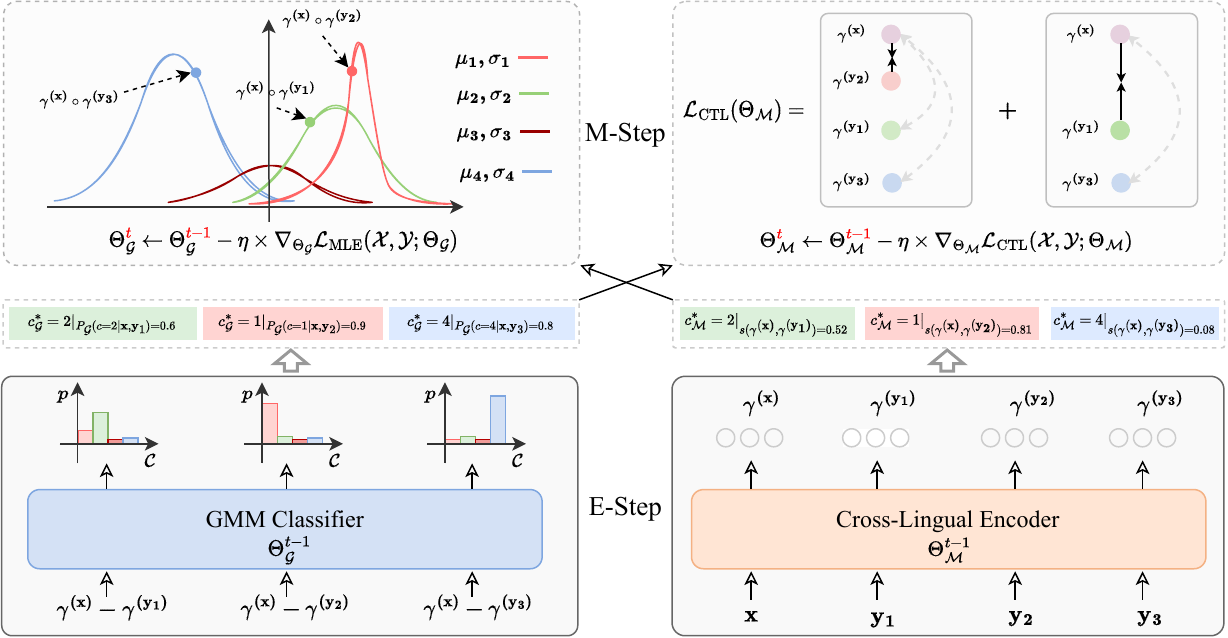}
\caption{Illustration of \textsc{Emma-X}, involving two modules (GMM classifier and Cross-lingual Encoder) that are reciprocated from each other and are updated alternatively. $\mathbf{x}$ means the current instance, $\{{\mathbf{y}_1}$, $\mathbf{y}_2$, $\mathbf{y}_3, ...\}$ are samples in various languages for comparison. $\boldsymbol{\gamma}^{(\mathbf{x})}$ is the continuous representation given a discrete sentence $\mathbf{x}$. $c_{\mathcal{M}}^*$ and $c_{\mathcal{G}}^*$ formulate the semantic ranks approximated according to Eq.~\ref{eq:ctl_pred} and Eq.~\ref{eq:gmm_pred}, to supervise the GMM classifier and cross-lingual encoder, respectively.}
\label{fig:emma-framework}
\end{figure*}

\section{Methodology}\label{sec:method}

We propose {\textsc{Emma}-X} that fully exploits massive monolingual data to learn cross-lingual universal representations. As illustrated in Figure \ref{fig:emma-framework}, \textsc{Emma}-X consists of two modules: 1) A GMM classifier ${\mathcal{G}(\cdot;\Theta_{\mathcal{G}})}$ to approximate the semantic relation of non-parallel sentences. 2) A cross-lingual encoder ${\mathcal{M}(\cdot;\Theta_{\mathcal{M}})}$ to convert multilingual sentences into universal representations. For optimization, \textsc{Emma-X} unifies these two modules in an EM framework with dual supervision. In this section, we begin with a definition of the semantic relation rank (\S \ref{sec:semantic-rank}). Then, we introduce model initialization (\S \ref{sec:model-init}) and the proposed training paradigm (\S \ref{sec:overall_model}), followed by a dual supervision strategy (\S \ref{sec:dual-supervision}). For a clearer presentation, an Algorithm of \textsc{Emma}-X is shown in Algorithm~\ref{alg:emma}.

\subsection{Rank of Semantic Relations} \label{sec:semantic-rank}

Mainstream methods model semantic relations with a strict binary separation: positives and negatives. However, the boundary between positives and negatives is blurry, and many samples cannot be clearly classified as either positives or negatives. So it cannot maximize the potential of models to perceive more subtle semantic changes. Also, a binary separation will lead to far more negative samples than positive ones (imbalanced data). To more accurately capture the semantic relation between two sentences and alleviate imbalanced problem, we subdivide the relation into ${N}$ semantic ranks, where the semantic similarity of each rank decreases as ${N}$ increases, e.g., $c=1$ denotes two sentences are paraphrases of each other, while $c=N$ implies sentences are irrelevant. In practice, we set $N$ to 4.

\subsection{Model Initialization}
\label{sec:model-init}
In \textsc{Emma}-X, the GMM classifier ${\mathcal{G}(\cdot;\Theta_{\mathcal{G}})}$ and cross-lingual encoder ${\mathcal{M}(\cdot;\Theta_{\mathcal{M}})}$ are initialized by training with massive parallel corpora, respectively.

\noindent\textbf{Initialization of Cross-lingual Encoder.} It is initialized with \textsc{Xlm-r}~\citep{conneau-etal-2020-unsupervised} and then continuously trained with InfoNCE \citep{Oord2018ctl} loss by Eq.\ref{eq:ctl}. Following \textsc{Hictl} \citep{wei2021on} and \textsc{InfoXLM} \citep{chi-etal-2021-infoxlm}, we treat the parallel sentence pairs as the query sentence $\mathbf{x}$ and its positive counterpart $\mathbf{y}$, while treating the randomly sampled sentence as a negative one $\mathbf{y}^{\prime}$. 

\noindent\textbf{Initialization of GMM Classifier.} A reasonable solution to warm up the GMM classifier is to use the available cross-lingual parallel corpora as training signals. Suppose $\mathbf{x}$ and $\mathbf{y}$ are parallel sentences, and ${\mathbf{y}^{\prime}}$ is an outlier. We set the semantic ranks for $(\gamma^{(\mathbf{x})}, \gamma^{(\mathbf{y})})$ and $(\gamma^{(\mathbf{x})}, \gamma^{(\mathbf{y}^{\prime})})$ as $c=1$ and $c=N$, respectively, according to the definition described in \S~\ref{sec:semantic-rank}. To obtain the fine-grained semantic ranks, we design a linear interpolation strategy similar to \citet{wei-etal-2022-learning} and mixup \citep{zhang2018mixup}, which provides virtual training examples for each semantic rank. Formally,
\begin{equation}
\footnotesize
\begin{split}
        \boldsymbol{\gamma^{(\mathbf{\tilde{y}})}} &= (1 - \lambda) \cdot \boldsymbol{\gamma^{(\mathbf{y})}} + \lambda \cdot \boldsymbol{\gamma^{(\mathbf{y}^{\prime})}},
\end{split}
\label{eq:interp}
\end{equation}
where $\lambda \in [0,1]$ is sampled from a uniform distribution. We compute $r= \lceil (1 - \lambda) \cdot (c=1) + \lambda \cdot (c=N) \rceil$ as the soft semantic rank for $(\gamma^{(\mathbf{x})}, \gamma^{(\mathbf{\tilde{y}})})$, where $\lceil \cdot \rceil$ means the least integer greater than or equal to the input. The virtual training examples are grouped together with the real parallel corpora to optimize the GMM classifier:
\begin{equation}
\footnotesize
    \mathcal{L}_{\mathbf{MLE}} = -\log P_{\mathcal{G}}(c=1|\mathbf{x}, \mathbf{y};\Theta_{\mathcal{G}}) 
    - \log P_{\mathcal{G}}(c=N|\mathbf{x}, \mathbf{y}^{\prime};\Theta_{\mathcal{G}}) 
    - \sum_{\tilde{\mathbf{y}}} \log P_{\mathcal{G}}(c=r|\mathbf{x}, \tilde{\mathbf{y}};\Theta_{\mathcal{G}}).
\label{eq:gmm-mle}
\end{equation}
The posterior probability $P_{\mathcal{G}}(\cdot)$ is formulated as
\begin{equation}
\footnotesize
P_{\mathcal{G}}(c=r|\mathbf{x}, \mathbf{y};\Theta_{\mathcal{G}}) = \frac{\pi_r \cdot \mathcal{N}_r\big(\gamma^{(\mathbf{x})} - \gamma^{(\mathbf{y})}|\mu_r, \sigma_r\big)}{\sum_{j=1}^{N} \pi_j \cdot \mathcal{N}_j\big(\gamma^{(\mathbf{x})} - \gamma^{(\mathbf{y})}|\mu_j, \sigma_j\big)},
\label{eq:gmm-posteriori}
\end{equation}
where the Gaussian form $\mathcal{N}_r$ is assigned with a prior probability $\pi_r$, mean $\mu_r$ and standard deviation $\sigma_r$ that are all parameterized by trainable variables, thus ${\Theta_{\mathcal{G}}=\{(\pi_r, \mu_r, \sigma_r) \mid r \in [1,N]\}}$. See \textbf{Appendix \ref{app:detail_eq}} for the specific calculation of Gaussian form $\mathcal{N}_r$.

\subsection{The EM Iteration Framework}
\label{sec:overall_model}
After initialization, \textsc{Emma}-X further trains the GMM classifier ${\mathcal{G}(\cdot;\Theta_{\mathcal{G}})}$ and cross-lingual encoder ${\mathcal{M}(\cdot;\Theta_{\mathcal{M}})}$ with only multilingual non-parallel data with an EM framework.

\noindent\textbf{E-Step.}
For optimization, we represent a training batch of multilingual non-parallel sentences as $\mathcal{X}=\{\mathbf{x}_{1},\mathbf{x}_{2},...,\mathbf{x}_{I}\}$ accompanied by a queue of random sentences as ${\mathcal{Y}=\{\mathbf{y}_{1},\mathbf{y}_{2},...,\mathbf{y}_{K}\}}$ for instance comparison. Formally, the expectation for GMM classifier is:
\begin{equation}
\footnotesize
\begin{split}
\mathcal{L}_{\mathbf{MLE}}(\mathcal{X},\mathcal{Y};\Theta_{\mathcal{G}}) = -\mathbb{E}_{\mathbf{x}_{i} \sim \mathcal{X}} \, \mathbb{E}_{\mathbf{y}_{k} \sim \mathcal{Y}}
\big [ \log P_{\mathcal{G}}(c=c_{\mathcal{M}}^*|\mathbf{x}_{i}, \mathbf{y}_{k};\Theta_{\mathcal{G}}) \big ],
\label{eq:gmm-expectation}
\end{split}
\end{equation}
where $c_{\mathcal{M}}^* \in [1,N]$ represents an approximated semantic rank for the combination of anchor ${\mathbf{x}_i}$ and another random sentence ${\mathbf{y}_k}$, based on the cosine similarity among representations (i.e., $\boldsymbol{\gamma^{(\mathbf{x}_i)}}$ and $\boldsymbol{\gamma^{(\mathbf{y}_k)}}$) produced by the cross-lingual encoder (i.e., ${\mathcal{M}(\cdot;\Theta_{\mathcal{M}})}$). Please refer to \S \ref{sec:dual-supervision} for details. 

Correspondingly, the expectation for the cross-lingual encoder can be calculated with contrastive learning, where the positive samples are established by the maximum a posteriori approximation (argmax prediction) $c_{\mathcal{G}}^*$ given by the GMM classifier. Specifically, we apply ranking InfoNCE~\citep{hoffmann2022ranking} as the training objective, which recursively takes parallel sentence pairs in each rank (e.g, $c_{\mathcal{G}}^*$) as positives and ranks that are larger than $c_{\mathcal{G}}^*$ as negatives. Formally,
\begin{equation}
\footnotesize
\begin{split}
\mathcal{L}_{\mathbf{CTL}}(\mathcal{X},\mathcal{Y};\Theta_{\mathcal{M}}) =
 - \mathbb{E}_{\mathbf{x}_{i} \sim \mathcal{X}} \Bigg [ &\log \frac{ \sum_{\mathbf{y}_{k} \sim \mathcal{Y}_{c_{\mathcal{G}}^*=1}} e^{s[\gamma^{(\mathbf{x}_{i})}, \gamma^{(\mathbf{y}_{k})}]}}{\sum_{\mathbf{y}_{t} \sim \mathcal{Y}_{c_{\mathcal{G}}^* \in [1, N]}} e^{s[\gamma^{(\mathbf{x}_{i})}, \gamma^{(\mathbf{y}_{t})}]} } + \log \frac{ \sum_{\mathbf{y}_{k} \sim \mathcal{Y}_{c_{\mathcal{G}}^*=2}} e^{s[\gamma^{(\mathbf{x}_{i})}, \gamma^{(\mathbf{y}_{k})}]}}{\sum_{\mathbf{y}_{t} \sim \mathcal{Y}_{c_{\mathcal{G}}^* \in [2, N]}} e^{s[\gamma^{(\mathbf{x}_{i})}, \gamma^{(\mathbf{y}_{t})}]} } \\
 &+...+\log \frac{ \sum_{\mathbf{y}_{k} \sim \mathcal{Y}_{c_{\mathcal{G}}^*=N-1}} e^{s[\gamma^{(\mathbf{x}_{i})}, \gamma^{(\mathbf{y}_{k})}]}}{\sum_{\mathbf{y}_{t} \sim \mathcal{Y}_{c_{\mathcal{G}}^* \in [N-1, N]}} e^{s[\gamma^{(\mathbf{x}_{i})}, \gamma^{(\mathbf{y}_{t})}]} }
 \Bigg ],
\label{eq:ctl-expectation}
\end{split}
\end{equation}
where $c_{\mathcal{G}}^* \in [1,N]$ represents a semantic rank approximated by the posteriori of GMM classifier (\S \ref{sec:dual-supervision}). For simplicity, we omit the temperature term in Eq. \ref{eq:ctl-expectation}, and please see \textbf{Appendix \ref{app:detail_eq}} for details.

\noindent\textbf{M-Step.} We use gradient descent algorithm to update the parameters of each module by optimizing its expectation. At each time step $t$, where $\eta$ and $\eta^{\prime}$ are learning rates for each expectation, formally, 
\begin{equation}
\footnotesize
\begin{split}
    \Theta_{\mathcal{G}}^{t+1} &\gets \Theta_{\mathcal{G}}^{t} - \eta \times \nabla_{\Theta_{\mathcal{G}}}  \mathcal{L}_{\mathbf{MLE}}(\mathcal{X},\mathcal{Y};\Theta_{\mathcal{G}}), \\
    \Theta_{\mathcal{M}}^{t+1} &\gets \Theta_{\mathcal{M}}^{t} - \eta^{\prime} \times \nabla_{\Theta_{\mathcal{M}}} \mathcal{L}_{\mathbf{CTL}}(\mathcal{X},\mathcal{Y};\Theta_{\mathcal{M}}).
\end{split}
\label{eq:m-step}
\end{equation}

\subsection{Dual Supervision}
\label{sec:dual-supervision}
The approximated semantic ranks $c_{\mathcal{G}}^*$ and $c_{\mathcal{M}}^*$ are critical in \textsc{Emma}-X training algorithm. To preserve their quality, we propose \texttt{dual supervision}: predictions from one module are fed to the other to calculate the expectation. In this section, we explain in detail how we approximate the semantic ranks for GMM classifier and cross-lingual encoder, respectively. 

\noindent\textbf{Approximate Semantic Rank with GMM classifier.} The way to obtain semantic rank with semantic classifier is straightforward. The semantic rank corresponding to the highest probability among multiple Gaussian distributions is chosen as the prediction, which is further used to supervise the cross-lingual encoder ${\mathcal{M}(\cdot;\Theta_{\mathcal{M}})}$, as illustrated in Eq.~\ref{eq:ctl-expectation}. Formally, 
\begin{equation}
\footnotesize
    c_{\mathcal{G}}^* =\argmax_{r} P_{\mathcal{G}}(c=r|\mathbf{x}_i, \mathbf{y}_k;\Theta_{\mathcal{G}}).
\label{eq:gmm_pred}
\end{equation}
\noindent\textbf{Approximate Semantic Rank with Cross-lingual Encoder.}
One common way to calculate sentence relation is to measure the similarity between two real-valued representations. Assuming $s_r$ (a scalar initialized as $\frac{r}{N}$) can reflect the general similarity score in semantic rank $c=r$. Given a random sentence pair $({\mathbf{x}_i}, \mathbf{y}_k)$, if its similarity score is close to $s_r$, the sentence pair is likely to belong to rank $c=r$. Cross-lingual encoder ${\mathcal{M}(\cdot;\Theta_{\mathcal{M}})}$ determines the semantic relation for each pair according to
\begin{equation}
\footnotesize
\label{eq:ctl_pred}
    c_{\mathcal{M}}^* =\argmin_{r} |s(\gamma^{(\mathbf{x}_i)}, \gamma^{(\mathbf{y}_k)}) - s_{r}|,
\end{equation}
where $|\cdot|$ refers to absolute value. Symmetrically, $c_{\mathcal{M}}^*$ is used to supervise GMM classifier (Eq.~\ref{eq:gmm-expectation}). 

During the training process, the general similarity score for each semantic rank may vary. Thus, we propose a moving-average strategy to adaptively adjust the value of $s_r$ to simulate this change. Specifically, at time step $t$, $s_r$ is updated by cosine similarity of all the sentence pairs, which are currently categorized into the rank $c=r$ according to the cross-lingual encoder in Eq.~\ref{eq:gmm_pred}. 
\begin{equation}
\footnotesize
    s_r^t \gets \epsilon \cdot s_r^{t-1} + (1 - \epsilon) \cdot s(\gamma^{(\mathbf{x}_i)}, \gamma^{(\mathbf{y}_k)}), \qquad  \  \text{if}\ \ \mathbf{y}_k \in \mathcal{Y}_{c_{\mathcal{G}}^*=r}.
\end{equation}
Here $\epsilon \in [0,1]$ is a momentum coefficient to make $s_n$ evolve smoothly during training.

\RestyleAlgo{ruled}
\begin{algorithm}[t]
\footnotesize
\KwData{multilingual parallel and non-parallel corpora}
\KwResult{$\mathcal{M}(\cdot;\Theta_{\mathcal{M}})$ and $\mathcal{G}(\cdot;\Theta_{\mathcal{G}})$}
\textcolor{blue}{Phase 1 \Comment*[r]{Warm-up two modules with multilingual parallel corpora}}
\While{\textit{not convergence}}{
    Sample a batch of multilingual bitexts $\mathcal{B}$\;
    \For{$(\mathbf{x},\mathbf{y}) \in \mathcal{B}$}{
        Sample $\mathbf{y}^{\prime} \in \mathcal{B}, \mathbf{y}^{\prime} \ne \mathbf{y}$, Build virtual training examples $\tilde{\mathbf{y}}$ with Eq.~\ref{eq:interp}\;
        Compute $\mathcal{L}_{\mathbf{CTL}}$ with Eq.~\ref{eq:ctl} to update $\Theta_{\mathcal{M}}$ and $\mathcal{L}_{\mathbf{MLE}}$ with Eq.~\ref{eq:gmm-mle} to update $\Theta_{\mathcal{G}}$\;
    }
}
\textcolor{blue}{Phase 2 \Comment*[r]{\textsc{Emma-X} training with EM framework using only non-parallel corpora}}
\While{\textit{not convergence}}{
  \SetKwBlock{Repeat}{E-Step}{}
  \Repeat{
    Sample a batch of multilingual anchors ${\mathcal{X}}$,
    Queue a batch of random sentences ${\mathcal{Y}}$\;
    \For{$(\mathbf{x}_i,\mathbf{y}_k) \in \mathcal{X} \times \mathcal{Y}$}{
      Approximate semantic rank $c_{\mathcal{G}}^*$ with Eq.~\ref{eq:gmm_pred} and $c_{\mathcal{M}}^*$ with Eq.~\ref{eq:ctl_pred} \;
    }
    Compute $\mathcal{L}_{\mathbf{MLE}}(\mathcal{X},\mathcal{Y};\Theta_{\mathcal{G}})$ according to Eq.~\ref{eq:gmm-expectation} and $\mathcal{L}_{\mathbf{CTL}}(\mathcal{X},\mathcal{Y};\Theta_{\mathcal{M}})$ according to Eq.~\ref{eq:ctl-expectation}\;
  }
  \SetKwBlock{Repeat}{M-Step}{}
  \Repeat{
    Update $\Theta_{\mathcal{M}}$ and $\Theta_{\mathcal{G}}$ according to Eq.~\ref{eq:m-step}\;
  }
}
\caption{\footnotesize \textsc{Emma-X} Training Algorithm}\label{alg:emma}
\end{algorithm}

\section{Theoretical Analysis}\label{sec:theoretical}
In this section, we provide theoretical justification for \textsc{Emma}-X and demonstrate the mutual influence between two modules with rigorous interpretation from an EM algorithm perspective. We show that under dual supervision, minimizing the positive terms in Eq.~\ref{eq:ctl-expectation} intrinsically maximizes the objective of a classical clustering algorithm. For simplicity, we assume that each semantic rank has the same number of sentence pairs $n$ and represents model parameters with $\Theta$. In \textsc{Emma}-X, we model the semantic relation of sentence pair $(\mathbf{x}_i, \mathbf{y}_k)$ through a joint distribution $P(\mathbf{x}_i, \mathbf{y}_k)$ with the semantic rank $c$ as a latent variable. Let $Q(c)$ be a prior distribution over the possible values of semantic ranks. That is $\sum_r Q(c=r)=1,Q(c) \geq 0$. The training goal is to maximize the following likelihood:
\begin{equation}
    \footnotesize
    \begin{split}
        \argmax_{\Theta} \sum_{\mathbf{x}_i \in \mathcal{X}} \sum_{\mathbf{y}_k \in \mathcal{Y}} \log P(\mathbf{x}_i, \mathbf{y}_k|\Theta)&= \argmax_{\Theta} \sum_{\mathbf{x}_i \in \mathcal{X}} \sum_{\mathbf{y}_k \in \mathcal{Y}} \log \sum_{r=1}^{N} P(\mathbf{x}_i, \mathbf{y}_k, c=r|\Theta) \\
        &\geq \argmax_{\Theta} \sum_{\mathbf{x}_i \in \mathcal{X}} \sum_{\mathbf{y}_k \in \mathcal{Y}} \sum_{r=1}^{N} Q(c=r) \log  \frac{P(\mathbf{x}_i, \mathbf{y}_k,c=r|\Theta)}{Q(c=r)}. 
    \end{split}
    \label{eq:gen-model-exp}
\end{equation}

\noindent\textbf{E-Step.} To make the inequality hold with equality, we have:
\begin{equation}
    \footnotesize
    \begin{split}
        Q(c=r) = \frac{P(\mathbf{x}_i, \mathbf{y}_k,c=r|\Theta)}{\sum_{j=1}^{N} P(\mathbf{x}_i, \mathbf{y}_k,c=j|\Theta)} = P(c=r|\mathbf{x}_i, \mathbf{y}_k,\Theta),
    \end{split}
\end{equation}
which is the posterior probability and is approximated by the prediction from GMM classifier. Since each sentence pair $(\mathbf{x}_i, \mathbf{y}_k)$ belongs to only one semantic rank, we approximate $Q(c=r) = \mathbb{I}(c_{\mathcal{G}}^* = r)$, which is a one-hot distribution. 

\noindent\textbf{M-Step.} We try to maximize the likelihood in Eq.~\ref{eq:gen-model-exp} under the semantic rank $c_{\mathcal{G}}^*$ :
\begin{equation}
    \footnotesize
    \begin{split}
        \argmax_{\Theta} \sum_{\mathbf{x}_i \in \mathcal{X}} \sum_{\mathbf{y}_k \in \mathcal{Y}} \sum_{r=1}^{N} Q(r) \log  \frac{P(\mathbf{x}_i, \mathbf{y}_k,r|\Theta)}{Q(r)} &\approx \argmax_{\Theta} \sum_{\mathbf{x}_i \in \mathcal{X}} \sum_{\mathbf{y}_k \in \mathcal{Y}} \sum_{r=1}^{N} \log P(\mathbf{x}_i, \mathbf{y}_k|c_{\mathcal{G}}^*=r,\Theta) \\
        & \geq \argmax_{\Theta} n (n - 1) \sum_{r=1}^{N}  \tilde{\mu}_r^2,
    \end{split}
    \label{eq:em_perspect}
\end{equation}
The above derivation uses the assumption that $P(\mathbf{x}_i, \mathbf{y}_k|c_{\mathcal{G}}^*=r,\Theta) \sim \mathcal{N}_r\big( 
\mathbf{x}_i - \mathbf{y}_k| \tilde{\mathbf{\mu}}_r, \tilde{\mathbf{\sigma}}_r\big)$, with $\tilde{\mu}_r$ and $\tilde{\sigma}_r$ being the mean value and standard deviation of the Euclidean distance between sentence pairs in semantic rank $r$. Detailed proof of Eq.~\ref{eq:em_perspect} is in \textbf{Appendix \ref{app:detail_eq}}.

Next, we prove that minimizing the positive terms in expectation $\mathcal{L}_{\mathbf{CTL}}(\mathcal{X},\mathcal{Y};\Theta_{\mathcal{M}})$ actually equal to maximizing a lower bound of Eq.~\ref{eq:em_perspect}. As we apply dual supervision, data in the contrastive label space also follows the distribution $\mathcal{N}_r\big( 
\mathbf{x}_i - \mathbf{y}_k| \tilde{\mathbf{\mu}}_r, \tilde{\mathbf{\sigma}}_r\big)$. Hence, under mild assumptions, we can get:
\begin{equation}
    \footnotesize
    \begin{split}
\mathcal{L}^+_{\mathbf{CTL}}(\mathcal{X},\mathcal{Y};\Theta_{\mathcal{M}})
    = n^2\sum_{r=1}^{N-1}\tilde{\mu}_r^2 
    \ <\ n (n - 1)\sum_{r=1}^{N} \tilde{\mu}_r^2
    \ \leq\  \sum_{\mathbf{x}_i \in \mathcal{X}} \sum_{\mathbf{y}_k \in \mathcal{Y}} \log P(\mathbf{x}_i, \mathbf{y}_k|\Theta),
    \end{split}
    \label{eq:em_prove}
\end{equation}
where $\mathcal{L}^+_{\mathbf{CTL}}(\cdot)$ means the positive terms. In the derivation, we use the intrinsic property of semantic ranks ($\tilde{\mu}_1 < \tilde{\mu}_2 < ...< \tilde{\mu}_N $). Detailed proof is in \textbf{Appendix \ref{app:detail_eq}}. Eq.~\ref{eq:em_prove} demonstrates that with dual supervision, minimizing the contrastive loss can partially maximize the likelihood in Eq.~\ref{eq:gen-model-exp}. 

\section{Experiments}

To thoroughly evaluate the performance of \textsc{Emma}-X, we conduct experiments on \textbf{\textsc{xrete}} benchmark to verify the transfer ability of \textsc{Emma}-X on various cross-lingual downstream tasks with strong baselines (pre-trained models: \textsc{Xlm-r} \citep{conneau-etal-2020-unsupervised}, \textsc{InfoXLM} \citep{chi-etal-2021-infoxlm}, \textsc{Hictl} \citep{wei2021on}, sentence models: LaBSE \citep{feng-etal-2022-language}, S-BERT \citep{reimers-gurevych-2020-making}) and ChatGPT in Section~\ref{sec:xrete_analysis}. See \textbf{Appendices \ref{app:benchmark_info}} and \textbf{\ref{app:baseline_info}} for details. We further conduct geometric analysis in Section~\ref{sec:geo_analysis} to better interpret the cross-lingual transferability in \textsc{Emma}-X. 

\subsection{Setup}

\noindent\textbf{Corpus \& Model.} We collect parallel corpora from CCAligned \citep{el-kishky-etal-2020-ccaligned}, CCMatrix \citep{schwenk-etal-2021-ccmatrix}, WMT \citep{akhbardeh-etal-2021-findings}, and MultiUN \citep{ziemski-etal-2016-united}, involving 94 languages with 3.2 billion sentence pairs. In addition, we add CC-100 \citep{conneau-etal-2020-unsupervised} as the large-scale monolingual corpus with about 800 billion sentences that covers 94 languages. The cross-lingual encoder starts from the well-trained \textsc{Xlm-r} large model \citep{conneau-etal-2020-unsupervised}. The GMM classifier is implemented as a mixture of Gaussian forms, each of which consists of a prior $\pi \in \mathbb{R}^1$, a mean $\mu \in \mathbb{R}^{1024}$ and a standard deviation $\sigma \in \mathbb{R}^{1024}$, all are trainable variables. We set the total semantic ranks as $N=4$. The statistics of all data and hyper-parameters are shown in \textbf{Appendix \ref{appendix:pretraining-data}}.

\subsection{\textsc{xrete} Evaluation}\label{sec:xrete_analysis}
\textbf{\textsc{xrete}} includes 12 cross-lingual tasks divided into 4 different categories. We report the ``translate-train''
performance in Table \ref{tbl:main-results} on most tasks but zero-shot performance on BUCC and Tatoeba following~\citep{ruder-etal-2021-xtreme}. Table~\ref{tbl:zero-results} presents zero-shot comparisons with sentence models.

\begin{table*}[t]
\centering
\footnotesize
\resizebox{\textwidth}{!}{
\begin{tabular}{lcccccccccccc}
\toprule
\multirow{2}{*}{\textbf{Model}} & \multicolumn{2}{c}{\textbf{Inference}} & \multicolumn{2}{c}{\textbf{Similarity}} & \multicolumn{4}{c}{\textbf{Retrieval}} 
& \multicolumn{4}{c}{\textbf{Classification}}
\\
& XNLI & ANLI & MultiSTS & QE & LAReQA & Mewsli-X & BUCC & Tatoeba & XCOPA & MultiEURLEX & MultiARC & PAWS-X\\
\cmidrule{1-13}
Metrics & Acc. ($\color{green}{\uparrow}$) & Acc. ($\color{green}{\uparrow}$) & Spearman ($\color{green}{\uparrow}$) & Pearson ($\color{green}{\uparrow}$) & mAP@20 ($\color{green}{\uparrow}$) & mAP@20 ($\color{green}{\uparrow}$) & F1 ($\color{green}{\uparrow}$) & Acc. ($\color{green}{\uparrow}$) & Acc. ($\color{green}{\uparrow}$) & Acc. ($\color{green}{\uparrow}$) & MAE ($\color{red}{\downarrow}$) & Acc. ($\color{green}{\uparrow}$)\\
\midrule
\textsc{mBert}$^*$ & 75.1$^a$ & - & 55.8$^s$ & - & 21.6$^d$ & 38.6$^d$ & 56.7$^a$ & 39.0$^a$ & 56.1$^d$ & 67.4$^s$ & 48.2$^s$ & 81.9$^a$ \\
\textsc{Xlm}$^*$ & 77.8$^b$ & - & - & - & - & - & 56.8$^a$ & 32.6$^a$ & - & - & - & 80.9$^a$ \\
\textsc{Xlm-r}$^*$ & 83.6$^b$ & 49.12$^s$ & 61.5$^s$ & 58.7$^s$ & 40.7$^d$ & 45.7$^d$ & 66.0$^a$ & 57.7$^a$ & 69.2$^d$ & 66.6$^s$ & - & 88.9$^a$ \\
\textsc{Hictl}$^*$ & 84.7$^c$ & - & - & - & - & - & 77.6$^c$ & 69.1$^c$ & - & - & - & 92.8$^c$ \\
\textsc{ChatGPT}$^\dag$ & 60.9 & 41.7 & 68.6 & 60.9 & - & - & - & - & 74.2 & 68.7 & 40.2 & 64.2 \\
\midrule
\multicolumn{13}{l}{\emph{Ours re-implementation, translate-train-all (models are trained on English training data and on its data translated to the target language)
}} \\
\midrule
\textsc{Xlm-r}$^\ddag$ & 82.8 & 48.48 & 65.9 & 63.2 & 40.3 & 48.6 & 67.9 & 59.1 & 71.2 & 66.9 & 44.9 & 90.1 \\
\textsc{InfoXLM}$^\ddag$ & 84.2 & 49.10 & 82.2 & 64.1 & 44.9 & 57.1 & 77.4 & 66.2 & 74.6 & 67.7 & 36.2 & 93.0 \\
\textsc{Hictl}$^\ddag$ & 85.1 & 49.02 & 81.6 & 64.9 & 46.1 & 54.8 & 77.6 & 65.8 & 74.8 & 68.3 & 38.2 & 92.8 \\
\midrule
\textbf{\textsc{Emma-X}} & \bf 88.1 & \bf 50.21 & \bf 87.3 & \bf 67.2 & \bf 50.6 & \bf 59.6 & \bf 87.1 & \bf 82.5 & \bf 78.2 & \bf 71.4 & \bf 32.7 & \bf 94.2 \\
\bottomrule
\end{tabular}%
}
\caption{Results on the \textsc{xrete} benchmark. $^*$ denotes the results from previous literature, $^a$\citet{pmlr-v119-hu20b} $^b$\citet{conneau-etal-2020-unsupervised} $^c$\citet{wei2021on} $^d$\citet{ruder-etal-2021-xtreme}. $^s$ denotes the results from the original paper. $^\ddag$ denotes results from our re-trained models with the same model size and training corpora as \textsc{Emma}-X. $^\dag$ denotes the zero-shot performance. Refer to \textbf{Appendix \ref{appendix:results-per-language}} for greater details on each task and language.}\label{tbl:main-results}
\end{table*}

\begin{table}[!t]
\centering
\begin{minipage}[t]{0.58\textwidth}
\makeatletter\def\@captype{table}
\footnotesize
\resizebox{\textwidth}{!}{
\begin{tabular}{lcccccc}
\toprule
\multirow{2}{*}{\textbf{Model}} & \multicolumn{2}{c}{\textbf{Similarity}} & \multicolumn{4}{c}{\textbf{Retrieval}} 
\\
& {MultiSTS} & {QE} & {LAReQA} & {Mewsli-X} & {BUCC} & {Tatoeba} \\
\cmidrule{1-7}
Metrics & {Spearman ($\color{green}{\uparrow}$)} & {Pearson ($\color{green}{\uparrow}$)} & {mAP@20 ($\color{green}{\uparrow}$)} & {mAP@20 ($\color{green}{\uparrow}$)} & {F1 ($\color{green}{\uparrow}$)} & {Acc. ($\color{green}{\uparrow}$)} \\
\midrule
\textsc{S-BERT} & \textbf{84.0} & 39.3 & \textbf{31.8} & \textbf{14.4} & \textbf{88.5} & 68.6 \\
LaBSE & \textbf{74.4} & 31.6 & 12.8 & 11.2 & \textbf{93.2} & \textbf{83.6} \\
\textsc{InfoXLM} &  55.2 & \textbf{49.4} & 16.9 & 23.9 & 77.4 & 66.2 \\
\textsc{Xlm-r} &  25.0 & 10.4 & 16.1 & 11.7 & 67.9 & 59.1 \\
\midrule
\textbf{\textsc{Emma-X}} & 62.9 & \textbf{54.7} & \textbf{19.4} & \textbf{29.4} & 87.1 & \textbf{82.5} \\
\bottomrule
\end{tabular}}
 \vspace{5pt}
\caption{Zero-shot Results on Similarity and Retrieved tasks. Results of LaBSE use Customized Vocab setting. Results of S-BERT are from \textbf{XLM-r}$\leftarrow$\textbf{SBERT-paraphrases}. The bold font denotes the best 2 results.}\label{tbl:zero-results}
\end{minipage}\quad
\begin{minipage}[t]{0.38\textwidth}
\makeatletter\def\@captype{table}
\footnotesize
\resizebox{\textwidth}{!}{
\begin{tabular}{lcccc}
\toprule
\multirow{2}{*}{\textbf{Model}} & \multicolumn{2}{c}{\textbf{FLORES-200}} & \multicolumn{2}{c}{\textbf{Tatoeba}} 
\\
& {Head} & {Long-tail} & {Head} & {Long-tail} \\
\midrule
Metrics & {Acc. ($\color{green}{\uparrow}$)} & {Acc. ($\color{green}{\uparrow}$)} & {Acc. ($\color{green}{\uparrow}$)} & {Acc. ($\color{green}{\uparrow}$)} \\
\midrule
\textsc{S-BERT} & 87.5 & 51.6 & 91.4 & 57.8 \\
LaBSE & \textbf{99.9} & 82.5 & \textbf{95.7} &  77.9 \\
\textsc{InfoXLM} &  83.4 & 53.8 & 87.8 &  56.1 \\
\textsc{Xlm-r} &  68.2 & 45.3 & 66.3 &  55.7 \\
\midrule
\textbf{\textsc{Emma-X}} & 94.5 & \textbf{84.2} & 91.9 & \textbf{78.1} \\
\bottomrule
\end{tabular}}
 \vspace{5pt}
\caption{Retrieval results on FLORES-200 and Tatoeba in xx $\rightarrow$ En direction. The bold font denotes the best results.}\label{tbl:long-tail-results}
\end{minipage}
\vspace{-15pt}
\end{table}

\noindent\textbf{Comparisons with Pre-trained Models.} In Table~\ref{tbl:main-results}, \textsc{Emma}-X consistently outperforms all baseline models (\textsc{Xlm-r}~\citep{conneau-etal-2020-unsupervised}, \textsc{Hictl} \citep{wei2021on} and \textsc{InfoXLM} \citep{chi-etal-2021-infoxlm}) with 7.97\% improvements on average. Specifically, \textsc{Emma}-X achieves 88.1\% accuracy on XNLI~\citep{conneau-etal-2018-xnli} and 50.21\% accuracy on ANLI~\citep{ebrahimi-etal-2022-americasnli} with up to 6.4\% improvements than baselines. On the MultiSTS~\citep{reimers-gurevych-2020-making} task, \textsc{Emma}-X achieves an 87.3 correlation score, outperforming several strong baselines by 5.1$\sim$21.4, and even achieves comparable performance in the cross-lingual and the monolingual settings (see \textbf{Appendix \ref{appendix:results-per-language}} for language-specific results). Furthermore, \textsc{Emma}-X obtains a 67.2 Pearson score on QE~\citep{specia-etal-2021-findings} task, which is comparable to the winner on the leaderboard\footnote{\url{https://www.statmt.org/wmt21/quality-estimation-task_results.html}} without any specific finetuning techniques. As for sentence retrieval, \textsc{Emma}-X consistently outperforms previous strong baselines among all 4 tasks~\citep{ruder-etal-2021-xtreme}, and demonstrates 2.6\%$\sim$39.6\% improvements over these baselines. Similar results can be found in sentence classification tasks. \textsc{Emma}-X obtains an 81.3\% top-1 accuracy averaged on XCOPA~\citep{ponti-etal-2020-xcopa}, MultiEURLEX~\citep{chalkidis-etal-2021-multieurlex} and PAWS-X~\citep{yang-etal-2019-paws} tasks, outperforming \textsc{Xlm-r}, \textsc{InfoXLM} and \textsc{Hictl} by 7.0\%, 3.9\% and 3.5\% improvements, respectively. On MultiARC~\citep{keung-etal-2020-multilingual} task, \textsc{Emma}-X shows the lowest error rates among all models. The consistent improvements on all tasks reveal that \textsc{Emma}-X can obtain better universal representations for different natural languages with various topics and domains. We further conduct experiments with ChatGPT on \textsc{xrete} tasks without 4 Retrieval tasks. We list the prompts for each task in \textbf{Appendix \ref{app:gpt-prompt}}. ChatGPT's zero-shot performance is worse than fine-tuned pre-trained models and the performance gap is very large on most tasks. 


\noindent\textbf{Zero-shot comparisons with sentence models.} 
Compared with \textsc{Xlm-r} and \textsc{InfoXLM}, which adopt the same amount of training data as \textsc{Emma}-X, \textsc{Emma}-X consistently outperforms \textsc{Xlm-r} and \textsc{InfoXLM} by 73.2\% and 25.1\% on average, as shown in Table~\ref{tbl:zero-results}. The results further prove the effectiveness of our pre-training technique. Through the reciprocation between GMM classifier and cross-lingual encoder, \textsc{Emma}-X can generate reliable semantic rank for multilingual non-parallel corpora, which can provide more supervision signals than previous baselines. \textsc{Emma}-X even achieves comparable results with strong supervised methods: LaBSE \citep{feng-etal-2022-language} and S-BERT \citep{reimers-gurevych-2020-making}, which both trained on supervised data. LaBSE is trained on a fine-filtered bilingual corpus with 6B translation pairs (2 times larger than \textsc{Emma}-X), while S-BERT is distilled from a S-BERT model fine-tuned on English NLI, STS datasets, and 50M English paraphrase pairs. Compared with these two methods, \textsc{Emma}-X can achieve the best results on QE and Mewsli-X by outperforming S-BERT and LaBSE by 71.7\% and 117.8\% averaged. \textsc{Emma}-X performs worse than these baselines on MultiSTS and BUCC, for these two tasks only contain rich-resource languages, which already have great deal of parallel data. 

\begin{table*}[ht]
\centering
\footnotesize
\resizebox{\textwidth}{!}{
\begin{tabular}{l|ccc|ccc|ccc}
\toprule
\multirow{2}{*}{\textbf{Model}} & \multicolumn{3}{c|}{\textbf{Head Langs.}} & \multicolumn{3}{c|}{\textbf{Long-tail Langs.}} & \multicolumn{3}{c}{\textbf{All Langs.}} \\
& Invariance & Canonical Form & Isotropy & Invariance & Canonical Form & Isotropy & Invariance & Canonical Form & Isotropy\\
\midrule
Metrics & KL-D ($\color{red}{\downarrow}$) & CH-I ($\color{green}{\uparrow}$) & PR ($\color{green}{\uparrow}$) & KL-D ($\color{red}{\downarrow}$) & CH-I ($\color{green}{\uparrow}$) & PR ($\color{green}{\uparrow}$) & KL-D ($\color{red}{\downarrow}$) & CH-I ($\color{green}{\uparrow}$) & PR ($\color{green}{\uparrow}$)\\
\midrule
\textsc{Xlm-r} (cls) & 0.7356 & 30.19 & 0.3681 & 2.0042 & 7.96 & 0.3686 & 1.6501 & 20.80 & 0.3683\\
\textsc{InfoXLM} (cls) & 0.4491 & 38.82 & 0.4478 & 1.8555 & 13.02 & 0.4406 & 1.4747 & 31.51 & 0.4665\\
S-BERT (mean) & \textbf{0.1115} & \textbf{108.22} & 0.4519 & 1.3112 & 44.32 & 0.4414 & \textbf{0.9782} & \textbf{102.36} & 0.4467\\
\midrule
\textsc{Emma-X} (cls) & 0.3603 & 43.52 & \bf 0.5318 & \bf 0.3963 & \bf 46.53 &  \textbf{0.5732} & 1.1904 & 48.70 & \bf 0.5918\\
\bottomrule
\end{tabular}%
}
\caption{Comparisons with existing methods on FLORES dataset for geometric analysis. ``cls'' and ``mean'' represent different pooling strategies to obtain sentence representations.}\label{tbl:geo-analysis}
\end{table*}

\noindent\textbf{Performance on Long-tail Languages.} 
One goal of \textsc{Emma}-X is to learn universal sentence representations accommodated for more languages. To better prove this, we report the retrieval accuracy on FLORES-200~\citep{costa2022no} and Tatoeba~\citep{artetxe-schwenk-2019-massively}. We reformulate FLORES-200, which contains manual translations in 204 languages (totaling 3001 sentences) to perform retrieval tasks in the same way as Tatoeba and report the performance in terms of language data scale in Table~\ref{tbl:long-tail-results}. Details about the separation of languages and FLORES-200  are shown in \textbf{Appendices A} and \textbf{B}. On head languages, \textsc{Emma}-X performs worse than LaBSE by about 4.6\% but outperforms S-BERT by 3.5\%. On the long-tail languages, \textsc{Emma}-X can surpass S-BERT by 4.3\% averaged on two tasks. \textsc{Emma}-X can even exceed the strongest LaBSE by 2.1\% on FLORES. One reason for the superior results on long-tail languages is that for those long-tail languages that have only bi-lingual parallel data with rich-resource languages (often English), \textsc{Emma}-X can provide multi-lingual semantic relation signals for them with arbitrary languages through dual supervision. 




\subsection{Geometric Analysis}\label{sec:geo_analysis}
\begin{figure*}[t]
\footnotesize
\begin{subfigure}{.24\linewidth}
    \centering
    \resizebox{\linewidth}{!}{
        \includegraphics{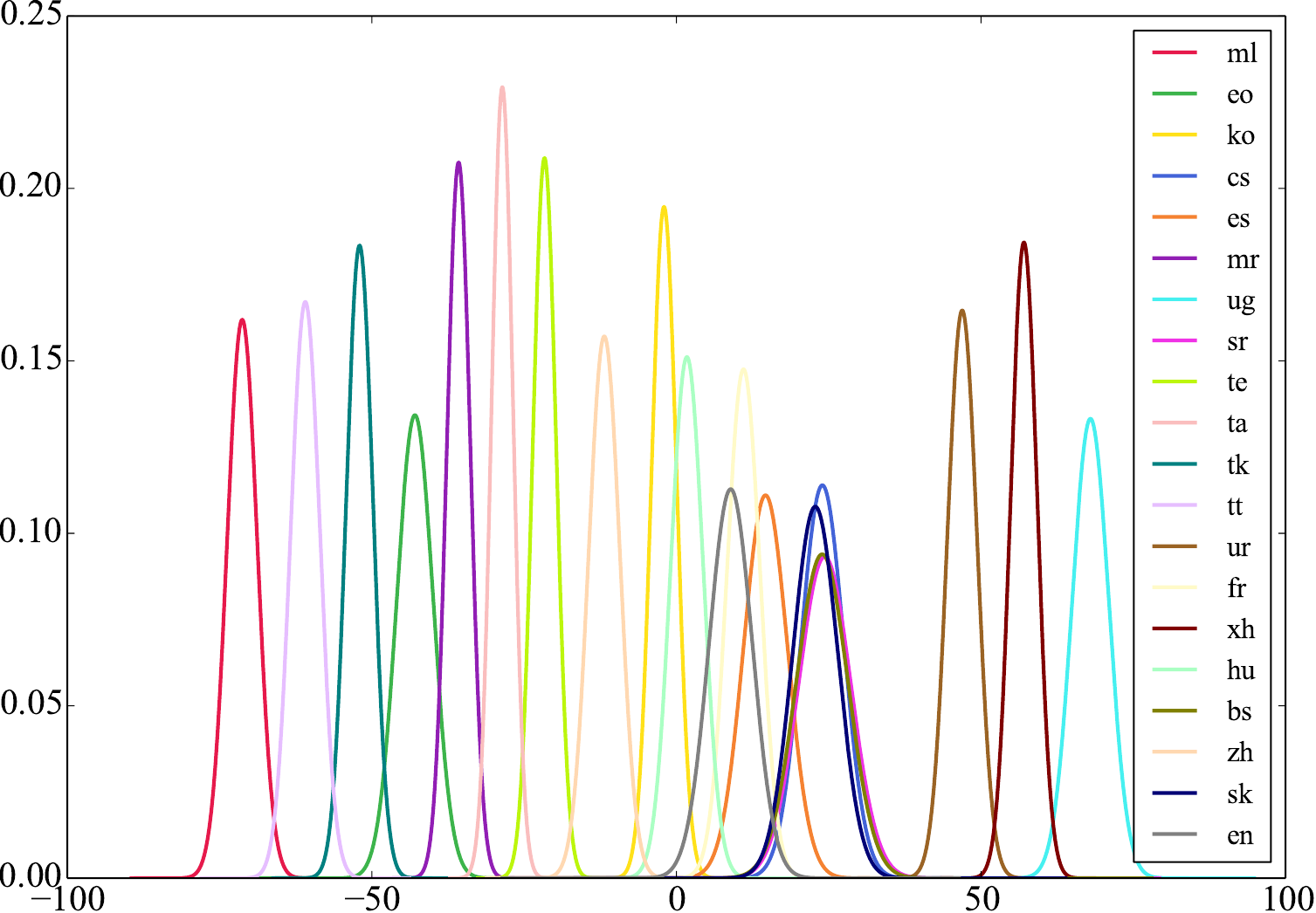}
    }
    \caption{\textsc{Xlm-r} (Invariance)}
    \label{fig:/xlmr_kl_div_2d_crop}
\end{subfigure}
\begin{subfigure}{.24\linewidth}
    \centering
    \resizebox{\linewidth}{!}{
        \includegraphics{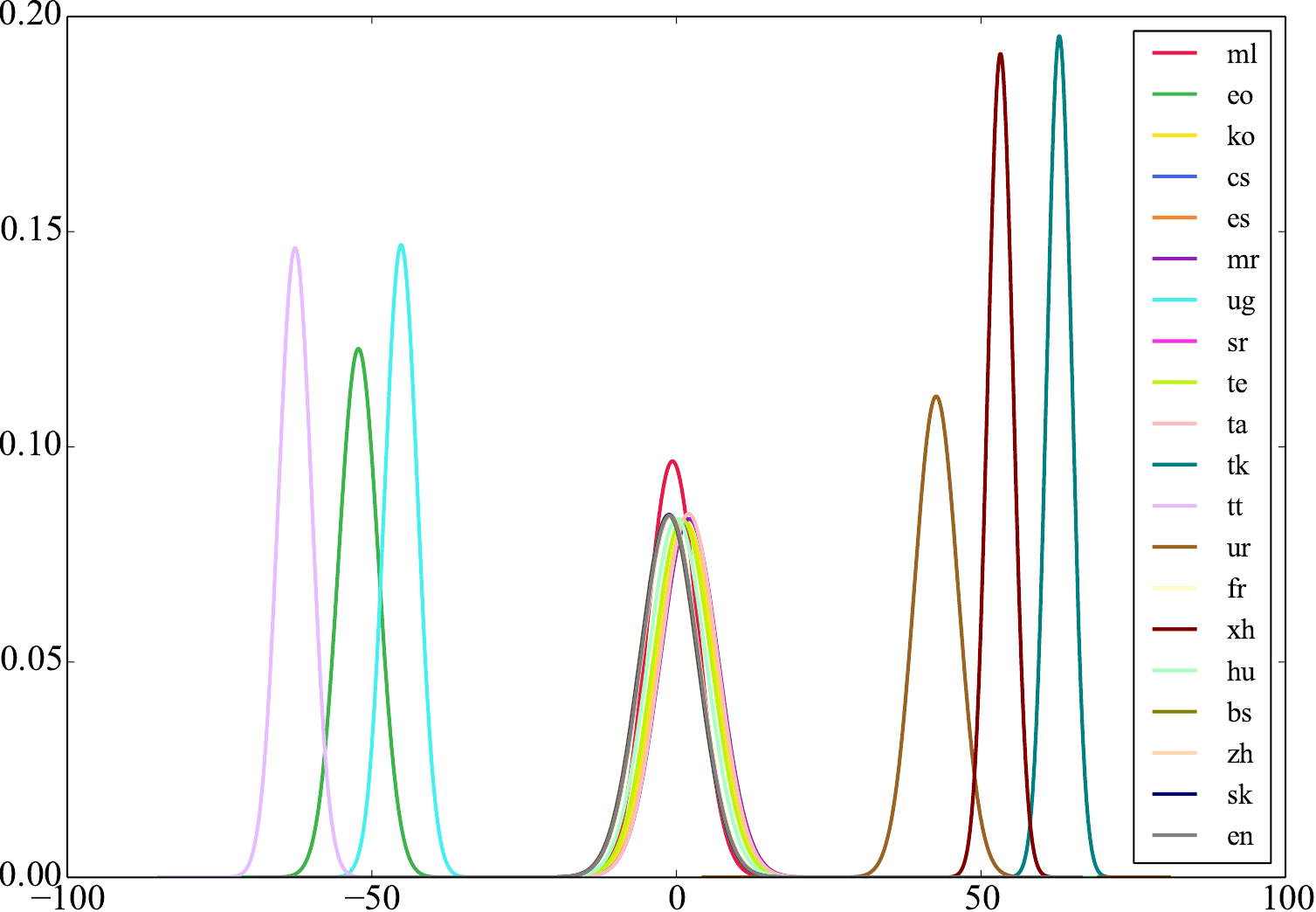}
    }
    \caption{\textsc{InfoXLM} (Invariance)}
    \label{fig:/infoxlm_kl_div_2d_crop}
\end{subfigure}
\begin{subfigure}{.24\linewidth}
    \centering
    \resizebox{\linewidth}{!}{
        \includegraphics{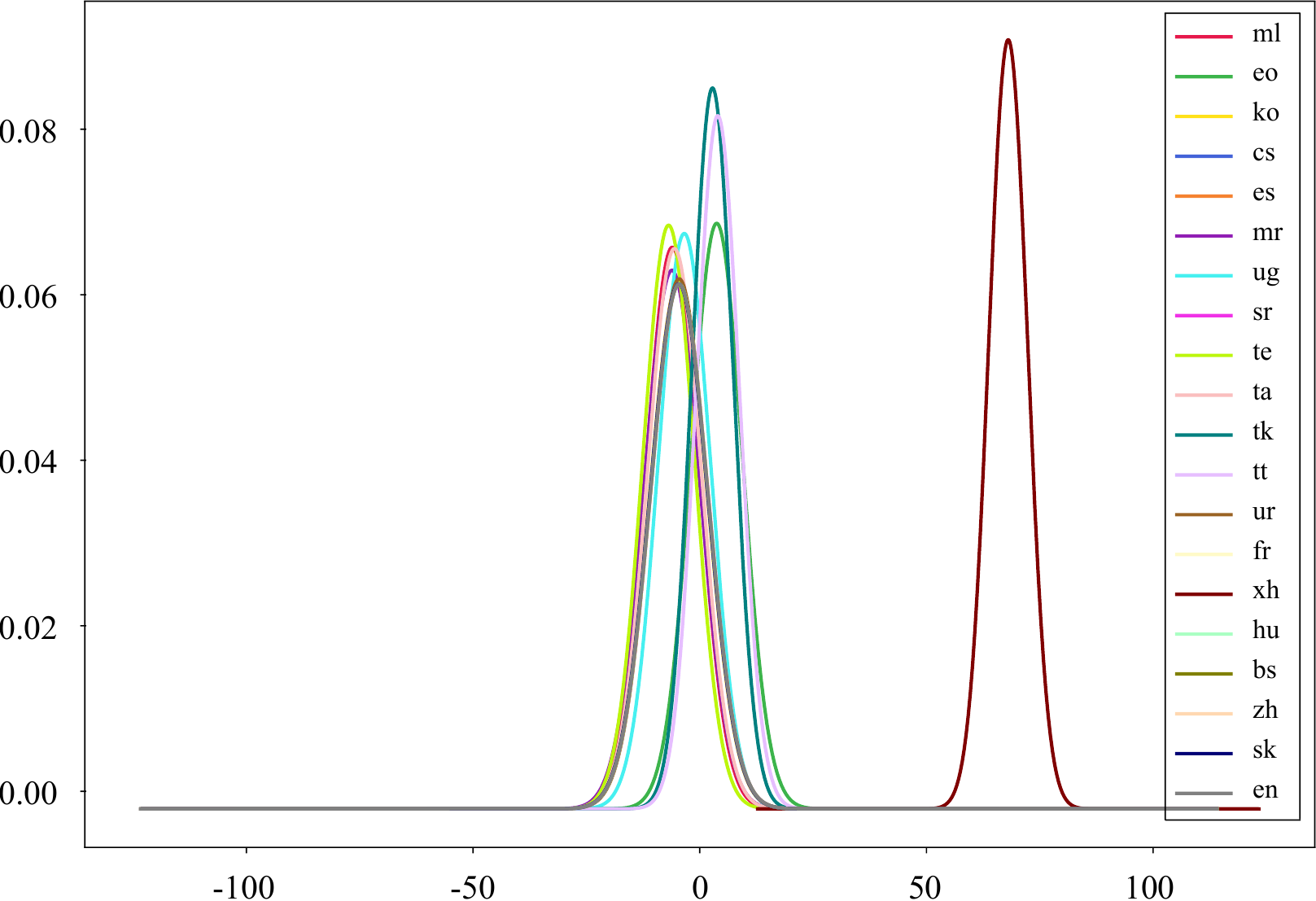}
    }
    \caption{S-BERT (Invariance)}
    \label{fig:/sbert_kl_div_2d_crop}
\end{subfigure}
\begin{subfigure}{.24\linewidth}
    \centering
    \resizebox{\linewidth}{!}{
        \includegraphics{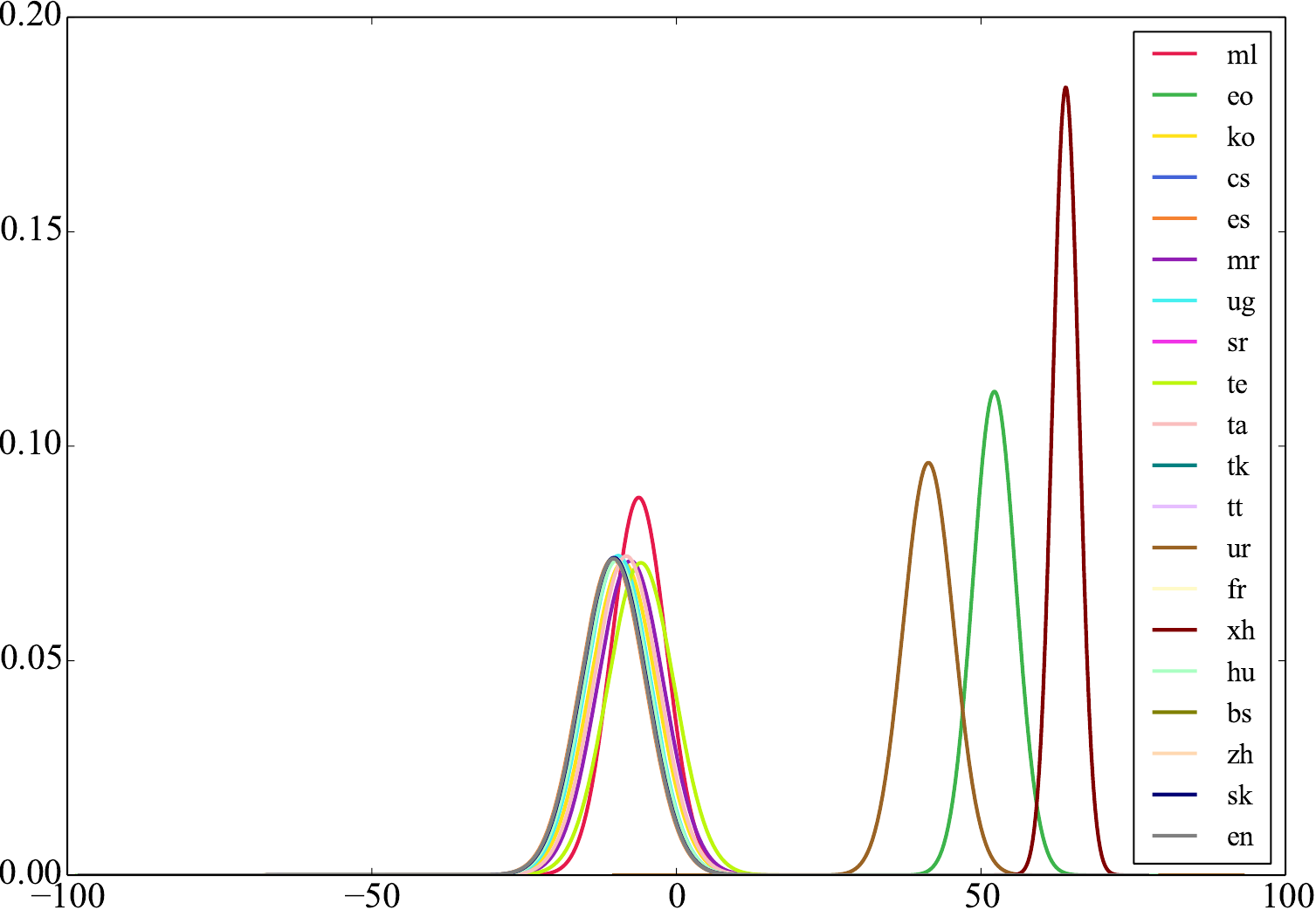}
    }
    \caption{\textsc{Emma}-X (Invariance)}
    \label{fig:/emma_kl_div_2d_crop}
\end{subfigure}
\begin{subfigure}{.24\linewidth}
    \centering
        \includegraphics[scale=0.135]{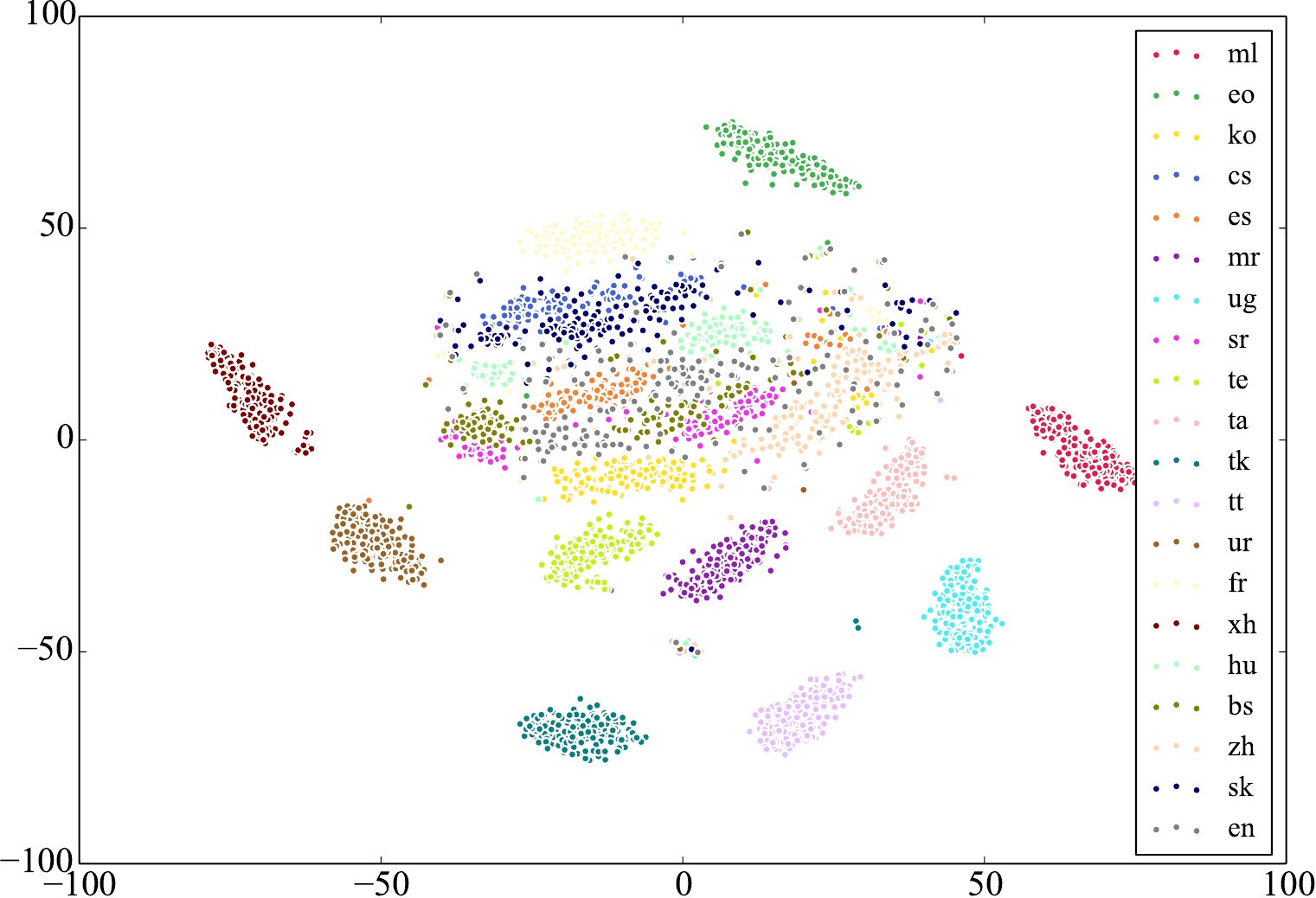}
    \caption{\textsc{Xlm-r} (Canonical)}
    \label{fig:/xlmr_canonical_crop}
\end{subfigure}
\begin{subfigure}{.24\linewidth}
    \centering
        \includegraphics[scale=0.135]{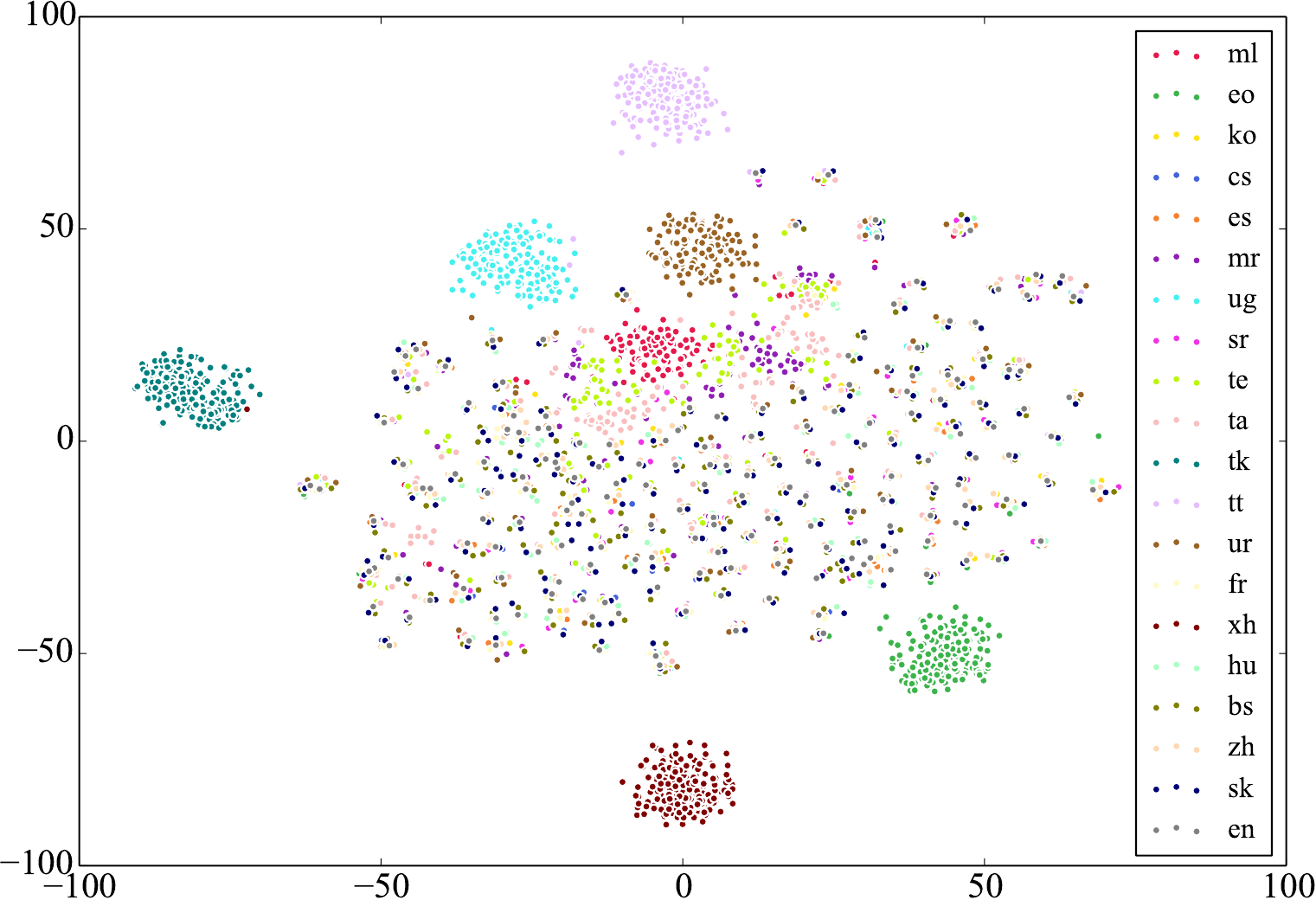}
    \caption{\textsc{InfoXLM} (Canonical)}
    \label{fig:/infoxlm_canonical_crop}
\end{subfigure}
\begin{subfigure}{.24\linewidth}
    \centering
        \includegraphics[scale=0.125]{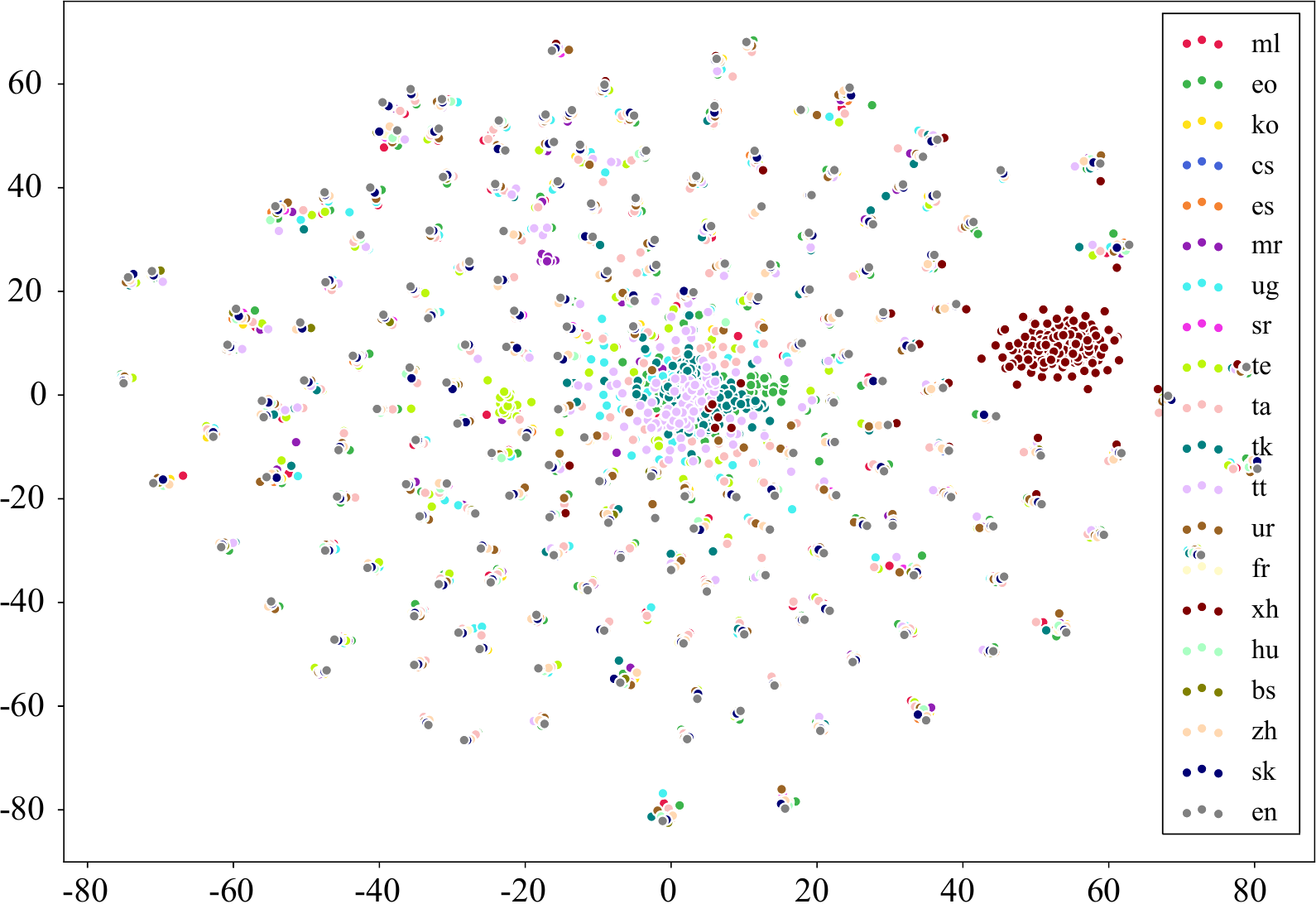}
    \caption{S-BERT (Canonical)}
    \label{fig:/sbert_canonical_crop}
\end{subfigure}
\begin{subfigure}{.24\linewidth}
    \centering
        \includegraphics[scale=0.13]{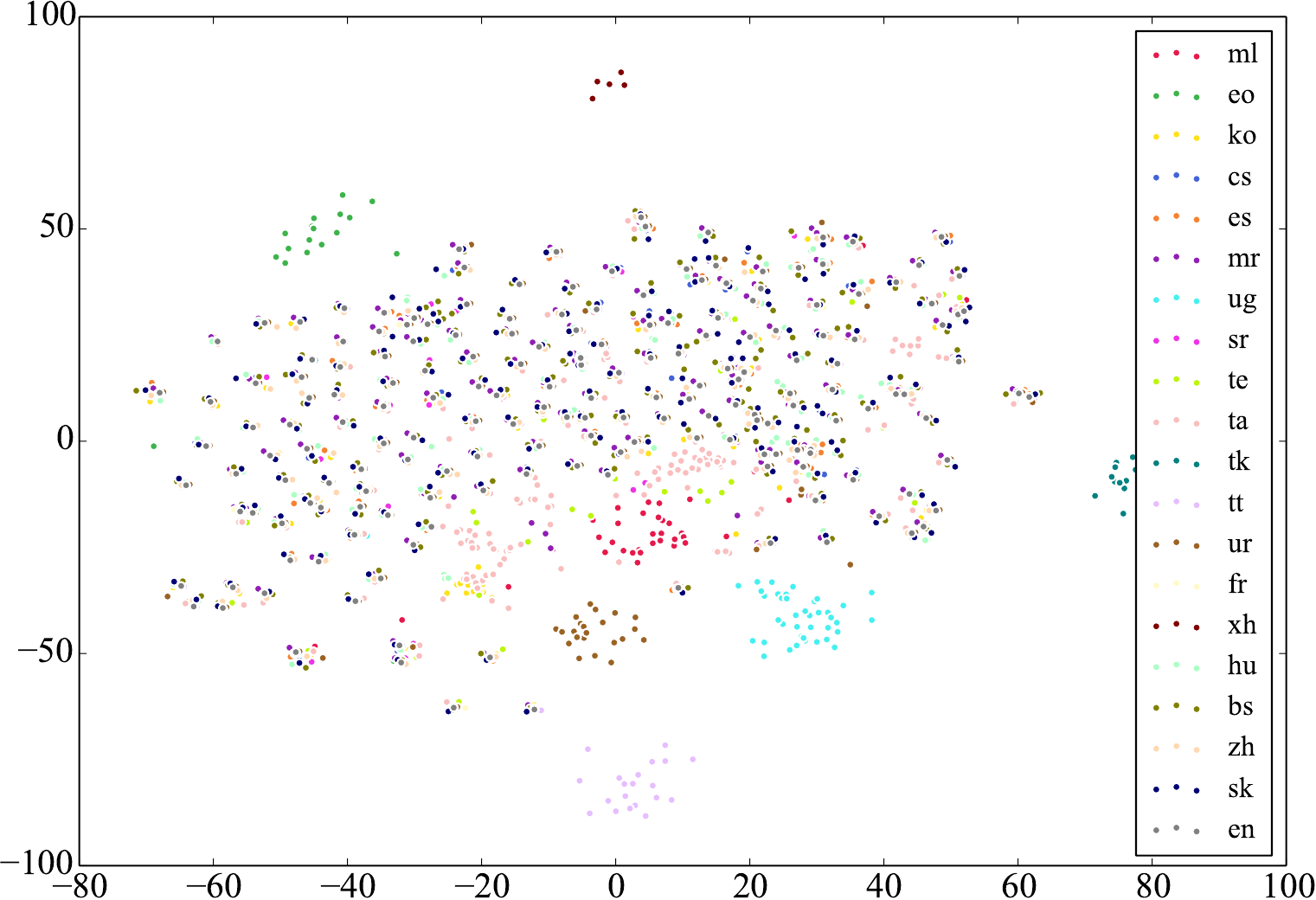}
    \caption{\textsc{Emma}-X (Canonical)}
    \label{fig:/emma_canonical_crop}
\end{subfigure}
\begin{subfigure}{.24\linewidth}
    \centering
        \includegraphics[scale=0.1]{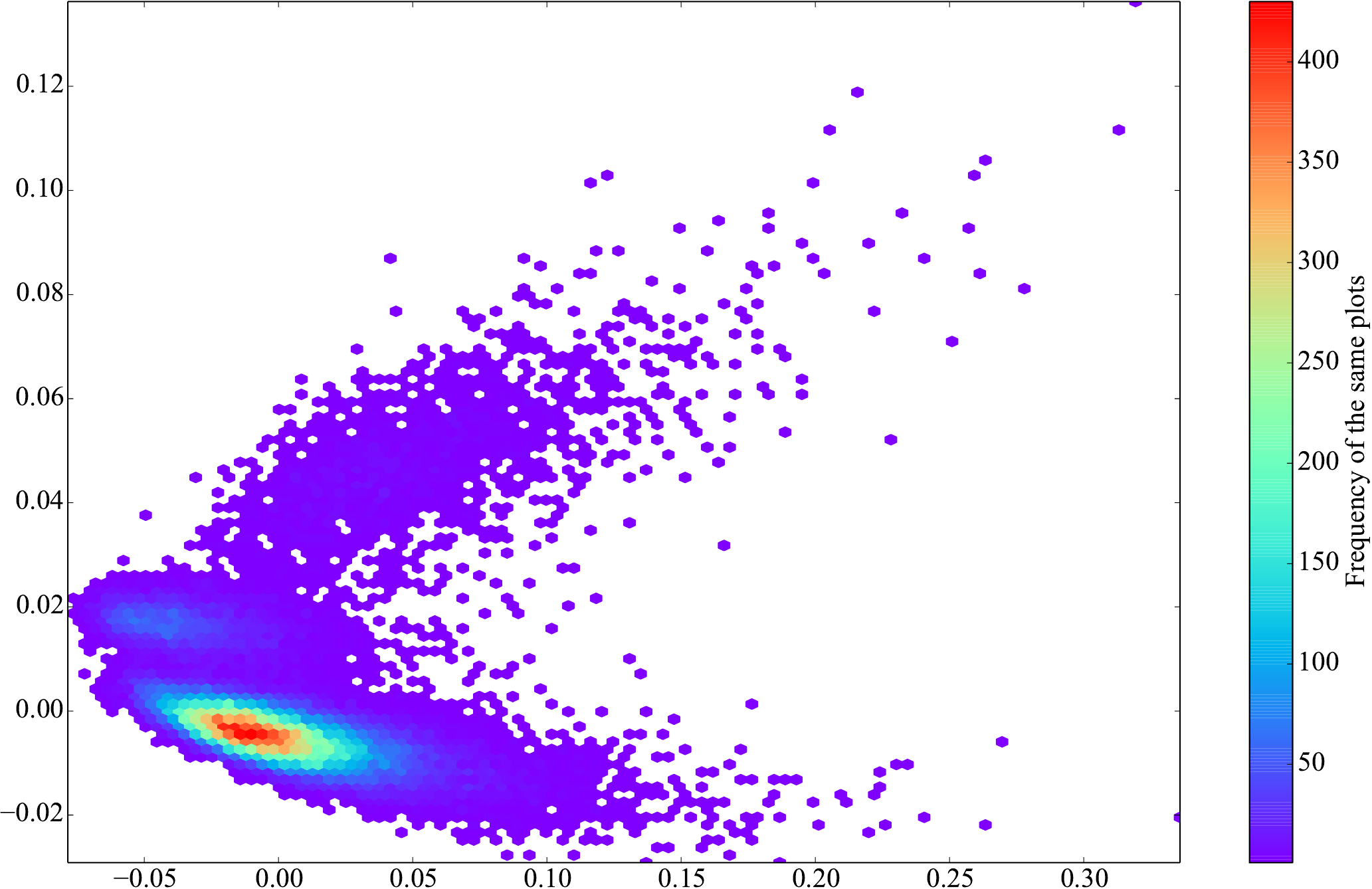}
    \caption{\textsc{Xlm-r} (Isotropy)}
    \label{fig:/xlmr_isotropy_crop}
\end{subfigure}
\hspace{0.5mm}
\begin{subfigure}{.24\linewidth}
    \centering
        \includegraphics[scale=0.1]{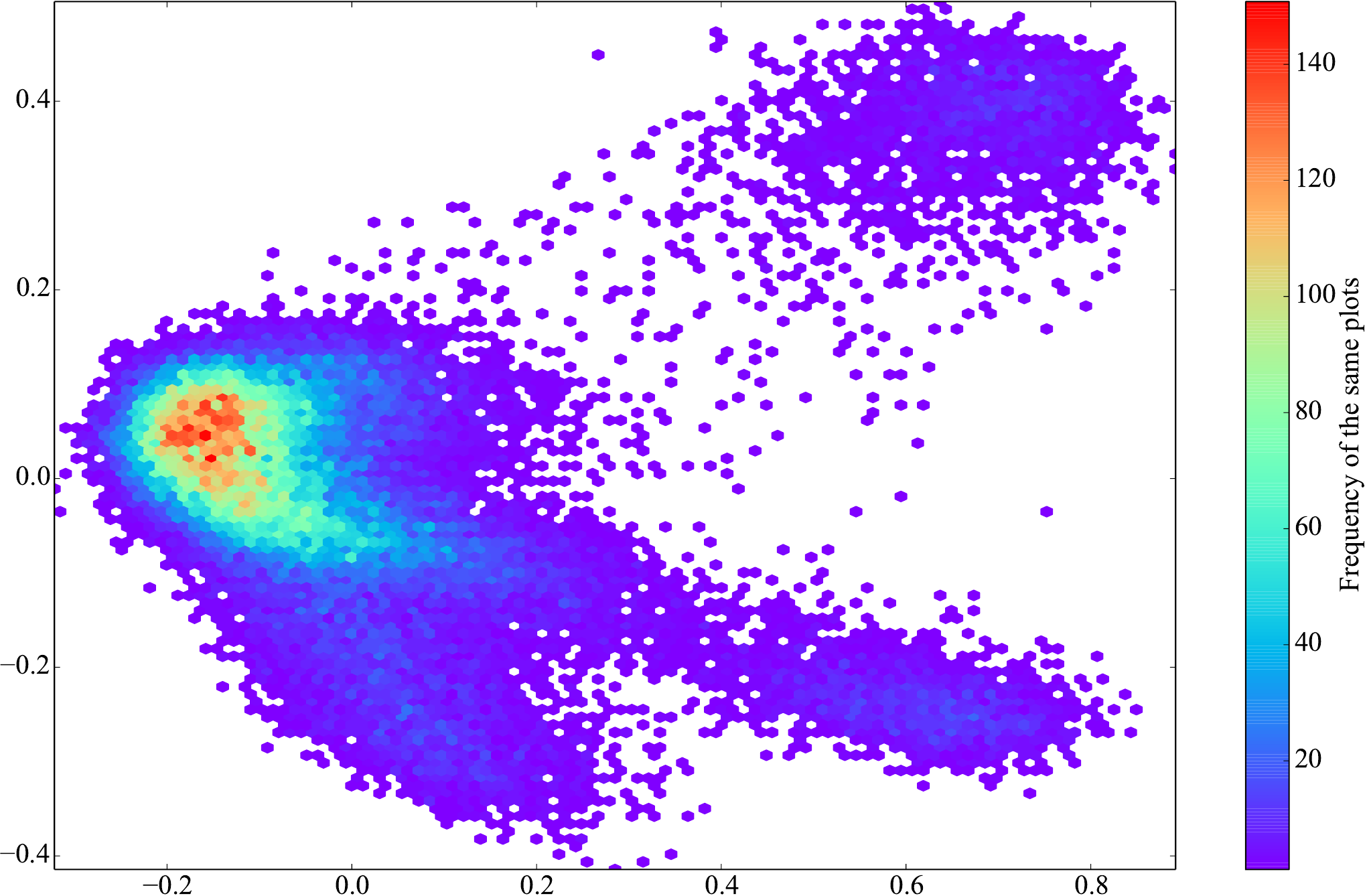}
    \caption{\textsc{InfoXLM} (Isotropy)}
    \label{fig:/infoxlm_isotropy_crop}
\end{subfigure}
\hspace{0.6mm}
\begin{subfigure}{.24\linewidth}
    \centering
        \includegraphics[scale=0.12]{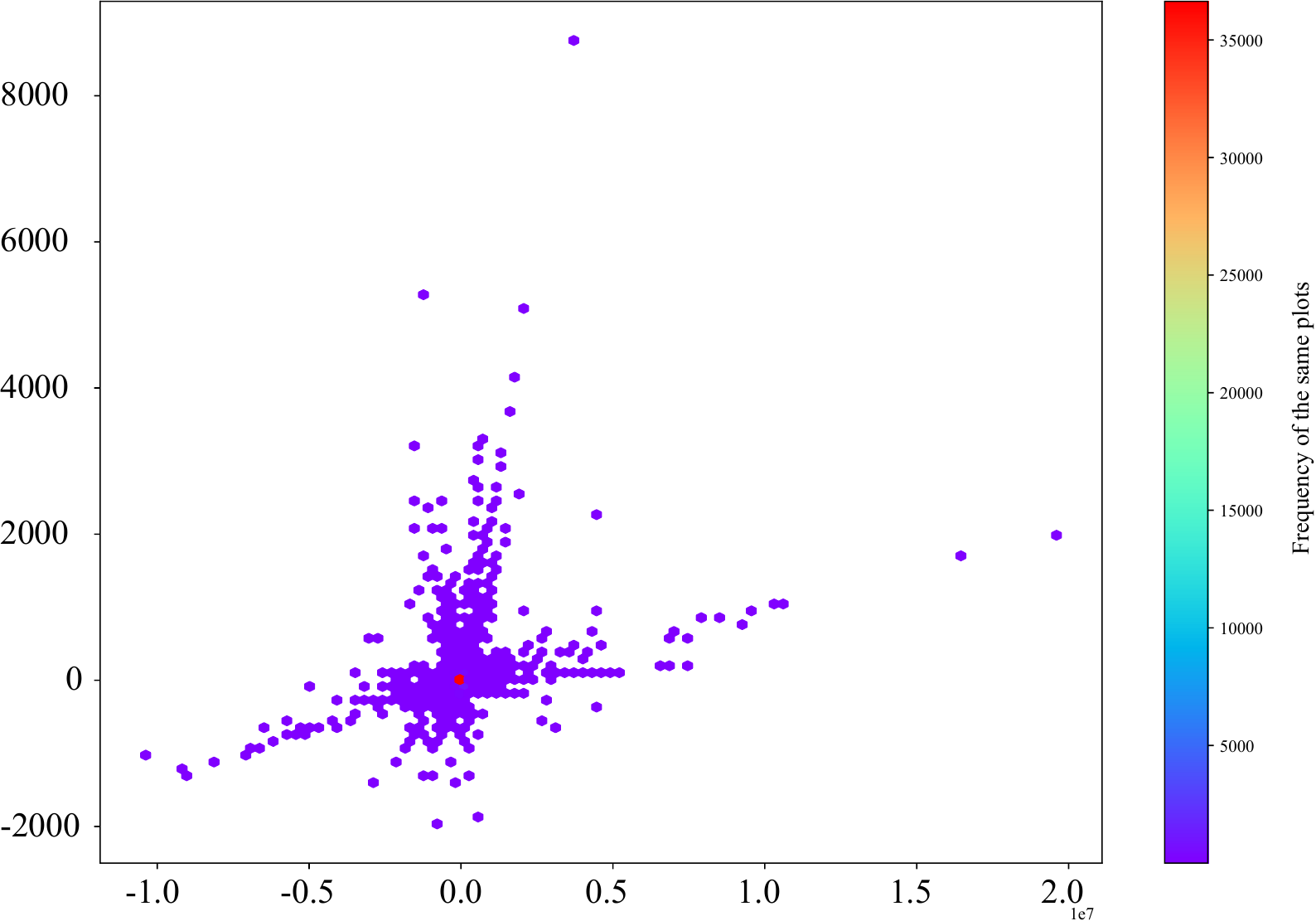}
    \caption{S-BERT (Isotropy)}
    \label{fig:/sbert_isotropy_crop}
\end{subfigure}
\hspace{0.5mm}
\begin{subfigure}{.24\linewidth}
    \centering
        \includegraphics[scale=0.1]{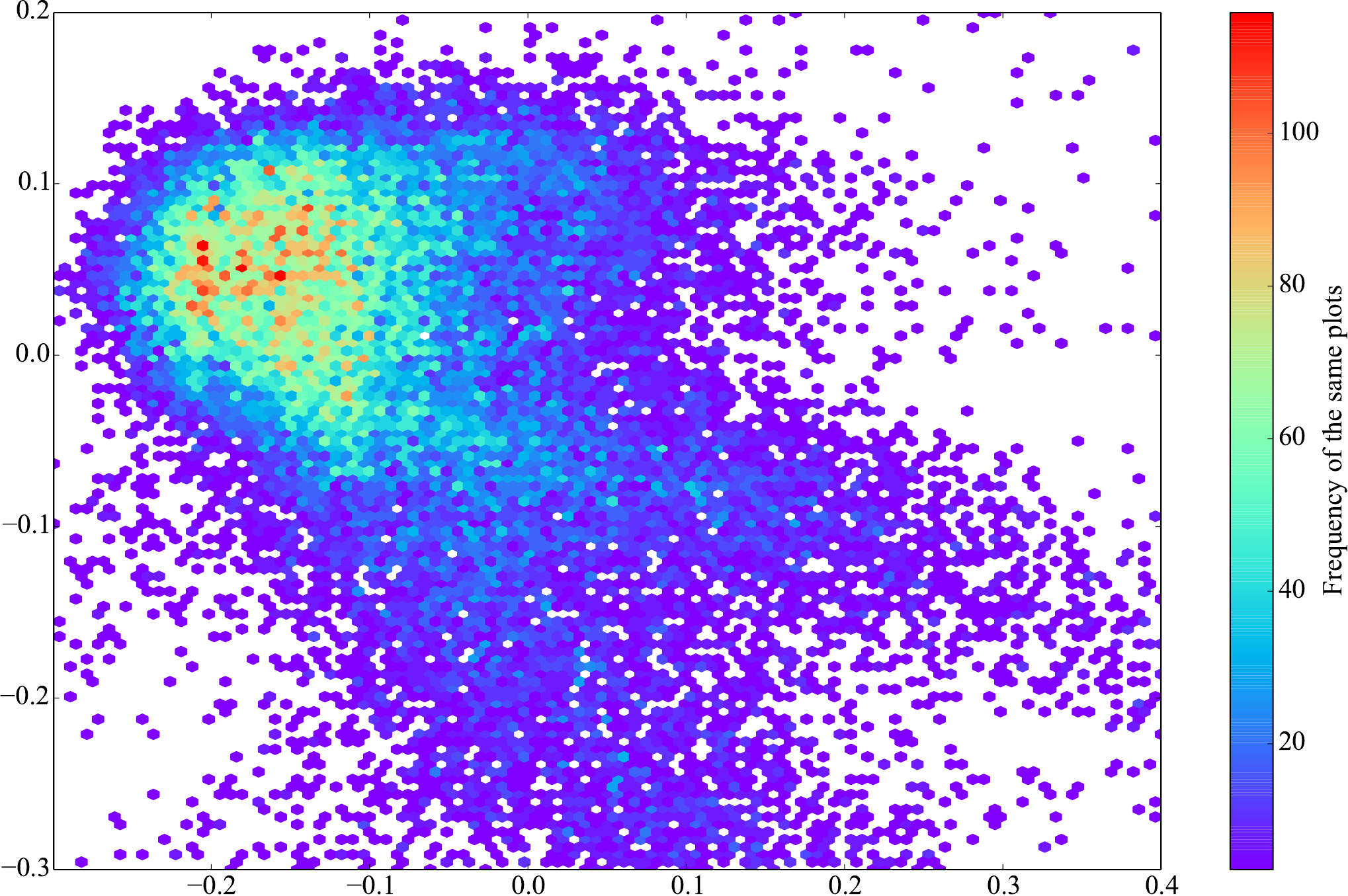}
    \caption{\textsc{Emma}-X (Isotropy)}
    \label{fig:/emma_isotropy_crop}
\end{subfigure}
\caption{Visualization of representations from \textsc{Emma}-X, \textsc{Xlm-r}, \textsc{InfoXLM} and S-BERT. We use t-SNE~\citep{JMLR:v9:vandermaaten08a} to visualize each geometric metrics.}
\label{fig:geometry_analysis}
\end{figure*}

To interpret the advantages of \textsc{Emma-X}, we evaluate the geometric characteristics of it on FLORES-200 dataset \citep{costa2022no} without any fine-tuning. The criteria of three requirements are Invariance, measured with KL-divergence (KL-D)~\citep{10.1214/aoms/1177729694}, Canonical Form, measured with Calinski-Harabasz Index (CH-I) \citep{doi:10.1080/03610927408827101} and Isotropy, measured with principal ratio (PR)~\citep{mu2018allbutthetop}. Details of them are shown in \textbf{Appendix \ref{app:flores-and-three-metrics}}. We report the numerical results in Table~\ref{tbl:geo-analysis} and visualize each characteristic in Figure~\ref{fig:geometry_analysis}.

\noindent\textbf{Invariance \& Canonical Form} aim to measure how languages are aligned in the representation space. If the sentence representations are universal, then sentences in different languages should follow a similar distribution, which is measured by invariance in KL-divergence. Similarly, canonical form measures how well semantic equivalent sentences are grouped into one cluster, with a clustering metric (CH-I). In Table~\ref{tbl:geo-analysis}, S-BERT outperforms other baselines in ``Invariance'' and ``Canonical Form'' on head languages. However, \textsc{Emma}-X shows better performance on long-tail languages in these two metrics, which is consistent with Table~\ref{tbl:long-tail-results}. Figure~\ref{fig:geometry_analysis} presents similar results. Among the 20 languages we randomly sampled from FLORES, \textsc{Emma}-X can align 17 languages as shown in Figure~\ref{fig:/emma_kl_div_2d_crop}, with ``xh, eo, ur'' as outliers. In Figure~\ref{fig:/xlmr_canonical_crop}, \ref{fig:/sbert_canonical_crop}, \ref{fig:/infoxlm_canonical_crop} and \ref{fig:/emma_canonical_crop}, different colors represent different languages. So a cluster with only one color means this language is isolated from other languages and not aligned well in representation space. Figure~\ref{fig:/emma_canonical_crop} shows that \textsc{Emma}-X performs well in most languages. 

\noindent\textbf{Isotropy} measures how expressive a representation space is since high-dimensional representation space can easily deteriorate into a low-dimensional manifold. From Table~\ref{tbl:geo-analysis}, \textsc{Emma}-X achieves the best results in isotropy. The Isotropy of S-BERT (k) is different from other methods. We conjecture the reason is that S-BERT removes MLM tasks during fine-tuning, so token embeddings will only receive sentence-level supervision, resulting in identical sentence representations for different languages. The abnormal results observed on the high KL-divergence, as depicted in Table \ref{tbl:geo-analysis} and Figure \ref{fig:/sbert_kl_div_2d_crop}, can be attributed to the representation space for S-BERT deteriorating into a low-dimensional manifold (low isotropy score in Table \ref{tbl:geo-analysis} and Figure \ref{fig:/sbert_isotropy_crop}), and different languages are not distributed uniformly across the whole representation space, which limits the expressive ability.

\begin{table}[h]
\centering
\resizebox{\textwidth}{!}{%
\begin{tabular}{@{}lcccccccccccccccccc@{}}
\toprule
Model\textbackslash{}Language & {\color[HTML]{FE0000} \textbf{af}} & \textbf{ar} & \textbf{bg} & {\color[HTML]{FE0000} \textbf{bn}} & \textbf{de} & \textbf{el} & \textbf{es} & \textbf{et} & {\color[HTML]{FE0000} \textbf{eu}} & {\color[HTML]{FE0000} \textbf{fa}} & \textbf{fi} & \textbf{fr} & \textbf{he} & {\color[HTML]{FE0000} \textbf{hi}} & \textbf{hu} & \textbf{id} & \textbf{it} & \textbf{ja} \\ \midrule
+ Phase 1 & {\ul 86.7} & {\ul 84.9} & 87.8 & {\ul 80.0} & \textbf{96.6} & {\ul 89.8} & 91.9 & 90.3 & 81.5 & 84.3 & 88.2 & 88.3 & {\ul 85.2} & 84.9 & {\ul 89.1} & \textbf{93.7} & {\ul 88.7} & {\ul 86.2} \\ \specialrule{0em}{0.5pt}{0.5pt}
+ fixed GMM & 86.2 & 84.3 & {\ul 89.6} & 79.4 & 92.6 & 87.5 & {\ul 93.9} & {\ul 92.6} & {\ul 84.5} & {\ul 87.9} & {\ul 92.1} & {\ul 91.4} & 84.4 & {\ul 92.6} & 87.1 & 90.9 & 87.1 & 82.4 \\ \specialrule{0em}{0.5pt}{0.5pt}
\textsc{Emma}-X & \textbf{93.9} & \textbf{90.2} & \textbf{91.4} & \textbf{92.1} & {\ul 95.9} & \textbf{92.0} & \textbf{95.7} & \textbf{94.3} & \textbf{96.4} & \textbf{96.0} & \textbf{94.7} & \textbf{92.1} & \textbf{90.3} & \textbf{98.5} & \textbf{92.0} & {\ul 93.4} & \textbf{91.6} & \textbf{91.3} \\ \midrule
Model\textbackslash{}Language & {\color[HTML]{FE0000} \textbf{jv}} & {\color[HTML]{FE0000} \textbf{ka}} & {\color[HTML]{FE0000} \textbf{kk}} & \textbf{ko} & {\color[HTML]{FE0000} \textbf{ml}} & {\color[HTML]{FE0000} \textbf{mr}} & \textbf{nl} & \textbf{pt} & \textbf{ru} & {\color[HTML]{FE0000} \textbf{sw}} & {\color[HTML]{FE0000} \textbf{ta}} & {\color[HTML]{FE0000} \textbf{te}} & {\color[HTML]{FE0000} \textbf{th}} & {\color[HTML]{FE0000} \textbf{tl}} & \textbf{tr} & {\color[HTML]{FE0000} \textbf{ur}} & \textbf{vi} & \textbf{zh} \\ \midrule
+ Phase 1 & 22.6 & 88.3 & 61.6 & 85.0 & {\ul 93.9} & 82.0 & {\ul 94.7} & {\ul 92.1} & 88.7 & 70.5 & 70.9 & 81.5 & 84.9 & 59.2 & 88.2 & 85.8 & {\ul 97.1} & {\ul 91.7} \\ \specialrule{0em}{0.5pt}{0.5pt}
+ fixed GMM & {\ul 37.9} & {\ul 91.8} & {\ul 67.1} & {\ul 87.6} & 91.8 & {\ul 85.3} & 93.1 & 88.9 & {\ul 90.8} & {\ul 75.4} & {\ul 72.4} & {\ul 82.3} & {\ul 86.7} & {\ul 68.1} & {\ul 91.6} & {\ul 89.3} & 94.9 & 88.8 \\ \specialrule{0em}{0.5pt}{0.5pt}
\textsc{Emma}-X & \textbf{56.7} & \textbf{93.2} & \textbf{86.0} & \textbf{91.4} & \textbf{96.7} & \textbf{94.8} & \textbf{93.6} & \textbf{92.7} & \textbf{91.6} & \textbf{87.2} & \textbf{87.8} & \textbf{95.4} & \textbf{96.9} & \textbf{85.7} & \textbf{96.5} & \textbf{93.4} & \textbf{96.1} & \textbf{92.7} \\ \bottomrule
\end{tabular}%
}
 \vspace{5pt}
\caption{Ablation Study about \textsc{Emma}-X on 36 languages in Tatoeba. We mark the 14 long-tail languages with color red. We highlight the best results with bold font and the second-best results with underlines.}
\label{tab:ablation_study}
\end{table}

\subsection{Ablation Analysis}
The primary goal of \textsc{Emma}-X is to acquire universal semantic representations for a multitude of languages. For the limited number of parallel data,  \textsc{Emma}-X strives to leverage non-parallel data to extend language coverage for universal semantic representations and to enhance representation performance for those languages. To provide additional evidence supporting our claim, we propose an ablation experiment in Table \ref{tab:ablation_study}. The results in Table \ref{tab:ablation_study} demonstrate that \textsc{Emma}-X significantly improves the retrieval accuracy by 8.1 when compared to solely using parallel data. Moreover, on 16 long-tail languages, the retrieval accuracy further increases from 76.2 to 90.7, representing a substantial improvement of 19.0\%. These findings support our claim that  \textsc{Emma}-X can indeed learn effective universal semantic representations from an abundance of non-parallel data.

\subsection{Semantic Rank and GMM Analysis}
We assessed \textsc{Emma}-X's performance on \textsc{xrete} benchmark with various settings in Table \ref{tbl:rank-class-ab}. The evaluation focuses on two aspects: the classifier type and the value of semantic similarity rank.

\noindent\textbf{Type of the classifier.} Without direct supervision, we propose each semantic rank follow a Gaussian distribution. Comparing ``\textbf{GMM}'' with ``\textbf{FCN}'' in Table \ref{tbl:rank-class-ab}, the performance drops from $70.7$ to $68.1$. This decline is attributed to the handling of outliers: the ``\textbf{GMM}'' handles them better due to its soft assignment approach, giving outliers low probabilities and thereby minimizing their influence.

\noindent\textbf{Value of Semantic Similarity Rank.} Different semantic similarity ranks were evaluated in Table \ref{tbl:rank-class-ab}. The peak performance at $N=4$ implies that a refined similarity rank ensures a balanced distribution of semantic ranks, which helps alleviate data imbalance. However, increasing the number of semantic ranks complicates the learning process for both ``\textbf{GMM}'' and the cross-lingual encoder. $N=4$ strikes a balance between data distribution and learning complexity.

\begin{table*}[t]
\centering
\footnotesize
\resizebox{\textwidth}{!}{
\begin{tabular}{lcccccccccccc}
\toprule
\multirow{2}{*}{\textbf{Setting}} & XNLI & ANLI & MultiSTS & QE & LAReQA & Mewsli-X & BUCC & Tatoeba & XCOPA & MultiEURLEX & MultiARC & PAWS-X\\
~ & Acc. ($\color{green}{\uparrow}$) & Acc. ($\color{green}{\uparrow}$) & Spearman ($\color{green}{\uparrow}$) & Pearson ($\color{green}{\uparrow}$) & mAP@20 ($\color{green}{\uparrow}$) & mAP@20 ($\color{green}{\uparrow}$) & F1 ($\color{green}{\uparrow}$) & Acc. ($\color{green}{\uparrow}$) & Acc. ($\color{green}{\uparrow}$) & Acc. ($\color{green}{\uparrow}$) & MAE ($\color{red}{\downarrow}$) & Acc. ($\color{green}{\uparrow}$)\\
\midrule
\textbf{(FCN, $N=4$)} & 86.4 & 46.63 & 87.0 & 66.1 & 46.6 & 54.2 & 81.1 & 78.2 & 76.2 & 68.5 & 34.9 & 90.8 \\
\midrule
\textbf{(GMM, $N=2$)} & 85.2 & 49.52 & 84.7 & 67.3 & 47.8 & 56.6 & 85.6 & 82.3 & 77.0 & 70.1 & \bf 30.6 & 90.9 \\
\textbf{(GMM, $N=4$)}$^\ddag$ & \bf 88.1 & 50.21 & \bf 87.3 & 67.2 & \bf 50.6 & \bf 59.6 & \bf 87.1 & \bf 82.5 & 78.2 & \bf 71.4 & 32.7 & 94.2 \\
\textbf{(GMM, $N=8$)} & 87.9 & \bf 50.58 & 86.3 & \bf 67.4 & 49.2 & 59.5 & 85.7 & \bf 82.5 & \bf 78.8 & 71.3 & 32.5 & \bf 95.1 \\
\bottomrule
\end{tabular}%
}
\caption{Effect of different settings on the \textsc{xrete} benchmark. $^\ddag$ means results with default settings (i.e., GMM classifier and $N=4$). ``FCN'' means the semantic classifier is developed as fully connected networks.}\label{tbl:rank-class-ab}
\end{table*}

\section{Related Work}\label{sec:bg}

\vspace{-2pt}
\noindent\textbf{Cross-lingual Representation Pre-training} In recent years, various studies~\citep{devlin-etal-2019-bert,conneau-etal-2020-unsupervised} have shifted the monolingual pre-training procedure to multilingual scenarios. Most of them often rely heavily on parallel data to learn cross-lingual sentence representations, with several improved techniques, such as language modeling~\citep{NEURIPS2019_c04c19c2} and contrastive learning~\citep{chi-etal-2021-infoxlm, hu-etal-2021-explicit,wei2021on}. Recently, some endeavors incorporate monolingual data into parallel corpora, by translating monolingual corpora into pseudo parallel corpora \citep{kvapilikova-etal-2020-unsupervised, ouyang-etal-2021-ernie}, or computing the semantic similarity in monolingual corpora off-the-shelf and using it as supervision signals \citep{goswami-etal-2021-cross}. 

\noindent\textbf{Contrastive Learning} has become a popular paradigm in NLP. Besides constructing the positives and negatives through parallel corpora \citep{zhang-etal-2021-pairwise} or other labeled data \citep{gunel2021supervised, ni-etal-2022-sentence}, researchers also adopt self-supervised methods, which build positives and negatives by corrupting sentences \citep{gao-etal-2021-simcse,chuang-etal-2022-diffcse} or data augmentation methods \citep{zhang-etal-2022-virtual,https://doi.org/10.48550/arxiv.2210.06432}. Another line of research improves the quality of negatives by preserving a memory queue \citep{yang-etal-2021-xmoco, wang-etal-2021-aligning} or generating high-quality negatives \citep{Zhang_Zhang_Mensah_Liu_Mao_2022, wei-etal-2022-learning}. 


\section{Conclusion}
In this paper, we study the problem of learning cross-lingual universal representations. Major contributions are fourfold: 1) We propose a novel paradigm \textsc{Emma}-X, which can make full use of massive monolingual data to learn universities for any two languages. 2) We summarize three requirements for the universal representation space among all languages and verify the superiority of \textsc{Emma}-X towards strong baselines. 3) To incentivize the research of cross-lingual universal representation learning, we form a novel benchmark (\textsc{xrete}) with 12 cross-lingual tasks fully depending on sentence-level representations. 4) Experiments on \textsc{xrete} demonstrate that \textsc{Emma}-X achieved state-of-the-art results over strong baselines.

\section*{Acknowledgements}
We thank the anonymous reviewers for their constructive comments. This work is supported by the National Natural Science Foundation of China (Grant No.U21B2009). This research is also supported by the Strategic Priority Research Program of the Chinese Academy of Science, Grant No.XDC02030400.

\bibliographystyle{plainnat}
\bibliography{custom}

\begin{thebibliography}{87}
\providecommand{\natexlab}[1]{#1}
\providecommand{\url}[1]{\texttt{#1}}
\expandafter\ifx\csname urlstyle\endcsname\relax
  \providecommand{\doi}[1]{doi: #1}\else
  \providecommand{\doi}{doi: \begingroup \urlstyle{rm}\Url}\fi

\bibitem[Abend and Rappoport(2017)]{abend-rappoport-2017-state}
Omri Abend and Ari Rappoport.
\newblock The state of the art in semantic representation.
\newblock In \emph{Proceedings of the 55th Annual Meeting of the Association
  for Computational Linguistics (Volume 1: Long Papers)}, pages 77--89,
  Vancouver, Canada, July 2017. Association for Computational Linguistics.
\newblock \doi{10.18653/v1/P17-1008}.
\newblock URL \url{https://aclanthology.org/P17-1008}.

\bibitem[Aharoni et~al.(2019)Aharoni, Johnson, and
  Firat]{aharoni-etal-2019-massively}
Roee Aharoni, Melvin Johnson, and Orhan Firat.
\newblock Massively multilingual neural machine translation.
\newblock In \emph{Proceedings of the 2019 Conference of the North {A}merican
  Chapter of the Association for Computational Linguistics: Human Language
  Technologies, Volume 1 (Long and Short Papers)}, pages 3874--3884,
  Minneapolis, Minnesota, June 2019. Association for Computational Linguistics.
\newblock \doi{10.18653/v1/N19-1388}.
\newblock URL \url{https://aclanthology.org/N19-1388}.

\bibitem[Akhbardeh et~al.(2021)Akhbardeh, Arkhangorodsky, Biesialska, Bojar,
  Chatterjee, Chaudhary, Costa-jussa, Espa{\~n}a-Bonet, Fan, Federmann,
  Freitag, Graham, Grundkiewicz, Haddow, Harter, Heafield, Homan, Huck,
  Amponsah-Kaakyire, Kasai, Khashabi, Knight, Kocmi, Koehn, Lourie, Monz,
  Morishita, Nagata, Nagesh, Nakazawa, Negri, Pal, Tapo, Turchi, Vydrin, and
  Zampieri]{akhbardeh-etal-2021-findings}
Farhad Akhbardeh, Arkady Arkhangorodsky, Magdalena Biesialska, Ond{\v{r}}ej
  Bojar, Rajen Chatterjee, Vishrav Chaudhary, Marta~R. Costa-jussa, Cristina
  Espa{\~n}a-Bonet, Angela Fan, Christian Federmann, Markus Freitag, Yvette
  Graham, Roman Grundkiewicz, Barry Haddow, Leonie Harter, Kenneth Heafield,
  Christopher Homan, Matthias Huck, Kwabena Amponsah-Kaakyire, Jungo Kasai,
  Daniel Khashabi, Kevin Knight, Tom Kocmi, Philipp Koehn, Nicholas Lourie,
  Christof Monz, Makoto Morishita, Masaaki Nagata, Ajay Nagesh, Toshiaki
  Nakazawa, Matteo Negri, Santanu Pal, Allahsera~Auguste Tapo, Marco Turchi,
  Valentin Vydrin, and Marcos Zampieri.
\newblock Findings of the 2021 conference on machine translation ({WMT}21).
\newblock In \emph{Proceedings of the Sixth Conference on Machine Translation},
  pages 1--88, Online, November 2021. Association for Computational
  Linguistics.
\newblock URL \url{https://aclanthology.org/2021.wmt-1.1}.

\bibitem[Artetxe and Schwenk(2019)]{artetxe-schwenk-2019-massively}
Mikel Artetxe and Holger Schwenk.
\newblock Massively multilingual sentence embeddings for zero-shot
  cross-lingual transfer and beyond.
\newblock \emph{Transactions of the Association for Computational Linguistics},
  7:\penalty0 597--610, 2019.
\newblock \doi{10.1162/tacl_a_00288}.
\newblock URL \url{https://aclanthology.org/Q19-1038}.

\bibitem[Artetxe et~al.(2020)Artetxe, Ruder, and
  Yogatama]{artetxe-etal-2020-cross}
Mikel Artetxe, Sebastian Ruder, and Dani Yogatama.
\newblock On the cross-lingual transferability of monolingual representations.
\newblock In \emph{Proceedings of the 58th Annual Meeting of the Association
  for Computational Linguistics}, pages 4623--4637, Online, July 2020.
  Association for Computational Linguistics.
\newblock \doi{10.18653/v1/2020.acl-main.421}.
\newblock URL \url{https://aclanthology.org/2020.acl-main.421}.

\bibitem[Bapna et~al.(2022)Bapna, Caswell, Kreutzer, Firat, van Esch, Siddhant,
  Niu, Baljekar, Garcia, Macherey, Breiner, Axelrod, Riesa, Cao, Chen,
  Macherey, Krikun, Wang, Gutkin, Shah, Huang, Chen, Wu, and Hughes]{51503}
Ankur Bapna, Isaac Caswell, Julia Kreutzer, Orhan Firat, Daan van Esch, Aditya
  Siddhant, Mengmeng Niu, Pallavi~Nikhil Baljekar, Xavier Garcia, Wolfgang
  Macherey, Theresa Breiner, Vera~Saldinger Axelrod, Jason Riesa, Yuan Cao, Mia
  Chen, Klaus Macherey, Maxim Krikun, Pidong Wang, Alexander Gutkin, Apu Shah,
  Yanping Huang, Zhifeng Chen, Yonghui Wu, and Macduff~Richard Hughes.
\newblock Building machine translation systems for the next thousand languages.
\newblock Technical report, Google Research, 2022.

\bibitem[Botha et~al.(2020)Botha, Shan, and Gillick]{botha-etal-2020-entity}
Jan~A. Botha, Zifei Shan, and Daniel Gillick.
\newblock {E}ntity {L}inking in 100 {L}anguages.
\newblock In \emph{Proceedings of the 2020 Conference on Empirical Methods in
  Natural Language Processing (EMNLP)}, pages 7833--7845, Online, November
  2020. Association for Computational Linguistics.
\newblock \doi{10.18653/v1/2020.emnlp-main.630}.
\newblock URL \url{https://aclanthology.org/2020.emnlp-main.630}.

\bibitem[Bowman et~al.(2015)Bowman, Angeli, Potts, and
  Manning]{bowman-etal-2015-large}
Samuel~R. Bowman, Gabor Angeli, Christopher Potts, and Christopher~D. Manning.
\newblock A large annotated corpus for learning natural language inference.
\newblock In \emph{Proceedings of the 2015 Conference on Empirical Methods in
  Natural Language Processing}, pages 632--642, Lisbon, Portugal, September
  2015. Association for Computational Linguistics.
\newblock \doi{10.18653/v1/D15-1075}.
\newblock URL \url{https://www.aclweb.org/anthology/D15-1075}.

\bibitem[Caliński and Harabasz(1974)]{doi:10.1080/03610927408827101}
T.~Caliński and J~Harabasz.
\newblock A dendrite method for cluster analysis.
\newblock \emph{Communications in Statistics}, 3\penalty0 (1):\penalty0 1--27,
  1974.
\newblock \doi{10.1080/03610927408827101}.
\newblock URL
  \url{https://www.tandfonline.com/doi/abs/10.1080/03610927408827101}.

\bibitem[Cer et~al.(2017)Cer, Diab, Agirre, Lopez-Gazpio, and
  Specia]{cer-etal-2017-semeval}
Daniel Cer, Mona Diab, Eneko Agirre, I{\~n}igo Lopez-Gazpio, and Lucia Specia.
\newblock {S}em{E}val-2017 task 1: Semantic textual similarity multilingual and
  crosslingual focused evaluation.
\newblock In \emph{Proceedings of the 11th International Workshop on Semantic
  Evaluation ({S}em{E}val-2017)}, pages 1--14, Vancouver, Canada, August 2017.
  Association for Computational Linguistics.
\newblock \doi{10.18653/v1/S17-2001}.
\newblock URL \url{https://aclanthology.org/S17-2001}.

\bibitem[Chalkidis et~al.(2021)Chalkidis, Fergadiotis, and
  Androutsopoulos]{chalkidis-etal-2021-multieurlex}
Ilias Chalkidis, Manos Fergadiotis, and Ion Androutsopoulos.
\newblock {M}ulti{EURLEX} - a multi-lingual and multi-label legal document
  classification dataset for zero-shot cross-lingual transfer.
\newblock In \emph{Proceedings of the 2021 Conference on Empirical Methods in
  Natural Language Processing}, pages 6974--6996, Online and Punta Cana,
  Dominican Republic, November 2021. Association for Computational Linguistics.
\newblock \doi{10.18653/v1/2021.emnlp-main.559}.
\newblock URL \url{https://aclanthology.org/2021.emnlp-main.559}.

\bibitem[Chen et~al.(2023)Chen, Shou, Gong, Pei, Cao, Chang, Jiang, and
  Li]{chen2023alleviating}
Nuo Chen, Linjun Shou, Ming Gong, Jian Pei, Bowen Cao, Jianhui Chang, Daxin
  Jiang, and Jia Li.
\newblock Alleviating over-smoothing for unsupervised sentence representation,
  2023.

\bibitem[Chi et~al.(2021)Chi, Dong, Wei, Yang, Singhal, Wang, Song, Mao, Huang,
  and Zhou]{chi-etal-2021-infoxlm}
Zewen Chi, Li~Dong, Furu Wei, Nan Yang, Saksham Singhal, Wenhui Wang, Xia Song,
  Xian-Ling Mao, Heyan Huang, and Ming Zhou.
\newblock {I}nfo{XLM}: An information-theoretic framework for cross-lingual
  language model pre-training.
\newblock In \emph{Proceedings of the 2021 Conference of the North American
  Chapter of the Association for Computational Linguistics: Human Language
  Technologies}, pages 3576--3588, Online, June 2021. Association for
  Computational Linguistics.
\newblock \doi{10.18653/v1/2021.naacl-main.280}.
\newblock URL \url{https://aclanthology.org/2021.naacl-main.280}.

\bibitem[Chowdhery et~al.(2022)Chowdhery, Narang, Devlin, Bosma, Mishra,
  Roberts, Barham, Chung, Sutton, Gehrmann, Schuh, Shi, Tsvyashchenko, Maynez,
  Rao, Barnes, Tay, Shazeer, Prabhakaran, Reif, Du, Hutchinson, Pope, Bradbury,
  Austin, Isard, Gur-Ari, Yin, Duke, Levskaya, Ghemawat, Dev, Michalewski,
  Garcia, Misra, Robinson, Fedus, Zhou, Ippolito, Luan, Lim, Zoph, Spiridonov,
  Sepassi, Dohan, Agrawal, Omernick, Dai, Pillai, Pellat, Lewkowycz, Moreira,
  Child, Polozov, Lee, Zhou, Wang, Saeta, Diaz, Firat, Catasta, Wei,
  Meier-Hellstern, Eck, Dean, Petrov, and Fiedel]{chowdhery2022palm}
Aakanksha Chowdhery, Sharan Narang, Jacob Devlin, Maarten Bosma, Gaurav Mishra,
  Adam Roberts, Paul Barham, Hyung~Won Chung, Charles Sutton, Sebastian
  Gehrmann, Parker Schuh, Kensen Shi, Sasha Tsvyashchenko, Joshua Maynez,
  Abhishek Rao, Parker Barnes, Yi~Tay, Noam Shazeer, Vinodkumar Prabhakaran,
  Emily Reif, Nan Du, Ben Hutchinson, Reiner Pope, James Bradbury, Jacob
  Austin, Michael Isard, Guy Gur-Ari, Pengcheng Yin, Toju Duke, Anselm
  Levskaya, Sanjay Ghemawat, Sunipa Dev, Henryk Michalewski, Xavier Garcia,
  Vedant Misra, Kevin Robinson, Liam Fedus, Denny Zhou, Daphne Ippolito, David
  Luan, Hyeontaek Lim, Barret Zoph, Alexander Spiridonov, Ryan Sepassi, David
  Dohan, Shivani Agrawal, Mark Omernick, Andrew~M. Dai,
  Thanumalayan~Sankaranarayana Pillai, Marie Pellat, Aitor Lewkowycz, Erica
  Moreira, Rewon Child, Oleksandr Polozov, Katherine Lee, Zongwei Zhou, Xuezhi
  Wang, Brennan Saeta, Mark Diaz, Orhan Firat, Michele Catasta, Jason Wei,
  Kathy Meier-Hellstern, Douglas Eck, Jeff Dean, Slav Petrov, and Noah Fiedel.
\newblock Palm: Scaling language modeling with pathways, 2022.

\bibitem[Christiano et~al.(2017)Christiano, Leike, Brown, Martic, Legg, and
  Amodei]{NIPS2017_d5e2c0ad}
Paul~F Christiano, Jan Leike, Tom Brown, Miljan Martic, Shane Legg, and Dario
  Amodei.
\newblock Deep reinforcement learning from human preferences.
\newblock In I.~Guyon, U.~Von Luxburg, S.~Bengio, H.~Wallach, R.~Fergus,
  S.~Vishwanathan, and R.~Garnett, editors, \emph{Advances in Neural
  Information Processing Systems}, volume~30. Curran Associates, Inc., 2017.
\newblock URL
  \url{https://proceedings.neurips.cc/paper_files/paper/2017/file/d5e2c0adad503c91f91df240d0cd4e49-Paper.pdf}.

\bibitem[Chuang et~al.(2022)Chuang, Dangovski, Luo, Zhang, Chang, Soljacic, Li,
  Yih, Kim, and Glass]{chuang-etal-2022-diffcse}
Yung-Sung Chuang, Rumen Dangovski, Hongyin Luo, Yang Zhang, Shiyu Chang, Marin
  Soljacic, Shang-Wen Li, Scott Yih, Yoon Kim, and James Glass.
\newblock {D}iff{CSE}: Difference-based contrastive learning for sentence
  embeddings.
\newblock In \emph{Proceedings of the 2022 Conference of the North American
  Chapter of the Association for Computational Linguistics: Human Language
  Technologies}, pages 4207--4218, Seattle, United States, July 2022.
  Association for Computational Linguistics.
\newblock \doi{10.18653/v1/2022.naacl-main.311}.
\newblock URL \url{https://aclanthology.org/2022.naacl-main.311}.

\bibitem[Conneau and Lample(2019)]{NEURIPS2019_c04c19c2}
Alexis Conneau and Guillaume Lample.
\newblock Cross-lingual language model pretraining.
\newblock In H.~Wallach, H.~Larochelle, A.~Beygelzimer, F.~d\textquotesingle
  Alch\'{e}-Buc, E.~Fox, and R.~Garnett, editors, \emph{Advances in Neural
  Information Processing Systems}, volume~32. Curran Associates, Inc., 2019.
\newblock URL
  \url{https://proceedings.neurips.cc/paper/2019/file/c04c19c2c2474dbf5f7ac4372c5b9af1-Paper.pdf}.

\bibitem[Conneau et~al.(2018)Conneau, Rinott, Lample, Williams, Bowman,
  Schwenk, and Stoyanov]{conneau-etal-2018-xnli}
Alexis Conneau, Ruty Rinott, Guillaume Lample, Adina Williams, Samuel Bowman,
  Holger Schwenk, and Veselin Stoyanov.
\newblock {XNLI}: Evaluating cross-lingual sentence representations.
\newblock In \emph{Proceedings of the 2018 Conference on Empirical Methods in
  Natural Language Processing}, pages 2475--2485, Brussels, Belgium,
  October-November 2018. Association for Computational Linguistics.
\newblock \doi{10.18653/v1/D18-1269}.
\newblock URL \url{https://www.aclweb.org/anthology/D18-1269}.

\bibitem[Conneau et~al.(2020)Conneau, Khandelwal, Goyal, Chaudhary, Wenzek,
  Guzm{\'a}n, Grave, Ott, Zettlemoyer, and
  Stoyanov]{conneau-etal-2020-unsupervised}
Alexis Conneau, Kartikay Khandelwal, Naman Goyal, Vishrav Chaudhary, Guillaume
  Wenzek, Francisco Guzm{\'a}n, Edouard Grave, Myle Ott, Luke Zettlemoyer, and
  Veselin Stoyanov.
\newblock Unsupervised cross-lingual representation learning at scale.
\newblock In \emph{Proceedings of the 58th Annual Meeting of the Association
  for Computational Linguistics}, pages 8440--8451, Online, July 2020.
  Association for Computational Linguistics.
\newblock \doi{10.18653/v1/2020.acl-main.747}.
\newblock URL \url{https://aclanthology.org/2020.acl-main.747}.

\bibitem[Costa-juss{\`a} et~al.(2022)Costa-juss{\`a}, Cross, {\c{C}}elebi,
  Elbayad, Heafield, Heffernan, Kalbassi, Lam, Licht, Maillard,
  et~al.]{costa2022no}
Marta~R Costa-juss{\`a}, James Cross, Onur {\c{C}}elebi, Maha Elbayad, Kenneth
  Heafield, Kevin Heffernan, Elahe Kalbassi, Janice Lam, Daniel Licht, Jean
  Maillard, et~al.
\newblock No language left behind: Scaling human-centered machine translation.
\newblock \emph{arXiv preprint arXiv:2207.04672}, 2022.

\bibitem[Devlin et~al.(2019)Devlin, Chang, Lee, and
  Toutanova]{devlin-etal-2019-bert}
Jacob Devlin, Ming-Wei Chang, Kenton Lee, and Kristina Toutanova.
\newblock {BERT}: Pre-training of deep bidirectional transformers for language
  understanding.
\newblock In \emph{Proceedings of the 2019 Conference of the North {A}merican
  Chapter of the Association for Computational Linguistics: Human Language
  Technologies, Volume 1 (Long and Short Papers)}, pages 4171--4186,
  Minneapolis, Minnesota, June 2019. Association for Computational Linguistics.
\newblock \doi{10.18653/v1/N19-1423}.
\newblock URL \url{https://aclanthology.org/N19-1423}.

\bibitem[Ebrahimi et~al.(2022)Ebrahimi, Mager, Oncevay, Chaudhary, Chiruzzo,
  Fan, Ortega, Ramos, Rios, Meza~Ruiz, Gim{\'e}nez-Lugo, Mager, Neubig, Palmer,
  Coto-Solano, Vu, and Kann]{ebrahimi-etal-2022-americasnli}
Abteen Ebrahimi, Manuel Mager, Arturo Oncevay, Vishrav Chaudhary, Luis
  Chiruzzo, Angela Fan, John Ortega, Ricardo Ramos, Annette Rios, Ivan~Vladimir
  Meza~Ruiz, Gustavo Gim{\'e}nez-Lugo, Elisabeth Mager, Graham Neubig, Alexis
  Palmer, Rolando Coto-Solano, Thang Vu, and Katharina Kann.
\newblock {A}mericas{NLI}: Evaluating zero-shot natural language understanding
  of pretrained multilingual models in truly low-resource languages.
\newblock In \emph{Proceedings of the 60th Annual Meeting of the Association
  for Computational Linguistics (Volume 1: Long Papers)}, pages 6279--6299,
  Dublin, Ireland, May 2022. Association for Computational Linguistics.
\newblock \doi{10.18653/v1/2022.acl-long.435}.
\newblock URL \url{https://aclanthology.org/2022.acl-long.435}.

\bibitem[El-Kishky et~al.(2020)El-Kishky, Chaudhary, Guzm{\'a}n, and
  Koehn]{el-kishky-etal-2020-ccaligned}
Ahmed El-Kishky, Vishrav Chaudhary, Francisco Guzm{\'a}n, and Philipp Koehn.
\newblock {CCA}ligned: A massive collection of cross-lingual web-document
  pairs.
\newblock In \emph{Proceedings of the 2020 Conference on Empirical Methods in
  Natural Language Processing (EMNLP)}, pages 5960--5969, Online, November
  2020. Association for Computational Linguistics.
\newblock \doi{10.18653/v1/2020.emnlp-main.480}.
\newblock URL \url{https://aclanthology.org/2020.emnlp-main.480}.

\bibitem[Everitt and Hand(1981)]{everitt1981finite}
BS~Everitt and DJ~Hand.
\newblock Finite mixture distributions.
\newblock \emph{Monographs on Applied Probability and Statistics}, 1981.

\bibitem[Feng et~al.(2022)Feng, Yang, Cer, Arivazhagan, and
  Wang]{feng-etal-2022-language}
Fangxiaoyu Feng, Yinfei Yang, Daniel Cer, Naveen Arivazhagan, and Wei Wang.
\newblock Language-agnostic {BERT} sentence embedding.
\newblock In \emph{Proceedings of the 60th Annual Meeting of the Association
  for Computational Linguistics (Volume 1: Long Papers)}, pages 878--891,
  Dublin, Ireland, May 2022. Association for Computational Linguistics.
\newblock \doi{10.18653/v1/2022.acl-long.62}.
\newblock URL \url{https://aclanthology.org/2022.acl-long.62}.

\bibitem[Gao et~al.(2021)Gao, Yao, and Chen]{gao-etal-2021-simcse}
Tianyu Gao, Xingcheng Yao, and Danqi Chen.
\newblock {S}im{CSE}: Simple contrastive learning of sentence embeddings.
\newblock In \emph{Proceedings of the 2021 Conference on Empirical Methods in
  Natural Language Processing}, pages 6894--6910, Online and Punta Cana,
  Dominican Republic, November 2021. Association for Computational Linguistics.
\newblock \doi{10.18653/v1/2021.emnlp-main.552}.
\newblock URL \url{https://aclanthology.org/2021.emnlp-main.552}.

\bibitem[Gordon et~al.(2012)Gordon, Kozareva, and
  Roemmele]{gordon-etal-2012-semeval}
Andrew Gordon, Zornitsa Kozareva, and Melissa Roemmele.
\newblock {S}em{E}val-2012 task 7: Choice of plausible alternatives: An
  evaluation of commonsense causal reasoning.
\newblock In \emph{*{SEM} 2012: The First Joint Conference on Lexical and
  Computational Semantics {--} Volume 1: Proceedings of the main conference and
  the shared task, and Volume 2: Proceedings of the Sixth International
  Workshop on Semantic Evaluation ({S}em{E}val 2012)}, pages 394--398,
  Montr{\'e}al, Canada, 7-8 June 2012. Association for Computational
  Linguistics.
\newblock URL \url{https://aclanthology.org/S12-1052}.

\bibitem[Goswami et~al.(2021)Goswami, Dutta, Assem, Fransen, and
  McCrae]{goswami-etal-2021-cross}
Koustava Goswami, Sourav Dutta, Haytham Assem, Theodorus Fransen, and John~P.
  McCrae.
\newblock Cross-lingual sentence embedding using multi-task learning.
\newblock In \emph{Proceedings of the 2021 Conference on Empirical Methods in
  Natural Language Processing}, pages 9099--9113, Online and Punta Cana,
  Dominican Republic, November 2021. Association for Computational Linguistics.
\newblock \doi{10.18653/v1/2021.emnlp-main.716}.
\newblock URL \url{https://aclanthology.org/2021.emnlp-main.716}.

\bibitem[Goyal et~al.(2022)Goyal, Gao, Chaudhary, Chen, Wenzek, Ju, Krishnan,
  Ranzato, Guzm{\'a}n, and Fan]{goyal-etal-2022-flores}
Naman Goyal, Cynthia Gao, Vishrav Chaudhary, Peng-Jen Chen, Guillaume Wenzek,
  Da~Ju, Sanjana Krishnan, Marc{'}Aurelio Ranzato, Francisco Guzm{\'a}n, and
  Angela Fan.
\newblock The {F}lores-101 evaluation benchmark for low-resource and
  multilingual machine translation.
\newblock \emph{Transactions of the Association for Computational Linguistics},
  10:\penalty0 522--538, 2022.
\newblock \doi{10.1162/tacl_a_00474}.
\newblock URL \url{https://aclanthology.org/2022.tacl-1.30}.

\bibitem[Gunel et~al.(2021)Gunel, Du, Conneau, and
  Stoyanov]{gunel2021supervised}
Beliz Gunel, Jingfei Du, Alexis Conneau, and Veselin Stoyanov.
\newblock Supervised contrastive learning for pre-trained language model
  fine-tuning.
\newblock In \emph{International Conference on Learning Representations}, 2021.
\newblock URL \url{https://openreview.net/forum?id=cu7IUiOhujH}.

\bibitem[Guo et~al.(2018)Guo, Shen, Yang, Ge, Cer, Hernandez~Abrego, Stevens,
  Constant, Sung, Strope, and Kurzweil]{guo-etal-2018-effective}
Mandy Guo, Qinlan Shen, Yinfei Yang, Heming Ge, Daniel Cer, Gustavo
  Hernandez~Abrego, Keith Stevens, Noah Constant, Yun-Hsuan Sung, Brian Strope,
  and Ray Kurzweil.
\newblock Effective parallel corpus mining using bilingual sentence embeddings.
\newblock In \emph{Proceedings of the Third Conference on Machine Translation:
  Research Papers}, pages 165--176, Brussels, Belgium, October 2018.
  Association for Computational Linguistics.
\newblock \doi{10.18653/v1/W18-6317}.
\newblock URL \url{https://aclanthology.org/W18-6317}.

\bibitem[He et~al.(2020)He, Fan, Wu, Xie, and Girshick]{he2020momentum}
Kaiming He, Haoqi Fan, Yuxin Wu, Saining Xie, and Ross Girshick.
\newblock Momentum contrast for unsupervised visual representation learning.
\newblock In \emph{2020 IEEE/CVF Conference on Computer Vision and Pattern
  Recognition (CVPR)}, pages 9726--9735. IEEE Computer Society, 2020.

\bibitem[Hoffmann et~al.(2022)Hoffmann, Behrmann, Gall, Brox, and
  Noroozi]{hoffmann2022ranking}
David~T Hoffmann, N~Behrmann, J~Gall, Thomas Brox, and M~Noroozi.
\newblock Ranking info noise contrastive estimation: Boosting contrastive
  learning via ranked positives.
\newblock In \emph{AAAI Conference on Artificial Intelligence}, 2022.

\bibitem[Hu et~al.(2020)Hu, Ruder, Siddhant, Neubig, Firat, and
  Johnson]{pmlr-v119-hu20b}
Junjie Hu, Sebastian Ruder, Aditya Siddhant, Graham Neubig, Orhan Firat, and
  Melvin Johnson.
\newblock {XTREME}: A massively multilingual multi-task benchmark for
  evaluating cross-lingual generalisation.
\newblock In Hal~Daumé III and Aarti Singh, editors, \emph{Proceedings of the
  37th International Conference on Machine Learning}, volume 119 of
  \emph{Proceedings of Machine Learning Research}, pages 4411--4421. PMLR,
  13--18 Jul 2020.
\newblock URL \url{https://proceedings.mlr.press/v119/hu20b.html}.

\bibitem[Hu et~al.(2021)Hu, Johnson, Firat, Siddhant, and
  Neubig]{hu-etal-2021-explicit}
Junjie Hu, Melvin Johnson, Orhan Firat, Aditya Siddhant, and Graham Neubig.
\newblock Explicit alignment objectives for multilingual bidirectional
  encoders.
\newblock In \emph{Proceedings of the 2021 Conference of the North American
  Chapter of the Association for Computational Linguistics: Human Language
  Technologies}, pages 3633--3643, Online, June 2021. Association for
  Computational Linguistics.
\newblock \doi{10.18653/v1/2021.naacl-main.284}.
\newblock URL \url{https://aclanthology.org/2021.naacl-main.284}.

\bibitem[Huang et~al.(2019)Huang, Liang, Duan, Gong, Shou, Jiang, and
  Zhou]{huang-etal-2019-unicoder}
Haoyang Huang, Yaobo Liang, Nan Duan, Ming Gong, Linjun Shou, Daxin Jiang, and
  Ming Zhou.
\newblock {U}nicoder: A universal language encoder by pre-training with
  multiple cross-lingual tasks.
\newblock In \emph{Proceedings of the 2019 Conference on Empirical Methods in
  Natural Language Processing and the 9th International Joint Conference on
  Natural Language Processing (EMNLP-IJCNLP)}, pages 2485--2494, Hong Kong,
  China, November 2019. Association for Computational Linguistics.
\newblock \doi{10.18653/v1/D19-1252}.
\newblock URL \url{https://aclanthology.org/D19-1252}.

\bibitem[Irwin et~al.(2009)Irwin, Harkema, Christensen, Schleyer, Haug, and
  Chapman]{irwin2009methodology}
Jeannie~Y Irwin, Henk Harkema, Lee~M Christensen, Titus Schleyer, Peter~J Haug,
  and Wendy~W Chapman.
\newblock Methodology to develop and evaluate a semantic representation for
  nlp.
\newblock In \emph{AMIA Annual Symposium Proceedings}, volume 2009, page 271.
  American Medical Informatics Association, 2009.

\bibitem[Kepler et~al.(2019)Kepler, Tr{\'e}nous, Treviso, Vera, and
  Martins]{kepler-etal-2019-openkiwi}
Fabio Kepler, Jonay Tr{\'e}nous, Marcos Treviso, Miguel Vera, and Andr{\'e}
  F.~T. Martins.
\newblock {O}pen{K}iwi: An open source framework for quality estimation.
\newblock In \emph{Proceedings of the 57th Annual Meeting of the Association
  for Computational Linguistics: System Demonstrations}, pages 117--122,
  Florence, Italy, July 2019. Association for Computational Linguistics.
\newblock \doi{10.18653/v1/P19-3020}.
\newblock URL \url{https://aclanthology.org/P19-3020}.

\bibitem[Keung et~al.(2020)Keung, Lu, Szarvas, and
  Smith]{keung-etal-2020-multilingual}
Phillip Keung, Yichao Lu, Gy{\"o}rgy Szarvas, and Noah~A. Smith.
\newblock The multilingual {A}mazon reviews corpus.
\newblock In \emph{Proceedings of the 2020 Conference on Empirical Methods in
  Natural Language Processing (EMNLP)}, pages 4563--4568, Online, November
  2020. Association for Computational Linguistics.
\newblock \doi{10.18653/v1/2020.emnlp-main.369}.
\newblock URL \url{https://aclanthology.org/2020.emnlp-main.369}.

\bibitem[Kingma and Ba(2015)]{DBLP:journals/corr/KingmaB14}
Diederik~P. Kingma and Jimmy Ba.
\newblock Adam: A method for stochastic optimization.
\newblock In \emph{3rd International Conference on Learning Representations,
  San Diego, CA}, 2015.
\newblock URL \url{http://arxiv.org/abs/1412.6980}.

\bibitem[Kudo and Richardson(2018)]{kudo-richardson-2018-sentencepiece}
Taku Kudo and John Richardson.
\newblock {S}entence{P}iece: A simple and language independent subword
  tokenizer and detokenizer for neural text processing.
\newblock In \emph{Proceedings of the 2018 Conference on Empirical Methods in
  Natural Language Processing: System Demonstrations}, pages 66--71, Brussels,
  Belgium, November 2018. Association for Computational Linguistics.
\newblock \doi{10.18653/v1/D18-2012}.
\newblock URL \url{https://aclanthology.org/D18-2012}.

\bibitem[Kullback and Leibler(1951)]{10.1214/aoms/1177729694}
S.~Kullback and R.~A. Leibler.
\newblock {On Information and Sufficiency}.
\newblock \emph{The Annals of Mathematical Statistics}, 22\penalty0
  (1):\penalty0 79 -- 86, 1951.
\newblock \doi{10.1214/aoms/1177729694}.
\newblock URL \url{https://doi.org/10.1214/aoms/1177729694}.

\bibitem[Kvapil{\'\i}kov{\'a} et~al.(2020)Kvapil{\'\i}kov{\'a}, Artetxe,
  Labaka, Agirre, and Bojar]{kvapilikova-etal-2020-unsupervised}
Ivana Kvapil{\'\i}kov{\'a}, Mikel Artetxe, Gorka Labaka, Eneko Agirre, and
  Ond{\v{r}}ej Bojar.
\newblock Unsupervised multilingual sentence embeddings for parallel corpus
  mining.
\newblock In \emph{Proceedings of the 58th Annual Meeting of the Association
  for Computational Linguistics: Student Research Workshop}, pages 255--262,
  Online, July 2020. Association for Computational Linguistics.
\newblock \doi{10.18653/v1/2020.acl-srw.34}.
\newblock URL \url{https://aclanthology.org/2020.acl-srw.34}.

\bibitem[Lee et~al.(2022)Lee, Lee, Jang, and Yu]{lee-etal-2022-toward}
Seonghyeon Lee, Dongha Lee, Seongbo Jang, and Hwanjo Yu.
\newblock Toward interpretable semantic textual similarity via optimal
  transport-based contrastive sentence learning.
\newblock In \emph{Proceedings of the 60th Annual Meeting of the Association
  for Computational Linguistics (Volume 1: Long Papers)}, pages 5969--5979,
  Dublin, Ireland, May 2022. Association for Computational Linguistics.
\newblock \doi{10.18653/v1/2022.acl-long.412}.
\newblock URL \url{https://aclanthology.org/2022.acl-long.412}.

\bibitem[Leibniz and von Leibniz(1996)]{leibniz1996leibniz}
Gottfried~Wilhelm Leibniz and Gottfried Wilhelm~Freiherr von Leibniz.
\newblock \emph{Leibniz: New essays on human understanding}.
\newblock Cambridge University Press, 1996.

\bibitem[Li et~al.(2023)Li, Huang, Zhang, Deng, Lou, Huang, Jiao, Wei, Deng,
  and Zhang]{li2023dualalignment}
Ziheng Li, Shaohan Huang, Zihan Zhang, Zhi-Hong Deng, Qiang Lou, Haizhen Huang,
  Jian Jiao, Furu Wei, Weiwei Deng, and Qi~Zhang.
\newblock Dual-alignment pre-training for cross-lingual sentence embedding,
  2023.

\bibitem[Luo et~al.(2021)Luo, Wang, Liu, Liu, Bi, Huang, Huang, and
  Si]{luo-etal-2021-veco}
Fuli Luo, Wei Wang, Jiahao Liu, Yijia Liu, Bin Bi, Songfang Huang, Fei Huang,
  and Luo Si.
\newblock {VECO}: Variable and flexible cross-lingual pre-training for language
  understanding and generation.
\newblock In \emph{Proceedings of the 59th Annual Meeting of the Association
  for Computational Linguistics and the 11th International Joint Conference on
  Natural Language Processing (Volume 1: Long Papers)}, pages 3980--3994,
  Online, August 2021. Association for Computational Linguistics.
\newblock \doi{10.18653/v1/2021.acl-long.308}.
\newblock URL \url{https://aclanthology.org/2021.acl-long.308}.

\bibitem[Moon(1996)]{543975}
T.K. Moon.
\newblock The expectation-maximization algorithm.
\newblock \emph{IEEE Signal Processing Magazine}, 13\penalty0 (6):\penalty0
  47--60, 1996.
\newblock \doi{10.1109/79.543975}.

\bibitem[Mu and Viswanath(2018)]{mu2018allbutthetop}
Jiaqi Mu and Pramod Viswanath.
\newblock All-but-the-top: Simple and effective postprocessing for word
  representations.
\newblock In \emph{International Conference on Learning Representations}, 2018.
\newblock URL \url{https://openreview.net/forum?id=HkuGJ3kCb}.

\bibitem[Ni et~al.(2022)Ni, Hernandez~Abrego, Constant, Ma, Hall, Cer, and
  Yang]{ni-etal-2022-sentence}
Jianmo Ni, Gustavo Hernandez~Abrego, Noah Constant, Ji~Ma, Keith Hall, Daniel
  Cer, and Yinfei Yang.
\newblock Sentence-t5: Scalable sentence encoders from pre-trained text-to-text
  models.
\newblock In \emph{Findings of the Association for Computational Linguistics:
  ACL 2022}, pages 1864--1874, Dublin, Ireland, May 2022. Association for
  Computational Linguistics.
\newblock \doi{10.18653/v1/2022.findings-acl.146}.
\newblock URL \url{https://aclanthology.org/2022.findings-acl.146}.

\bibitem[Oord et~al.(2018)Oord, Li, and Vinyals]{Oord2018ctl}
A{\"{a}}ron van~den Oord, Yazhe Li, and Oriol Vinyals.
\newblock Representation learning with contrastive predictive coding.
\newblock \emph{CoRR}, abs/1807.03748, 2018.
\newblock URL \url{http://arxiv.org/abs/1807.03748}.

\bibitem[Ouyang et~al.(2022)Ouyang, Wu, Jiang, Almeida, Wainwright, Mishkin,
  Zhang, Agarwal, Slama, Ray, Schulman, Hilton, Kelton, Miller, Simens, Askell,
  Welinder, Christiano, Leike, and Lowe]{ouyang2022training}
Long Ouyang, Jeff Wu, Xu~Jiang, Diogo Almeida, Carroll~L. Wainwright, Pamela
  Mishkin, Chong Zhang, Sandhini Agarwal, Katarina Slama, Alex Ray, John
  Schulman, Jacob Hilton, Fraser Kelton, Luke Miller, Maddie Simens, Amanda
  Askell, Peter Welinder, Paul Christiano, Jan Leike, and Ryan Lowe.
\newblock Training language models to follow instructions with human feedback,
  2022.

\bibitem[Ouyang et~al.(2021)Ouyang, Wang, Pang, Sun, Tian, Wu, and
  Wang]{ouyang-etal-2021-ernie}
Xuan Ouyang, Shuohuan Wang, Chao Pang, Yu~Sun, Hao Tian, Hua Wu, and Haifeng
  Wang.
\newblock {ERNIE}-{M}: Enhanced multilingual representation by aligning
  cross-lingual semantics with monolingual corpora.
\newblock In \emph{Proceedings of the 2021 Conference on Empirical Methods in
  Natural Language Processing}, pages 27--38, Online and Punta Cana, Dominican
  Republic, November 2021. Association for Computational Linguistics.
\newblock \doi{10.18653/v1/2021.emnlp-main.3}.
\newblock URL \url{https://aclanthology.org/2021.emnlp-main.3}.

\bibitem[Pierre~Zweigenbaum and Rapp(2018)]{ZWEIGENBAUM18.12}
Serge~Sharoff Pierre~Zweigenbaum and Reinhard Rapp.
\newblock Overview of the third bucc shared task: Spotting parallel sentences
  in comparable corpora.
\newblock In Reinhard Rapp, Pierre Zweigenbaum, and Serge Sharoff, editors,
  \emph{Proceedings of the Eleventh International Conference on Language
  Resources and Evaluation (LREC 2018)}, Paris, France, may 2018. European
  Language Resources Association (ELRA).
\newblock ISBN 979-10-95546-07-8.

\bibitem[Ponti et~al.(2020)Ponti, Glava{\v{s}}, Majewska, Liu, Vuli{\'c}, and
  Korhonen]{ponti-etal-2020-xcopa}
Edoardo~Maria Ponti, Goran Glava{\v{s}}, Olga Majewska, Qianchu Liu, Ivan
  Vuli{\'c}, and Anna Korhonen.
\newblock {XCOPA}: A multilingual dataset for causal commonsense reasoning.
\newblock In \emph{Proceedings of the 2020 Conference on Empirical Methods in
  Natural Language Processing (EMNLP)}, pages 2362--2376, Online, November
  2020. Association for Computational Linguistics.
\newblock \doi{10.18653/v1/2020.emnlp-main.185}.
\newblock URL \url{https://aclanthology.org/2020.emnlp-main.185}.

\bibitem[Rajpurkar et~al.(2016)Rajpurkar, Zhang, Lopyrev, and
  Liang]{rajpurkar-etal-2016-squad}
Pranav Rajpurkar, Jian Zhang, Konstantin Lopyrev, and Percy Liang.
\newblock {SQ}u{AD}: 100,000+ questions for machine comprehension of text.
\newblock In \emph{Proceedings of the 2016 Conference on Empirical Methods in
  Natural Language Processing}, pages 2383--2392, Austin, Texas, November 2016.
  Association for Computational Linguistics.
\newblock \doi{10.18653/v1/D16-1264}.
\newblock URL \url{https://aclanthology.org/D16-1264}.

\bibitem[Reimers and Gurevych(2020)]{reimers-gurevych-2020-making}
Nils Reimers and Iryna Gurevych.
\newblock Making monolingual sentence embeddings multilingual using knowledge
  distillation.
\newblock In \emph{Proceedings of the 2020 Conference on Empirical Methods in
  Natural Language Processing (EMNLP)}, pages 4512--4525, Online, November
  2020. Association for Computational Linguistics.
\newblock \doi{10.18653/v1/2020.emnlp-main.365}.
\newblock URL \url{https://aclanthology.org/2020.emnlp-main.365}.

\bibitem[Roy et~al.(2020)Roy, Constant, Al-Rfou, Barua, Phillips, and
  Yang]{roy-etal-2020-lareqa}
Uma Roy, Noah Constant, Rami Al-Rfou, Aditya Barua, Aaron Phillips, and Yinfei
  Yang.
\newblock {LAR}e{QA}: Language-agnostic answer retrieval from a multilingual
  pool.
\newblock In \emph{Proceedings of the 2020 Conference on Empirical Methods in
  Natural Language Processing (EMNLP)}, pages 5919--5930, Online, November
  2020. Association for Computational Linguistics.
\newblock \doi{10.18653/v1/2020.emnlp-main.477}.
\newblock URL \url{https://aclanthology.org/2020.emnlp-main.477}.

\bibitem[Ruder et~al.(2021)Ruder, Constant, Botha, Siddhant, Firat, Fu, Liu,
  Hu, Garrette, Neubig, and Johnson]{ruder-etal-2021-xtreme}
Sebastian Ruder, Noah Constant, Jan Botha, Aditya Siddhant, Orhan Firat, Jinlan
  Fu, Pengfei Liu, Junjie Hu, Dan Garrette, Graham Neubig, and Melvin Johnson.
\newblock {XTREME}-{R}: Towards more challenging and nuanced multilingual
  evaluation.
\newblock In \emph{Proceedings of the 2021 Conference on Empirical Methods in
  Natural Language Processing}, pages 10215--10245, Online and Punta Cana,
  Dominican Republic, November 2021. Association for Computational Linguistics.
\newblock \doi{10.18653/v1/2021.emnlp-main.802}.
\newblock URL \url{https://aclanthology.org/2021.emnlp-main.802}.

\bibitem[Sap et~al.(2019)Sap, Rashkin, Chen, Le~Bras, and
  Choi]{sap-etal-2019-social}
Maarten Sap, Hannah Rashkin, Derek Chen, Ronan Le~Bras, and Yejin Choi.
\newblock Social {IQ}a: Commonsense reasoning about social interactions.
\newblock In \emph{Proceedings of the 2019 Conference on Empirical Methods in
  Natural Language Processing and the 9th International Joint Conference on
  Natural Language Processing (EMNLP-IJCNLP)}, pages 4463--4473, Hong Kong,
  China, November 2019. Association for Computational Linguistics.
\newblock \doi{10.18653/v1/D19-1454}.
\newblock URL \url{https://aclanthology.org/D19-1454}.

\bibitem[Saunshi et~al.(2019)Saunshi, Plevrakis, Arora, Khodak, and
  Khandeparkar]{saunshi2019theoretical}
Nikunj Saunshi, Orestis Plevrakis, Sanjeev Arora, Mikhail Khodak, and
  Hrishikesh Khandeparkar.
\newblock A theoretical analysis of contrastive unsupervised representation
  learning.
\newblock In \emph{International Conference on Machine Learning}, pages
  5628--5637. PMLR, 2019.

\bibitem[Schwenk et~al.(2021)Schwenk, Wenzek, Edunov, Grave, Joulin, and
  Fan]{schwenk-etal-2021-ccmatrix}
Holger Schwenk, Guillaume Wenzek, Sergey Edunov, Edouard Grave, Armand Joulin,
  and Angela Fan.
\newblock {CCM}atrix: Mining billions of high-quality parallel sentences on the
  web.
\newblock In \emph{Proceedings of the 59th Annual Meeting of the Association
  for Computational Linguistics and the 11th International Joint Conference on
  Natural Language Processing (Volume 1: Long Papers)}, pages 6490--6500,
  Online, August 2021. Association for Computational Linguistics.
\newblock \doi{10.18653/v1/2021.acl-long.507}.
\newblock URL \url{https://aclanthology.org/2021.acl-long.507}.

\bibitem[Specia et~al.(2021)Specia, Blain, Fomicheva, Zerva, Li, Chaudhary, and
  Martins]{specia-etal-2021-findings}
Lucia Specia, Fr{\'e}d{\'e}ric Blain, Marina Fomicheva, Chrysoula Zerva,
  Zhenhao Li, Vishrav Chaudhary, and Andr{\'e} F.~T. Martins.
\newblock Findings of the {WMT} 2021 shared task on quality estimation.
\newblock In \emph{Proceedings of the Sixth Conference on Machine Translation},
  pages 684--725, Online, November 2021. Association for Computational
  Linguistics.
\newblock URL \url{https://aclanthology.org/2021.wmt-1.71}.

\bibitem[Teller(2000)]{10.1162/089120100750105975}
Virginia Teller.
\newblock {Speech and Language Processing: An Introduction to Natural Language
  Processing, Computational Linguistics, and Speech Recognition}.
\newblock \emph{Computational Linguistics}, 26\penalty0 (4):\penalty0 638--641,
  12 2000.
\newblock ISSN 0891-2017.
\newblock \doi{10.1162/089120100750105975}.
\newblock URL \url{https://doi.org/10.1162/089120100750105975}.

\bibitem[Tiyajamorn et~al.(2021)Tiyajamorn, Kajiwara, Arase, and
  Onizuka]{tiyajamorn-etal-2021-language}
Nattapong Tiyajamorn, Tomoyuki Kajiwara, Yuki Arase, and Makoto Onizuka.
\newblock Language-agnostic representation from multilingual sentence encoders
  for cross-lingual similarity estimation.
\newblock In \emph{Proceedings of the 2021 Conference on Empirical Methods in
  Natural Language Processing}, pages 7764--7774, Online and Punta Cana,
  Dominican Republic, November 2021. Association for Computational Linguistics.
\newblock \doi{10.18653/v1/2021.emnlp-main.612}.
\newblock URL \url{https://aclanthology.org/2021.emnlp-main.612}.

\bibitem[van~der Maaten and Hinton(2008)]{JMLR:v9:vandermaaten08a}
Laurens van~der Maaten and Geoffrey Hinton.
\newblock Visualizing data using t-sne.
\newblock \emph{Journal of Machine Learning Research}, 9\penalty0
  (86):\penalty0 2579--2605, 2008.
\newblock URL \url{http://jmlr.org/papers/v9/vandermaaten08a.html}.

\bibitem[Vaswani et~al.(2017)Vaswani, Shazeer, Parmar, Uszkoreit, Jones, Gomez,
  Kaiser, and Polosukhin]{Vaswani2017Attention}
Ashish Vaswani, Noam Shazeer, Niki Parmar, Jakob Uszkoreit, Llion Jones,
  Aidan~N Gomez, Lukasz Kaiser, and Illia Polosukhin.
\newblock Attention is all you need.
\newblock In \emph{Advances in Neural Information Processing Systems 30, {NIPS}
  2017 4-9 December 2017, Long Beach, CA, {USA}}, pages 5998--6008, 2017.
\newblock URL \url{http://papers.nips.cc/paper/7181-attention-is-all-you-need}.

\bibitem[Wang and Liu(2021)]{Wang_2021_CVPR}
Feng Wang and Huaping Liu.
\newblock Understanding the behaviour of contrastive loss.
\newblock In \emph{Proceedings of the IEEE/CVF Conference on Computer Vision
  and Pattern Recognition (CVPR)}, pages 2495--2504, June 2021.

\bibitem[Wang et~al.(2021)Wang, Zhao, and Liu]{wang-etal-2021-aligning}
Liang Wang, Wei Zhao, and Jingming Liu.
\newblock Aligning cross-lingual sentence representations with dual momentum
  contrast.
\newblock In \emph{Proceedings of the 2021 Conference on Empirical Methods in
  Natural Language Processing}, pages 3807--3815, Online and Punta Cana,
  Dominican Republic, November 2021. Association for Computational Linguistics.
\newblock \doi{10.18653/v1/2021.emnlp-main.309}.
\newblock URL \url{https://aclanthology.org/2021.emnlp-main.309}.

\bibitem[Wei et~al.(2021)Wei, Weng, Hu, Xing, Yu, and Luo]{wei2021on}
Xiangpeng Wei, Rongxiang Weng, Yue Hu, Luxi Xing, Heng Yu, and Weihua Luo.
\newblock On learning universal representations across languages.
\newblock In \emph{International Conference on Learning Representations}, 2021.
\newblock URL \url{https://openreview.net/forum?id=Uu1Nw-eeTxJ}.

\bibitem[Wei et~al.(2022)Wei, Yu, Hu, Weng, Luo, and
  Jin]{wei-etal-2022-learning}
Xiangpeng Wei, Heng Yu, Yue Hu, Rongxiang Weng, Weihua Luo, and Rong Jin.
\newblock Learning to generalize to more: Continuous semantic augmentation for
  neural machine translation.
\newblock In \emph{Proceedings of the 60th Annual Meeting of the Association
  for Computational Linguistics (Volume 1: Long Papers)}, pages 7930--7944,
  Dublin, Ireland, May 2022. Association for Computational Linguistics.
\newblock \doi{10.18653/v1/2022.acl-long.546}.
\newblock URL \url{https://aclanthology.org/2022.acl-long.546}.

\bibitem[Wenzek et~al.(2020)Wenzek, Lachaux, Conneau, Chaudhary, Guzm{\'a}n,
  Joulin, and Grave]{wenzek-etal-2020-ccnet}
Guillaume Wenzek, Marie-Anne Lachaux, Alexis Conneau, Vishrav Chaudhary,
  Francisco Guzm{\'a}n, Armand Joulin, and Edouard Grave.
\newblock {CCN}et: Extracting high quality monolingual datasets from web crawl
  data.
\newblock In \emph{Proceedings of the Twelfth Language Resources and Evaluation
  Conference}, pages 4003--4012, Marseille, France, May 2020. European Language
  Resources Association.
\newblock ISBN 979-10-95546-34-4.
\newblock URL \url{https://aclanthology.org/2020.lrec-1.494}.

\bibitem[Wierzbicka(1999)]{wierzbicka1999emotions}
Anna Wierzbicka.
\newblock \emph{Emotions across languages and cultures: Diversity and
  universals}.
\newblock Cambridge university press, 1999.

\bibitem[Williams et~al.(2018)Williams, Nangia, and
  Bowman]{williams-etal-2018-broad}
Adina Williams, Nikita Nangia, and Samuel Bowman.
\newblock A broad-coverage challenge corpus for sentence understanding through
  inference.
\newblock In \emph{Proceedings of the 2018 Conference of the North {A}merican
  Chapter of the Association for Computational Linguistics: Human Language
  Technologies, Volume 1 (Long Papers)}, pages 1112--1122, New Orleans,
  Louisiana, June 2018. Association for Computational Linguistics.
\newblock \doi{10.18653/v1/N18-1101}.
\newblock URL \url{https://aclanthology.org/N18-1101}.

\bibitem[Workshop et~al.(2023)Workshop, :, Scao, Fan, Akiki, Pavlick, Ilić,
  Hesslow, Castagné, Luccioni, and et~al.]{workshop2023bloom}
BigScience Workshop, :, Teven~Le Scao, Angela Fan, Christopher Akiki, Ellie
  Pavlick, Suzana Ilić, Daniel Hesslow, Roman Castagné, Alexandra~Sasha
  Luccioni, and et~al.
\newblock Bloom: A 176b-parameter open-access multilingual language model,
  2023.

\bibitem[Wu et~al.(2022)Wu, Gao, Lin, Han, Wang, and
  Hu]{https://doi.org/10.48550/arxiv.2210.06432}
Xing Wu, Chaochen Gao, Zijia Lin, Jizhong Han, Zhongyuan Wang, and Songlin Hu.
\newblock Infocse: Information-aggregated contrastive learning of sentence
  embeddings, 2022.
\newblock URL \url{https://arxiv.org/abs/2210.06432}.

\bibitem[Yang et~al.(2021)Yang, Wei, Jiao, Jiang, and
  Yang]{yang-etal-2021-xmoco}
Nan Yang, Furu Wei, Binxing Jiao, Daxing Jiang, and Linjun Yang.
\newblock x{M}o{C}o: Cross momentum contrastive learning for open-domain
  question answering.
\newblock In \emph{Proceedings of the 59th Annual Meeting of the Association
  for Computational Linguistics and the 11th International Joint Conference on
  Natural Language Processing (Volume 1: Long Papers)}, pages 6120--6129,
  Online, August 2021. Association for Computational Linguistics.
\newblock \doi{10.18653/v1/2021.acl-long.477}.
\newblock URL \url{https://aclanthology.org/2021.acl-long.477}.

\bibitem[Yang et~al.(2019{\natexlab{a}})Yang, Hernandez~Abrego, Yuan, Guo,
  Shen, Cer, Sung, Strope, and Kurzweil]{ijcai2019p0746}
Yinfei Yang, Gustavo Hernandez~Abrego, Steve Yuan, Mandy Guo, Qinlan Shen,
  Daniel Cer, Yun-hsuan Sung, Brian Strope, and Ray Kurzweil.
\newblock Improving multilingual sentence embedding using bi-directional dual
  encoder with additive margin softmax.
\newblock In \emph{Proceedings of the Twenty-Eighth International Joint
  Conference on Artificial Intelligence, {IJCAI-19}}, pages 5370--5378.
  International Joint Conferences on Artificial Intelligence Organization, 7
  2019{\natexlab{a}}.
\newblock \doi{10.24963/ijcai.2019/746}.
\newblock URL \url{https://doi.org/10.24963/ijcai.2019/746}.

\bibitem[Yang et~al.(2019{\natexlab{b}})Yang, Zhang, Tar, and
  Baldridge]{yang-etal-2019-paws}
Yinfei Yang, Yuan Zhang, Chris Tar, and Jason Baldridge.
\newblock {PAWS}-{X}: A cross-lingual adversarial dataset for paraphrase
  identification.
\newblock In \emph{Proceedings of the 2019 Conference on Empirical Methods in
  Natural Language Processing and the 9th International Joint Conference on
  Natural Language Processing (EMNLP-IJCNLP)}, pages 3687--3692, Hong Kong,
  China, November 2019{\natexlab{b}}. Association for Computational
  Linguistics.
\newblock \doi{10.18653/v1/D19-1382}.
\newblock URL \url{https://aclanthology.org/D19-1382}.

\bibitem[Zhang et~al.(2021)Zhang, Li, Xiao, Zhu, Nallapati, Arnold, and
  Xiang]{zhang-etal-2021-pairwise}
Dejiao Zhang, Shang-Wen Li, Wei Xiao, Henghui Zhu, Ramesh Nallapati, Andrew~O.
  Arnold, and Bing Xiang.
\newblock Pairwise supervised contrastive learning of sentence representations.
\newblock In \emph{Proceedings of the 2021 Conference on Empirical Methods in
  Natural Language Processing}, pages 5786--5798, Online and Punta Cana,
  Dominican Republic, November 2021. Association for Computational Linguistics.
\newblock \doi{10.18653/v1/2021.emnlp-main.467}.
\newblock URL \url{https://aclanthology.org/2021.emnlp-main.467}.

\bibitem[Zhang et~al.(2022{\natexlab{a}})Zhang, Xiao, Zhu, Ma, and
  Arnold]{zhang-etal-2022-virtual}
Dejiao Zhang, Wei Xiao, Henghui Zhu, Xiaofei Ma, and Andrew Arnold.
\newblock Virtual augmentation supported contrastive learning of sentence
  representations.
\newblock In \emph{Findings of the Association for Computational Linguistics:
  ACL 2022}, pages 864--876, Dublin, Ireland, May 2022{\natexlab{a}}.
  Association for Computational Linguistics.
\newblock \doi{10.18653/v1/2022.findings-acl.70}.
\newblock URL \url{https://aclanthology.org/2022.findings-acl.70}.

\bibitem[Zhang et~al.(2018)Zhang, Cisse, Dauphin, and
  Lopez-Paz]{zhang2018mixup}
Hongyi Zhang, Moustapha Cisse, Yann~N. Dauphin, and David Lopez-Paz.
\newblock mixup: Beyond empirical risk minimization.
\newblock In \emph{International Conference on Learning Representations}, 2018.
\newblock URL \url{https://openreview.net/forum?id=r1Ddp1-Rb}.

\bibitem[Zhang et~al.(2022{\natexlab{b}})Zhang, Roller, Goyal, Artetxe, Chen,
  Chen, Dewan, Diab, Li, Lin, Mihaylov, Ott, Shleifer, Shuster, Simig, Koura,
  Sridhar, Wang, and Zettlemoyer]{zhang2022opt}
Susan Zhang, Stephen Roller, Naman Goyal, Mikel Artetxe, Moya Chen, Shuohui
  Chen, Christopher Dewan, Mona Diab, Xian Li, Xi~Victoria Lin, Todor Mihaylov,
  Myle Ott, Sam Shleifer, Kurt Shuster, Daniel Simig, Punit~Singh Koura, Anjali
  Sridhar, Tianlu Wang, and Luke Zettlemoyer.
\newblock Opt: Open pre-trained transformer language models,
  2022{\natexlab{b}}.

\bibitem[Zhang et~al.(2022{\natexlab{c}})Zhang, Zhang, Mensah, Liu, and
  Mao]{Zhang_Zhang_Mensah_Liu_Mao_2022}
Yanzhao Zhang, Richong Zhang, Samuel Mensah, Xudong Liu, and Yongyi Mao.
\newblock Unsupervised sentence representation via contrastive learning with
  mixing negatives.
\newblock \emph{Proceedings of the AAAI Conference on Artificial Intelligence},
  36\penalty0 (10):\penalty0 11730--11738, Jun. 2022{\natexlab{c}}.
\newblock \doi{10.1609/aaai.v36i10.21428}.
\newblock URL \url{https://ojs.aaai.org/index.php/AAAI/article/view/21428}.

\bibitem[Zhang et~al.(2019)Zhang, Baldridge, and He]{zhang-etal-2019-paws}
Yuan Zhang, Jason Baldridge, and Luheng He.
\newblock {PAWS}: Paraphrase adversaries from word scrambling.
\newblock In \emph{Proceedings of the 2019 Conference of the North {A}merican
  Chapter of the Association for Computational Linguistics: Human Language
  Technologies, Volume 1 (Long and Short Papers)}, pages 1298--1308,
  Minneapolis, Minnesota, June 2019. Association for Computational Linguistics.
\newblock \doi{10.18653/v1/N19-1131}.
\newblock URL \url{https://aclanthology.org/N19-1131}.

\bibitem[Ziemski et~al.(2016)Ziemski, Junczys-Dowmunt, and
  Pouliquen]{ziemski-etal-2016-united}
Micha{\l} Ziemski, Marcin Junczys-Dowmunt, and Bruno Pouliquen.
\newblock The {U}nited {N}ations parallel corpus v1.0.
\newblock In \emph{Proceedings of the Tenth International Conference on
  Language Resources and Evaluation ({LREC}'16)}, pages 3530--3534,
  Portoro{\v{z}}, Slovenia, May 2016. European Language Resources Association
  (ELRA).
\newblock URL \url{https://aclanthology.org/L16-1561}.

\bibitem[Zweigenbaum et~al.(2017)Zweigenbaum, Sharoff, and
  Rapp]{zweigenbaum-etal-2017-overview}
Pierre Zweigenbaum, Serge Sharoff, and Reinhard Rapp.
\newblock Overview of the second {BUCC} shared task: Spotting parallel
  sentences in comparable corpora.
\newblock In \emph{Proceedings of the 10th Workshop on Building and Using
  Comparable Corpora}, pages 60--67, Vancouver, Canada, August 2017.
  Association for Computational Linguistics.
\newblock \doi{10.18653/v1/W17-2512}.
\newblock URL \url{https://www.aclweb.org/anthology/W17-2512}.

\end{thebibliography}

\newpage
\appendix

\section{Training Corpora and Hyper-parameters}
\label{appendix:pretraining-data}

\subsection{Training Corpora}
As for monolingual data, we follow \citet{conneau-etal-2020-unsupervised} to build a Common-Crawl Corpus using the CCNet \citep{wenzek-etal-2020-ccnet} tool\footnote{\url{https://github.com/facebookresearch/cc_net}}, which is widely used in the literature~\cite{huang-etal-2019-unicoder,luo-etal-2021-veco,chi-etal-2021-infoxlm,wei2021on}. Further, we collect parallel corpora from CCAligned \cite{el-kishky-etal-2020-ccaligned}, CCMatrix \cite{schwenk-etal-2021-ccmatrix}, WMT \cite{akhbardeh-etal-2021-findings}, and MultiUN \cite{ziemski-etal-2016-united}, involving 94 languages with more than 4.8 billion sentence pairs. We use the OpusFilter\footnote{\url{https://github.com/Helsinki-NLP/OpusFilter}} tool to remove noisy bitexts, which results in 3.2 billion sentence pairs. Table~\ref{training-corpora-table} shows the statistics for both monolingual and parallel data. We apply subword tokenization directly on raw text data using Sentence Piece Model~\cite{kudo-richardson-2018-sentencepiece} without any additional preprocessing. To better support our motivation that \textsc{Emma}-X can cover more languages than previous cross-lingual sentence representations, we divide Tatoeba~\cite{artetxe-schwenk-2019-massively} into two subsets: ``Head'', containing languages usually covered in previous methods, and ``Long-tail'', with other languages. We treat the 36 languages contained in XTREME~\citet{ruder-etal-2021-xtreme} as head languages, which are: \textbf{``ar, he, vi, id, jv, tl, eu, ml, ta, te, af, nl, en, de, el, bn, hi, mr, ur, fa, fr, it, pt, es, bg, ru, ja, ka, ko, th, sw, zh, kk, tr, et, fi, hu, az, lt, pl, uk, ro''}. The remaining 76 languages in Tatoeba are treated as long-tail ones.

\subsection{Hyper-parameters}
The parameters of \textsc{Emma}-X are first initialized with \textsc{Xlm-r}, with 24 layers of Transformer \citep{Vaswani2017Attention} encoder, 1024 hidden states, and 16 attention heads. We set the total semantic ranks as 4. The GMM classifier is implemented as a mixture of Gaussian forms, each of which consists of a prior $\pi \in \mathbb{R}^1$, a mean $\mu \in \mathbb{R}^{1024}$, and a variance $\sigma \in \mathbb{R}^{1024}$, all are trainable variables. We optimize the GMM classifier with Adam ($\beta_1$=0.9, $\beta_2$=0.999) \cite{DBLP:journals/corr/KingmaB14} using a batch size of 1024 and a learning rate of 3e-5. For cross-lingual encoder, we apply the same training setting as MoCo \cite{he2020momentum}, with the momentum queue $K$ to be 256 and temperature as 0.04. We set the momentum coefficient to 0.999 and use the Adam optimizer with a cosine decay learning rate whose peak is 5e-4. 

\section{FLORES-200 Dataset and Geometric Analysis}
\label{app:flores-and-three-metrics}

\subsection{FLORES-200 dataset}
FLORES-200 \cite{goyal-etal-2022-flores,costa2022no} is a many-to-many multilingual benchmark, which consists of 3001 sentences in 204 total languages. FLORES-200 sourced all sentences from English WikiMedia and translated these English sentences into 204 languages by human translators. In particular, sentences in FLORES-200 have a much larger breadth of topics, for they are collected from three different sources: WikiNews\footnote{\url{https://en.wikinews.org/wiki/Main Page}}, WikiJunior\footnote{\url{https://en.wikibooks.org/wiki/Wikijunior}} and WikiVoyage\footnote{\url{https://en.wikivoyage.org/wiki/Main_Page}}. We summarize the basic statistics of all languages in FLORES-200 in Table~\ref{tbl:flores-table}. Similar to Tatoeba~\citep{artetxe-schwenk-2019-massively}, we treat English data ``eng\_Latn'' as retrieval labels and report the retrieval accuracy using the same scripts as Tatoeba in XTREME~\citep{ruder-etal-2021-xtreme}. We set the 68 languages: \textbf{``bel\_Cyrl, bos\_Latn, hun\_Latn, epo\_Latn, khm\_Khmr, urd\_Arab, srp\_Cyrl, jav\_Latn, hye\_Armn, gla\_Latn, por\_Latn, lit\_Latn, bul\_Cyrl, slk\_Latn, mal\_Mlym, ita\_Latn, nno\_Latn, mar\_Deva, hrv\_Latn, hin\_Deva, kat\_Geor, ben\_Beng, fin\_Latn, cym\_Latn, oci\_Latn, cat\_Latn, fao\_Latn, xho\_Latn, spa\_Latn, ron\_Latn, amh\_Ethi, ces\_Latn, swe\_Latn, nld\_Latn, tat\_Cyrl, kor\_Hang, glg\_Latn, fra\_Latn, eus\_Latn, ind\_Latn, dan\_Latn, tha\_Thai, deu\_Latn, tel\_Telu, afr\_Latn, pol\_Latn, est\_Latn, uig\_Arab, ukr\_Cyrl, uzn\_Latn, heb\_Hebr, kaz\_Cyrl, nob\_Latn, rus\_Cyrl, vie\_Latn, arb\_Arab, zho\_Hans, tuk\_Latn, khk\_Cyrl, jpn\_Jpan, ell\_Grek, isl\_Latn, tam\_Taml, slv\_Latn, tur\_Latn, mkd\_Cyrl, tgl\_Latn, gle\_Latn''} as ``Head'' languages, and the remaining 135 languages (excluded English data) as ``Long-tail'' ones.

\begin{table*}[t]
\begin{center}
\resizebox{\textwidth}{!}{
\begin{tabular}{crrcrrcrrcrrcrr}
\toprule
\bf Code & \bf Size (GB) & \bf Sent. (M) & \bf Code & \bf Size (GB) & \bf Sent. (M) & \bf Code & \bf Size (GB) & \bf Sent. (M) & \bf Code & \bf Size (GB) & \bf Sent. (M) & \bf Code & \bf Size (GB) & \bf Sent. (M) \\ \cmidrule(r){1-3}\cmidrule{4-6}\cmidrule(l){7-9}\cmidrule(l){10-12}\cmidrule(l){13-15}
af & 1.3 & - & et & 6.1 & 22.3 & ja & 24.2 & 89.2 & mt & 0.2 & - & sq & 3.0 & - \\
am & 0.7 & - & eu & 2.0 & 0.81 & jv & 0.2 & - & my & 0.9 & - & sr & 5.1 & - \\
ar & 20.4 & 72.3 & fa & 21.6 & 7.5 & ka & 3.4 & 2.0 & ne & 2.6 & - & su & 0.1 & - \\
as & 0.1 & - & fi & 19.2 & 92.8 & kk & 2.6 & 2.8 & nl & 15.8 & 66.0 & sv & 10.8 & 74.2\\
az & 3.6 & 0.82 & fr & 46.5 & 331.5 & km & 1.0 & 0.84 & no & 3.7 & - & sw & 1.6 & 1.7\\
be & 3.5 & 0.51 & fy & 0.2 & 0.13 & kn & 1.2 & - & om & 0.1 & - & ta & 8.2 & 2.79\\
bg & 22.6 & 47.2 & ga & 0.5 & - & ko & 17.2 & 79.3 & or & 0.6 & - & te & 2.6 & -\\
bn & 7.9 & 7.52 & gd & 0.1 & 0.05 & ku & 0.4 & - & pa & 0.8 & - & th & 14.7 & 13.1\\
br & 0.1 & - & gl & 2.9 & 0.77 & ky & 1.2 & - & pl & 16.8 & 79.7 & tl & 0.8 & -\\
bs & 0.1 & - & gu & 0.3 & - & la & 2.5 & - & ps & 0.7 & - & tr & 17.3 & 93.8\\
ca & 10.1 & 14.9 & ha & 0.3 & - & lo & 0.6 & - & pt & 15.9 & 247.6 & ug & 0.4 & - \\
cs & 16.3 & 108.4 & he & 6.7 & 47.1 & lt & 7.2 & 11.0 & ro & 8.6 & 60.4 & uk & 9.1 & 0.78\\
cy & 0.8 & - & hi & 20.2 & 3.2 & lv & 6.4 & 0.37 & ru & 48.1 & 134.9 & ur & 5.0 & 1.15\\
da & 15.2 & 8.0 & hr & 5.4 & - & mg & 0.2  & - & sa & 0.3 & - & uz & 0.7 & - \\
de & 46.3 & 283.4 & hu & 9.5 & 55.2 & mk & 1.9 & - & sd & 0.4 & - & vi & 44.6 & 15.3\\
el & 29.3 & 95.1 & hy & 5.5 & 1.7 & ml & 4.3 & 1.07 & si & 2.1 & 0.60 & xh & 0.1 & - \\
en & 49.7 & - & id & 10.6 & 184.6 & mn & 1.7 & 0.19 & sk & 4.9 & - & yi & 0.3 & - \\
eo & 0.9 & 0.18 & is & 1.3 & - & mr & 1.3 & - & sl & 2.8 & 9.8 & zh & 36.8 & 379.4\\
es & 44.6 & 279.6 & it & 19.8 & 179.3 & ms & 3.2 & 2.1 & so & 0.4 & - & - & -\\
\bottomrule
\end{tabular}}
\end{center}
\caption{The statistics of CC-100 and the collected parallel corpora used for training. We report the list of 94 languages and include
the size of the monolingual data (in GiB) and the number of sentence pairs (in Millions, which denotes the number of sentence pairs between the specific language and English) in parallel corpora for each language. ``-'' means the number of sentence pairs is less than 0.1 million.}
\label{training-corpora-table}
\end{table*}

\begin{table}[t]
\begin{center}
\footnotesize
\resizebox{0.48\linewidth}{!}{
\begin{tabular}{lrr}
\toprule
\multicolumn{2}{l}{Number of Sentences} & 3001 \\
\multicolumn{2}{l}{Average Words per Sentence} & 21 \\
\multicolumn{2}{l}{Number of Articles} & 842 \\
\multicolumn{2}{l}{Average Number of Sentences per Article} & 3.5 \\
\midrule
\textbf{Domain} &\textbf{Articles} &\textbf{Sentences} \\
WikiNews & 309 & 993 \\
WikiJunior & 284 & 1006 \\
WikiVoyage & 249 & 1002 \\
\midrule
\textbf{Sub-Topic} &\textbf{Articles} &\textbf{Sentences} \\
Crime & 155 & 313 \\
Disasters & 27 & 65 \\
Entertainment & 28 & 68 \\
Geography & 36 & 86 \\
Health & 27 & 67 \\
Nature & 17 & 45 \\
Politics & 171 & 341 \\
Science & 154 & 325 \\
Sports & 154 & 162 \\
Travel & 505 & 1529 \\
\bottomrule
\end{tabular}}
\end{center}
\caption{Basic Statistics of FLORES-200.}
\label{tbl:flores-table}
\end{table}

\subsection{Three measurements in Geometric Analysis}
\noindent\textbf{Invariance Measurement} implies whether the semantic distributions of all languages are similar~\citep{abend-rappoport-2017-state}. We adopt a Gaussian form ${\scriptstyle \mathcal{N}_l(\mu_l,\sigma_l^2)}$ where ${\scriptstyle \mu_l=\frac{\sum_{\mathbf{x} \in l}\gamma^{(\mathbf{x})}}{3001}}$ and ${\scriptstyle \sigma_l^2= \sum_{\mathbf{x} \in l} (\gamma^{(\mathbf{x})} -\mu_l)(\gamma^{(\mathbf{x})} -\mu_l)^T}$, to approximate the semantic distribution of each language $l$. Further, we compute the mean averaged KL-divergence (KL-D for short)~\citep{10.1214/aoms/1177729694} among all language pairs as the overall Invariance score $\boldsymbol{\scriptstyle \mathcal{I}_v}$ with $L$ as the total number of languages: 
\begin{equation}
\scriptsize
    \mathcal{I}_v = \frac{1}{L \times (L-1)} \sum_{l_1 \neq l_2} \frac{\mathbf{KL}(\mathcal{N}_{l_1} || \mathcal{N}_{l_2}) + \mathbf{KL}(\mathcal{N}_{l_2} || \mathcal{N}_{l_1})}{2}.
\end{equation}

\noindent\textbf{Canonical Form Measurement} Previous works \citep{10.1162/089120100750105975,irwin2009methodology} have demonstrated that a good multilingual space should distribute sentence representations based on their semantic similarities rather than language families. To measure this in quantity, we focus on Calinski-Harabasz Index (CH-I) \citep{doi:10.1080/03610927408827101}, which measures how similar an object is to its own cluster compared to other clusters. We group all semantically equivalent sentences in a cluster, which leads to 3001 clusters and each obsesses 204 sentences in 204 different languages. Assuming $c_k$ and $c$ are the centroid of cluster $k$ and the whole dataset ${\scriptstyle \mathcal{S}}$, respectively. The CH-I $\boldsymbol{\scriptstyle \mathcal{C}_h}$ is defined as:
\begin{equation}
    \scriptsize
    \mathcal{C}_h = \Big[ 204 \times \sum_{k=1}^{K} \Vert c_k - c \Vert ^2 \Big] / \Big[ \sum_{k=1}^{K} \sum_{s \in \mathcal{S}} \Vert s - c_k \Vert ^2 \Big].
\end{equation}
The higher the CH-I is, the better the semantically equivalent sentences are clustered.

\noindent\textbf{Isotropy Measurement} A high-dimensional embedding space often demonstrates poor isotropy, and deteriorates into a low-dimensional manifold that greatly limits the expressive ability of the embedding space. We adopt principal ratio (PR)~\citep{mu2018allbutthetop} to measure isotropy. Let $\boldsymbol{\scriptstyle \mathcal{E}}$ be the sentence representation matrix, $\boldsymbol{\scriptstyle \mathcal{V}}$ be the set of the eigenvectors of $\boldsymbol{\scriptstyle \mathcal{E}}$, the Isotropy $\boldsymbol{\scriptstyle \mathcal{I}_{so}}$ is
\begin{equation}
    \footnotesize
    \mathcal{I}_{so} = {\min_{v \in \mathcal{V}}  \sum_{e \in \mathcal{E} } \exp({v^{\mathsf{T}}e}) } / {\max_{v \in \mathcal{V}}  \sum_{e \in \mathcal{E} } \exp({v^{\mathsf{T}}e}) }.
\end{equation}
The closer $\boldsymbol{\scriptstyle \mathcal{I}_{so}}$ is to $1$, the more isotropic the representation space is. 

\section{\textsc{xrete}: Cross-lingual Representation Transfer Evaluation}
\label{app:benchmark_info}
\textsc{xrete} consists of 12 tasks that fall into four different categories. In our ``translate-train-all'' setting, we individually fine-tune models with English training set and its translated training sets on each task. Then we report the performance of our fine-tuned model. We give an overview in Table~\ref{tbl:overview-of-evaluation-dataset} and describe the task details as follows.

\begin{table*}[t]
\centering
\footnotesize
\resizebox{\textwidth}{!}{
\begin{tabular}{llrrrrrr}\\ 
\toprule
\textbf{Task category} & \textbf{Task} & \textbf{Train} & \textbf{Dev} & \textbf{Test} & \textbf{Lang.} & \textbf{Metric} & \textbf{Domain} \\
\toprule
\multirow{2}{*}{Inference} & AmericasNLI & 392,702 & 222-743 & 738-750    & 10 & Accuracy & Misc.\\
~& XNLI & 392,702 & 2,490 & 5,010 & 15 & Accuracy & Misc. \\
\midrule
\multirow{2}{*}{Semantic Similarity} & Multi-STS & 550,152+5,749 & 10,000+1,500 & 250 & 7 & Spearman & Misc. \\
~& WMT21QETask1 & 7,000 & 1,000 & 1,000 & 7 (11) & Pearson & News \\
\midrule
\multirow{4}{*}{Sentence Retrieval} & LAReQA & 87,599 & 10,579 & 1,190 & 11 & mAP@20  & Wikipedia \\
~& Mewsli-X & 116,093 & 10,252 & 428-1,482 & 11 (50) & mAP@20  & News \\
~& BUCC       & -           & -          & 1,896-14,330 & 5       & F1   & Wiki/News    \\
~& Tatoeba      & -           & -          & 1,000       & 36 (122) & Accuracy & Misc. \\
\midrule
\multirow{4}{*}{Classification} & XCOPA        & 33,410+400   & 100        & 500        & 11      & Accuracy & Misc. \\
~& MultiEURLEX  & 55,000       & 5,000       & 5,000       & 23      & Accuracy & Legal \\
~& MultiARC     & 200,000      & 5,000       & 5,000       & 6       & MAE   & Reviews   \\
~& PAWS-X       & 49,401       & 2,000       & 2,000       & 7       & Accuracy & Wiki/Quora \\
\bottomrule
\end{tabular}
}
\caption{Overview of \textsc{xrete} tasks. For tasks that have training and dev sets in other languages, we only report the number of sentences in English sets. We report the number of test examples per language.}
\label{tbl:overview-of-evaluation-dataset}
\end{table*}

\paragraph{XNLI} The Cross-lingual Natural Language Inference corpus \cite{conneau-etal-2018-xnli} tasks the systems with reading two sentences and determining whether one entails the other, contradicts it, or neither (neutral). A crowdsourcing-based procedure is used for collecting English examples, which are later translated into ten target languages for evaluation. Training data stays consistent with the English training data of MultiNLI \cite{williams-etal-2018-broad}. For evaluation, we concatenate two sentences as input and apply a new classification head to distinguish sentence relationships. We perform ``translate-train-all'' evaluation, where the model is first fine-tuned on English training data and its translated data in other languages, then evaluated on test sets.

\paragraph{AmericasNLI (ANLI)} The AmericasNLI \cite{ebrahimi-etal-2022-americasnli} is an extension of XNLI task to 10 Indigenous languages of the Americas. All of these languages are truly low-resource languages and serve as a good testbed for zero-shot cross-lingual transferability. As Spanish is more relative to the target languages, the Spanish version of XNLI subset is translated for evaluation. For training, both English and Spanish versions of MultiNLI training data are provided. We evaluate on ANLI following the same settings as in XNLI.

\paragraph{MultiSTS} The Multilingual Semantic Textual Similarity dataset \cite{cer-etal-2017-semeval, reimers-gurevych-2020-making} aims to assign a semantic similarity score for a pair of sentences. The MultiSTS dataset contains 7 cross-lingual sentence pairs and 3 monolingual pairs. Stanford NLI \cite{bowman-etal-2015-large} and English STS \cite{cer-etal-2017-semeval} are provided as training sets. We report the results after first fine-tuning on English training set using a Siamese network structure~\citep{reimers-gurevych-2020-making}. Then we compute the cosine similarity between the sentence pairs and compute Spearman's rank correlation between the predicted score and gold score following \citet{reimers-gurevych-2020-making}.

\paragraph{WMT21QETask1 (QE)} The WMT21 Quality Estimation Task 1 Sentence-level Direct Assessment \cite{specia-etal-2021-findings} aims at testing the translation quality and this task has been applied to test the sensitivity of language models to semantic similarity \cite{tiyajamorn-etal-2021-language}. The training and evaluation sets are collected from Wikipedia by translating sentences using state-of-the-art translation models to 6 languages and annotated by professional translators. In WMT21, 4 new language pairs with no training data are given to test zero-shot cross-lingual transferability. Our evaluation setting on QE is similar to that on MultiSTS, but we report Pearson's rank correlation \citep{kepler-etal-2019-openkiwi}.

\paragraph{LAReQA} The Language-Agnostic Retrieval Question Answering \cite{roy-etal-2020-lareqa} is a QA retrieval task where models are required to retrieve all relevant answers in different languages over a large multilingual pool. The dataset is constructed on XQuAD \cite{artetxe-etal-2020-cross} and a question is linked with answer sentences in different languages. The training set of SQuAD v1.1 \cite{rajpurkar-etal-2016-squad} is used to fine-tune the model to adapt to QA retrieval task. During the evaluation, sentence embeddings are also obtained by a siamese network, and we retrieve the sentences with the highest cosine similarity as predictions.

\paragraph{Mewsli-X} Mewsli (\textbf{M}ultilingual Entities in N\textbf{ews}, \textbf{li}nked) requires linking an entity mention to its entry in a language-agnostic knowledge base \cite{botha-etal-2020-entity}. Mewsli-X \cite{ruder-etal-2021-xtreme} features 15k mentions in 11 languages. For each mention, Mewsli-X offers entity descriptions candidate pool containing 1M candidates across 50 languages. Fine-tuning is done on a predefined set of English-only mention-entity pairs from English Wikipedia hyperlinks. Our evaluation setting is identical to LAReQA.

\paragraph{BUCC} The second and third shared task of the workshop on Building and Using Parallel Corpora \cite{zweigenbaum-etal-2017-overview, ZWEIGENBAUM18.12} aims to examine the ability of models to detect parallel sentence pairs in a pair of monolingual corpora. The dataset provides train and test splits in 5 languages. Following XTREME \cite{pmlr-v119-hu20b}, we directly evaluate on BUCC without fine-tuning and retrieve sentences with the highest cosine similarity. 

\paragraph{Tatoeba} The goal of the Tatoeba dataset \cite{artetxe-schwenk-2019-massively} is to find the nearest neighbor for each sentence in the other language according to cosine similarity and compute the error rate. The dataset consists of up to 1,000 English-aligned sentence pairs covering 122 languages. Following XTREME \cite{pmlr-v119-hu20b}, we directly evaluate on Tatoeba without fine-tuning and retrieve sentences with the highest cosine similarity.

\paragraph{XCOPA} In the Cross-lingual Choice of Plausible Alternatives dataset \cite{ponti-etal-2020-xcopa}, each XCOPA instance corresponds to a premise and two alternatives. The task is formulated as a binary classification to predict the more plausible choice. The English COPA \cite{gordon-etal-2012-semeval} training set and Social IQa \cite{sap-etal-2019-social} training data are used for fine-tuning, while the validation and test sets of English COPA are translated and re-annotated into 11 languages for evaluation. 

\paragraph{MultiEURLEX} The MultiEURLEX dataset \cite{chalkidis-etal-2021-multieurlex} is a legal topic classification task that comprises 65k European Union (EU) laws in 23 official EU languages. The dataset provides multi-granular labels per document. The dataset is split into training, development, and test subsets chronologically, resulting in 55k training documents for 7 languages, and 5k each for development and test subsets in all 23 languages. 

\paragraph{MultiARC} The Multilingual Amazon Reviews Corpus \cite{keung-etal-2020-multilingual} is a large-scale collection of Amazon reviews for multilingual text classification in 6 languages. Different languages are directly gathered from the marketplaces in different countries. The goal is to predict the reviewer's rating on the 5-star scale using the test of the review as input. The data is clearly split into training (200,000 reviews), development (5,000 reviews), and test sets (5,000 reviews) for each language. 

\paragraph{PAWS-X} The Cross-lingual Paraphrase Adversaries from Word Scrambling \cite{yang-etal-2019-paws} dataset requires identifying whether two sentences are paraphrases. A subset of the evaluation pairs in English PAWS \cite{zhang-etal-2019-paws} are human-translated into 6 typologically distinct languages for evaluation, while the English PAWS training set is used for training. 

\section{Baseline Methods} \label{app:baseline_info}
To fairly evaluate the performance of \textsc{Emma}-X, we choose \textsc{Xlm-r} \cite{NEURIPS2019_c04c19c2} and its several derivatives as our baselines, which contain: (1) \textsc{Xlm-r}, which applies multilingual MLM tasks as pre-training objectives on CCNet-100 corpus; (2) \textsc{Hictl} \cite{wei2021on}, which continues training on \textsc{Xlm-r} using hierarchical contrastive learning; and (3) \textsc{InfoXLM}, which is initialized with \textsc{Xlm-r} and trains with cross-lingual contrast, multilingual \textsc{Mlm} and \textsc{Tlm}. Also, we compare \textsc{Emma}-X to strong sentence models: (1) S-BERT~\citep{reimers-gurevych-2020-making}, which adopts multilingual knowledge distillation to extend monolingual sentence representations to multilingual. We use the strongest baseline, \textbf{XLM-R $\leftarrow$ SBERT-paraphrase}, proposed in the original paper as a baseline. (2) LaBSE~\citep{feng-etal-2022-language}, which systematically combines several best methods, including masked language modeling, translation language modeling~\citep{NEURIPS2019_c04c19c2}, dual encoder translation ranking~\citep{guo-etal-2018-effective}, and additive margin softmax~\citep{ijcai2019p0746}, to learn cross-lingual sentence representations. It filters 17B monolingual sentences and 6B translation pairs for sentence representation learning. We take the best model, LaBSE with Customized Vocab as our baseline. We further report the zero-shot results on Large Language Model (\textsc{Llm}), ChatGPT, which is trained on a wide variety of multilingual sentences and instruction tuning based on Reinforcement Learning with Human Feedback~\citep{NIPS2017_d5e2c0ad,ouyang2022training}.

\section{Prompts for ChatGPT} \label{app:gpt-prompt}
In this section, we show the input prompts of ChatGPT on each task in Table~\ref{tbl:gpt_prompt}.

\begin{table}[t]
\begin{center}
\footnotesize
\resizebox{\textwidth}{!}{
\begin{tabular}{lcccccccccccccccc}
\toprule
\textbf{Model} & \textbf{en} & \textbf{ar} & \textbf{bg} & \textbf{de} & \textbf{el} & \textbf{es} & \textbf{fr} & \textbf{hi} & \textbf{ru} & \textbf{sw} & \textbf{th} & \textbf{tr} & \textbf{ur} & \textbf{vi} & \textbf{zh} & \cellcolor{lightgray}\textbf{Avg.} \\  
\midrule
\textsc{Xlm-r} & 88.6 &84.5 &86.7 &84.6 &85.2 &84.7 &82.0 &82.5&82.6 &82.4 &80.6 &83.1 &80.3 &77.3 &77.2 & \cellcolor{lightgray}\textbf{82.8}  \\
\textsc{InfoXLM} & 90.4 & 83.9 & 85.8 & 86.0 & 85.6 & 87.8 & 86.9 & 83.9 & 83.5 & 83.3 & 81.2 & 84.6 & 82.7 & 81.6 & 75.7 & \cellcolor{lightgray}\textbf{84.2} \\
\textsc{Hictl} & 90.6 & 86.8 & 88.2 & 87.4 & 87.0 & 87.4 & 85.0 & 83.9 & 83.3 & 84.8 & 83.1 & 85.7 & 82.8 & 79.7 & 80.9 & \cellcolor{lightgray}\textbf{85.1} \\
ChatGPT & 70.4 & 61.0 & 64.5 & 64.8 & 62.8 & 65.7 & 66.3 & 51.5 & 63.4 & 55.7 & 53.0 & 61.6 & 47.9 & 61.6 & 62.6 &
\cellcolor{lightgray}\textbf{60.9} \\
\midrule
\textbf{\textsc{Emma}-X} & \textbf{91.9} & \textbf{89.2} & \textbf{90.1} & \textbf{89.6} & \textbf{89.5} & \textbf{90.3} & \textbf{88.7} & \textbf{86.7} & \textbf{85.4} & \textbf{88.5} & \textbf{86.7} & \textbf{89.6} & \textbf{87.7} & \textbf{83.6} & \textbf{83.9} & \cellcolor{lightgray}\textbf{88.1}\\

\bottomrule
\end{tabular}}
\end{center}
\caption{XNLI results (accuracy) for each language.} \label{tbl:xnli-all-langs}
\end{table}

\begin{table}[t]
\begin{center}
\footnotesize
\resizebox{\textwidth}{!}{
\begin{tabular}{lccccccccccc}
\toprule
\textbf{Model} & \textbf{aym} & \textbf{bzd} & \textbf{cni} & \textbf{gn} & \textbf{hch} & \textbf{nah}& \textbf{oto} & \textbf{quy} & \textbf{shp} & \textbf{tar} & \cellcolor{lightgray}\textbf{Avg.} \\
\midrule
\textsc{Xlm-r} & 49.01 & 50.61 & 41.72 & 58.34 & 42.46 & 54.63& 35.57 & 59.29 & 51.62 & 41.54 & \cellcolor{lightgray}\textbf{48.48} \\
\textsc{InfoXLM} & 49.87 & 51.29 & 42.41 & 58.83 & 43.07 & 55.25& 36.14 & 59.87 & 52.20 & 42.12 & \cellcolor{lightgray}\textbf{49.10}  \\
\textsc{Hictl} & 49.65 & 51.22 & 42.36 & 58.82 & 43.09 & 55.13& 36.04 & 59.61 & 52.17 & 42.08 & \cellcolor{lightgray}\textbf{49.02}  \\
ChatGPT & 42.0 & 43.6 & 40.8 & 40.4 & 40.0 & 43.8 & 41.1 & 43.1 & 42.0 & 40.0 & \cellcolor{lightgray}\textbf{41.7} \\
\midrule
\textbf{\textsc{Emma}-X} & \textbf{51.19} & \textbf{52.50} & \textbf{43.62} & \textbf{59.88} & \textbf{44.31} & \textbf{55.44}& \textbf{39.16} & \textbf{60.14} & \textbf{52.84} & \textbf{43.10} & \cellcolor{lightgray}\textbf{50.21}\\
\bottomrule
\end{tabular}}
\end{center}
\caption{AmericasNLI (ANLI) results (top-1 accuracy) across different input languages.} \label{tbl:americasnli-all-langs}
\end{table}

\begin{table}[!t]
\begin{center}
\footnotesize
\resizebox{\textwidth}{!}{
\begin{tabular}{lccccccccccc}
\toprule
\textbf{Model} & \textbf{en-ar} & \textbf{en-de} & \textbf{en-tr} & \textbf{en-es} & \textbf{en-fr} & \textbf{en-it}& \textbf{en-nl} & \textbf{ar-ar} & \textbf{en-en} & \textbf{es-es} & \cellcolor{lightgray}\textbf{Avg.} \\ 
\midrule
\textsc{Xlm-r} & 50.2 & 63.7 & 45.8 & 59.6 & 68.0 & 63.4 & 69.6 & 87.7 & 82.5 & 68.5 & \cellcolor{lightgray}\textbf{65.9} \\
\textsc{InfoXLM} & 81.7 & 80.3 & 79.9 & 79.1 & 80.6 & 83.4 & 81.2 & 86.7 & 87.2 & 81.7 & \cellcolor{lightgray}\textbf{82.2} \\
\textsc{Hictl} & 80.4 & 81.8 & 78.3 & 80.6 & 81.2 & 80.9 & 79.3 & 88.4 & 86.1 & 79.6 & \cellcolor{lightgray}\textbf{81.6}\\
\midrule
\textbf{\textsc{Emma}-X} & \textbf{86.6} & \textbf{85.0} & \textbf{87.1} & \textbf{84.4} & \textbf{85.2} & \textbf{89.4} & \textbf{88.3} & \textbf{90.9} & \textbf{92.0} & \textbf{84.5} & \cellcolor{lightgray}\textbf{87.3}\\
\bottomrule
\end{tabular}}
\end{center}
\caption{MultiSTS results (Spearman) across different input languages.} \label{tbl:multists-all-langs}
\end{table}

\begin{table}[t]
    \centering
    \footnotesize
    \resizebox{\textwidth}{!}{
    \begin{tabularx}{\linewidth}{X}
    \toprule
    \textbf{Basic Prompt for XNLI/ANLI} \\
    \midrule
    \colorbox{lightgray}{\textbf{Task Description: }}Read the following and determine the relationship between Hypothesis and Premise. Choose relation from ``contradiction'', ``neutral'', or ``entailment''. \\
    \colorbox{lightgray}{\textbf{Hypothesis: }}Yo... no puedo pensar por qué deberías hablarme así, dijo ella, con menos de lo que le había asegurado antes.\\
    \colorbox{lightgray}{\textbf{Premise: }}Ella era una buena amiga de él, por esto le dolía que le hablara así. \\
    \midrule
    \textbf{Basic Prompt for MultiSTS} \\
    \midrule
    \colorbox{lightgray}{\textbf{Task Description: }}Read the following sentences and measure the real-valued meaning similarity between these two sentences. You can choose the meaning similarity score, ranging from 0 for no meaning overlap to 5 for meaning equivalence. \\
    \colorbox{lightgray}{\textbf{Sentence1: }}A person is on a baseball team. \\
    \colorbox{lightgray}{\textbf{Sentence2: }}Eine Person spielt in einem Team Basketball. \\
    \midrule    
    \textbf{Basic Prompt for QE} \\
    \midrule
    \colorbox{lightgray}{\textbf{Task Description: }}Read the Source sentence and its Translation, and estimate the quality of the Translation. You can rate the translation from 0-1 according to the perceived translation quality. \\
    \colorbox{lightgray}{\textbf{Source: }}În Franța a început stagnarea demografică de lungă durată, refacerea durând o generație. \\
    \colorbox{lightgray}{\textbf{Translation: }}In France, long-term demographic stagnation has started, restoring a generation. \\
    \midrule    
    \textbf{Basic Prompt for XCOPA} \\
    \midrule
    \colorbox{lightgray}{\textbf{Task Description: }}Read the Premise and determine which choice is the effect(or cause) of the Premise . Choose from ``Choice1'' or ``Choice2''. \\
    \colorbox{lightgray}{\textbf{Premise: }}Kuki kurukuna wasiman haykurqanku. \\
    \colorbox{lightgray}{\textbf{Choice1: }}Kuki kurukunaqa wasimanta chinkarqanku. \\
    \colorbox{lightgray}{\textbf{Choice2: }}Kuki kuruqa wasip kurkunta mikhurqanku. \\
    \midrule    
    \textbf{Basic Prompt for MultiEURLEX} \\
    \midrule
    \colorbox{lightgray}{\textbf{Task Description: }}Read the following sentences and determine the legal topic of the given sentence. The legal topic should choose from `international organisations', `social questions', `production', `technology and research', `environment', `energy', `transport', `law', `finance', `education and communications', `trade', `agriculture', `forestry and fisheries', `economics', `agri-foodstuffs', `EUROPEAN UNION', `science', `politics', `international relations', `industry', `geography', `business and competition', `employment and working conditions'. \\
    \colorbox{lightgray}{\textbf{Sentence: }}NEUVOSTON ASETUS (EU) N:o 1390/2013, annettu 16 päivänä joulukuuta 2013, Euroopan unionin ja Komorien liiton kesken näiden välisessä kalastuskumppanuussopimuksessa määrätyjen kalastusmahdollisuuksien ja taloudellisen korvauksen vahvistamisesta hyväksytyn pöytäkirjan mukaisten kalastusmahdollisuuksien jakamisesta ... \\
    \midrule    
    \textbf{Basic Prompt for MultiARC} \\
    \midrule
    \colorbox{lightgray}{\textbf{Task Description: }}Read the following review and predict a 5-star scale rating (1 means the poorest experience and 5 represents excellent or outstanding performance) that can best match the review. \\
    \colorbox{lightgray}{\textbf{Review: }}no me llego el articulo me lo mando por correos normal sin seguimiento y nunca me llego tota un desastre \\
    \midrule    
    \textbf{Basic Prompt for PAWS-X} \\
    \midrule
    \colorbox{lightgray}{\textbf{Task Description: }}Read the following sentences and determine whether two sentences are paraphrases. Return yes or no. \\
    \colorbox{lightgray}{\textbf{Sentence1: }}La excepción fue entre fines de 2005 y 2009 cuando jugó en Suecia con Carlstad United BK, Serbia con FK Borac Čačak y el FC Terek Grozny de Rusia. \\
    \colorbox{lightgray}{\textbf{Sentence2: }}La excepción se dio entre fines del 2005 y 2009, cuando jugó con Suecia en el Carlstad United BK, Serbia con el FK Borac Čačak y el FC Terek Grozny de Rusia. \\
    \midrule    
    \end{tabularx}}
    \caption{Prompts of ChatGPT on each task.}
    \label{tbl:gpt_prompt}
\end{table}

\begin{table}[t]
\begin{center}
\footnotesize
\resizebox{\textwidth}{!}{
\begin{tabular}{lcccccccccccc}
\toprule
\textbf{Model} & \textbf{en-de} & \textbf{en-zh} & \textbf{et-en} & \textbf{ne-en} & \textbf{ro-en} & \textbf{ru-en} & \textbf{si-en} & \textbf{en-cs} & \textbf{en-ja} & \textbf{km-en} & \textbf{ps-en} & \cellcolor{lightgray}\textbf{Avg.} \\ 
\midrule
\textsc{Xlm-r} & 0.412 & 0.566 & 0.797 & 0.812 & 0.891 & 0.774 & 0.578 & 0.547 & 0.335 & 0.612 & 0.635 & \cellcolor{lightgray}\textbf{0.632}  \\
\textsc{InfoXLM} & 0.517 & 0.534 & 0.775 & 0.834 & 0.890 & 0.788& 0.581 & 0.564 & 0.325 & 0.635 & 0.616 & \cellcolor{lightgray}\textbf{0.641}  \\
\textsc{Hictl} & 0.495 & 0.579 & 0.792 & 0.835 & \textbf{0.904} & 0.787& 0.575 & 0.556 & 0.342 & 0.625 & 0.648 & \cellcolor{lightgray}\textbf{0.649}  \\
\midrule
\textbf{\textsc{Emma}-X} & \textbf{0.580} & \textbf{0.589} & \textbf{0.809} & \textbf{0.854} & 0.897 & \textbf{0.829}& \textbf{0.593} & \textbf{0.577} & \textbf{0.370} & \textbf{0.641} & \textbf{0.651} & \cellcolor{lightgray}\textbf{0.672}\\

\bottomrule
\end{tabular}}
\end{center}
\caption{WMT21-QE-Task1 results (Pearson) across different input languages.} \label{tbl:wmtqe-all-langs}
\end{table}

\begin{table}[t]
\begin{center}
\footnotesize
\resizebox{\textwidth}{!}{
\begin{tabular}{lcccccccccccc}
\toprule
\textbf{Model} & \textbf{ar} & \textbf{de} & \textbf{el} & \textbf{en} & \textbf{es} & \textbf{hi}& \textbf{ru} & \textbf{th} & \textbf{tr} & \textbf{vi} & \textbf{zh} & \cellcolor{lightgray}\textbf{Avg.} \\ 
\midrule
\textsc{Xlm-r} & 34.1 & 42.4 & 39.3 & 44.8 & 44.0 & 37.3& 41.7 & 38.6 & 40.9 & 40.4 & 39.5 & \cellcolor{lightgray}\textbf{40.3}  \\
\textsc{InfoXLM} & 39.7 & 52.6 & 39.2 & 55.1 & 53.4 & 36.8 & 51.0 & 28.5 & 41.1 & 48.9 & 47.3 & \cellcolor{lightgray}\textbf{44.9}  \\
\textsc{Hictl} & 40.3 & 53.2 & 41.7 & 56.3 & 54.3 & 39.6 & 51.7 & 30.1 & 42.8 & 48.9 & 48.5 & \cellcolor{lightgray}\textbf{46.1}  \\
\midrule
\textbf{\textsc{Emma}-X} & \textbf{45.1} & \textbf{58.4} & \textbf{45.4} & \textbf{60.6} & \textbf{59.8} & \textbf{41.4} & \textbf{56.3} & \textbf{34.7} & \textbf{47.1} & \textbf{54.6} & \textbf{53.4} & \cellcolor{lightgray}\textbf{50.6} \\

\bottomrule
\end{tabular}}
\end{center}
\caption{LAReQA results (mean average precision@20, mAP@20) across different input languages.} \label{tbl:lareqa-all-langs}
\end{table}

\begin{table}[t]
\begin{center}
\footnotesize
\resizebox{\textwidth}{!}{
\begin{tabular}{lcccccccccccc}
\toprule
\textbf{Model} & \textbf{ar} & \textbf{de} & \textbf{en} & \textbf{es} & \textbf{fa} & \textbf{ja}& \textbf{pl} & \textbf{ro} & \textbf{ta} & \textbf{tr} & \textbf{uk} & \cellcolor{lightgray}\textbf{Avg.} \\ 
\midrule
\textsc{Xlm-r} & 34.6 & 66.0 & 62.6 & 64.8 & 27.1 & 47.8& 64.8 & 33.7 & 17.8 & 62.3 & 53.2 & \cellcolor{lightgray}\textbf{48.6}  \\
\textsc{InfoXLM} & 40.8 & 71.6 & 66.3 & 68.7 & 48.7 & 61.0 & 66.7 & 39.2 & 42.0 & 64.6 & 58.1 & \cellcolor{lightgray}\textbf{57.1}  \\
\textsc{Hictl} & 41.7 & 68.5 & 64.2 & 65.6 & 45.6 & 51.9 & 67.6 & 40.4 & 32.8 & 65.5 & 58.9 & \cellcolor{lightgray}\textbf{54.8}  \\
\midrule
\textbf{\textsc{Emma}-X} & \textbf{50.2} & \textbf{78.7} & \textbf{69.1} & \textbf{63.7} & 47.9& \textbf{59.6} & \textbf{70.0} & \textbf{50.2} & \textbf{43.5} & \textbf{68.0} & \textbf{60.9} & \cellcolor{lightgray}\textbf{59.6} \\

\bottomrule
\end{tabular}}
\end{center}
\caption{Mewsli-X results (mean average precision@20, mAP@20) across different input languages.} \label{tbl:mewsli-x-all-langs}
\end{table}

\begin{table}[t]
\begin{center}
\footnotesize
\resizebox{\textwidth}{!}{
\begin{tabular}{lcccccccccccc}
\toprule
\textbf{Model} & \textbf{et} & \textbf{ht} & \textbf{id} & \textbf{it} & \textbf{qu} & \textbf{sw}& \textbf{ta} & \textbf{th} & \textbf{tr} & \textbf{vi} & \textbf{zh} & \cellcolor{lightgray}\textbf{Avg.} \\ 
\midrule
\textsc{Xlm-r} & 73.8 & 67.4 & 77.8 & 72.2 & 52.3 & 70.9& 72.1 & \textbf{74.6} & 73.4 & 73.2 & 75.7 & \cellcolor{lightgray}\textbf{71.2}  \\
\textsc{InfoXLM} & 75.1 & 73.4 & \textbf{78.3} & 80.7 & 65.6 & 69.1& 72.7 & 73.9 & 76.9 & 77.8 & 77.5 & \cellcolor{lightgray}\textbf{74.6}  \\
\textsc{Hictl} & 75.9 & 73.1 & 77.8 & \textbf{81.2} & 65.5 & 73.8& 72.6 & 73.2 & 76.1 & 75.4 & 78.0 & \cellcolor{lightgray}\textbf{74.8}  \\
ChatGPT & 80.6 & 64.1 & 85.6 & 89.2 & 47.4 & 75.9 & 56.4 & 67.3 & 82.2 & 81.5 & 85.8 & \cellcolor{lightgray}\textbf{74.2} \\
\midrule
\textbf{\textsc{Emma}-X} & \textbf{76.8} & \textbf{74.0} & {77.6} & {79.8} & \textbf{76.2} & \textbf{74.4}& \textbf{77.8} & {74.2} & \textbf{77.6} & \textbf{82.6} & \textbf{89.6} & \cellcolor{lightgray}\textbf{78.2} \\
\bottomrule
\end{tabular}}
\end{center}
\caption{XCOPA results (accuracy) across different input languages.}
\label{tbl:xcopa-all-langs}
\end{table}

\begin{table}[!t]
\centering
\begin{minipage}[t]{0.4\textwidth}
\makeatletter\def\@captype{table}
\footnotesize
\resizebox{\textwidth}{!}{
\begin{tabular}{lccccc}
\toprule
\textbf{Model} & \textbf{de} & \textbf{fr} & \textbf{ru} & \textbf{zh} & \cellcolor{lightgray}\textbf{Avg.}\\ 
\midrule
\textsc{Xlm-r} & 76.1 & 72.3 & 62.3 & 60.8 & \cellcolor{lightgray}\textbf{67.9}\\
\textsc{InfoXLM} & 81.3 & 78.2 & 76.0 & 74.2 & \cellcolor{lightgray}\textbf{77.4}\\
\textsc{Hictl} & 80.5 & 79.2 & 76.0 & 74.8 & \cellcolor{lightgray}\textbf{77.6}\\
\midrule
\textbf{\textsc{Emma}-X} & \textbf{85.1} & \textbf{82.8} & \textbf{81.3} & \textbf{78.3} & \cellcolor{lightgray}\textbf{81.9} \\
\bottomrule
\end{tabular}}
\caption{BUCC results (F1) across different languages.} \label{tbl:bucc-all-langs}
\end{minipage}\quad
\begin{minipage}[t]{0.48\textwidth}
\makeatletter\def\@captype{table}
\footnotesize
\resizebox{\textwidth}{!}{
\begin{tabular}{lcccccccc}
\toprule
\textbf{Model} & \textbf{en} & \textbf{de} & \textbf{es} & \textbf{fr} & \textbf{ja} & \textbf{ko} & \textbf{zh} & \cellcolor{lightgray}\textbf{Avg.} \\ 
\midrule
\textsc{Xlm-r} & 95.7 & 92.2 & 92.7 & 92.5 & 84.7 & 85.9 & 87.1 & \cellcolor{lightgray}\textbf{90.1}   \\
\textsc{InfoXLM} & \textbf{97.7} & 94.6 & \textbf{95.2} & 95.1 & 88.9 & 89.0 & 90.2 & \cellcolor{lightgray}\textbf{93.0}   \\
\textsc{Hictl} & 97.4 & 94.2 & 95.0 & 94.2 & 89.1 & 89.5 & 90.2 & \cellcolor{lightgray}\textbf{92.8}   \\
ChatGPT & 71.9 & 67.8 & 67.9 & 67.0 & 58.3 & 54.7 & 61.4 & \cellcolor{lightgray}\textbf{64.2} \\
\midrule
\textbf{\textsc{Emma}-X} & 97.3 & \textbf{95.6} & 94.7 & \textbf{96.0} & \textbf{92.9} & \textbf{89.8} & \textbf{93.0} & \cellcolor{lightgray}\textbf{94.2}  \\
\bottomrule
\end{tabular}}
\caption{PAWS-X results (accuracy) for each language.} \label{tbl:pawsx-all-langs}
\end{minipage}
\vspace{-15pt}
\end{table}

\section{Results of each Language}
\label{appendix:results-per-language}
We show the details for tasks and all languages in Tables~\ref{tbl:xnli-all-langs} (XNLI), \ref{tbl:americasnli-all-langs} (AmericasNLI), \ref{tbl:multists-all-langs} (MultiSTS), \ref{tbl:wmtqe-all-langs} (QE), \ref{tbl:lareqa-all-langs} (LAReQA), \ref{tbl:mewsli-x-all-langs} (Mewsli-X), \ref{tbl:xcopa-all-langs} (XCOPA), \ref{tbl:bucc-all-langs} (BUCC) and \ref{tbl:pawsx-all-langs} (PAWS-X).

\section{Equations and Theoretical Analysis}
\label{app:detail_eq}
\subsection{Details of Equations}
\paragraph{Details of Gaussian Form $\mathcal{N}_r$} In \textsc{Emma}-X, GMM classifier is introduced to determine the semantic rank of sentence pairs. The posterior probability $P_{\mathcal{G}}(\cdot)$ of GMM classifier is already discussed in Eq. 5. We show the explicit calculation of  Gaussian form $\mathcal{N}_r(\gamma^{(\mathbf{x}_i)}, \gamma^{(\mathbf{y}_k)})$ as:
\begin{equation}
\footnotesize
\begin{split}
        \mathcal{N}_r(\gamma^{(\mathbf{x}_i)} - \gamma^{(\mathbf{y}_k)}|\mu_r, \sigma_r)&=\frac{\pi_r}{(2\pi)^{(d/2)}|\text{diag}(\sigma_r)|} 
        \cdot e^{\Big(-\frac{1}{2}\big[(\gamma^{(\mathbf{x}_i)} - \gamma^{(\mathbf{y}_k)}) - \mu_r\big]^T\text{diag}(\sigma_r^{-2})\big[(\gamma^{(\mathbf{x}_i)} - \gamma^{(\mathbf{y}_k)}) - \mu_r\big]\Big)},
\end{split}
\label{eq:gmm-explicit}
\end{equation}
where $d$ is the dimension of hidden states of $\gamma^{(\mathbf{x}_i)}$ and $\gamma^{(\mathbf{y}_k)}$.

\paragraph{Details of contrastive learning} The training objective of cross-lingual encoder in \textsc{Emma}-X is the ranking InfoNCE loss. We show the explicit expansion of this loss (Eq. 7) as:
\begin{equation}
\footnotesize
    \begin{split}
        &\mathcal{L}_{\mathbf{CTL}}(\mathcal{X},\mathcal{Y};\Theta_{\mathcal{M}}) =  - \mathbb{E}_{\mathbf{x}_{i} \sim \mathcal{X}} \Bigg [ \\
        &\underbrace{\log \frac{ \sum_{\mathbf{y}_{k} \sim \mathcal{Y}_{c_{\mathcal{G}}^*=1}} e^{\frac{s[\gamma^{(\mathbf{x}_{i})}, \gamma^{(\mathbf{y}_{k})}]}{\tau_1}}}{\sum_{\mathbf{y}_{t} \sim \mathcal{Y}_{c_{\mathcal{G}}^*=1}} e^{\frac{s[\gamma^{(\mathbf{x}_{i})}, \gamma^{(\mathbf{y}_{t})}]}{\tau_1}}+
        \sum_{\mathbf{y}_{t} \sim \mathcal{Y}_{c_{\mathcal{G}}^*=2}} e^{\frac{s[\gamma^{(\mathbf{x}_{i})}, \gamma^{(\mathbf{y}_{t})}]}{\tau_1}}+
        ...+
        \sum_{\mathbf{y}_{t} \sim \mathcal{Y}, \mathcal{Y}_{c_{\mathcal{G}}^*=4}} e^{\frac{s[\gamma^{(\mathbf{x}_{i})}, \gamma^{(\mathbf{y}_{t})}]}{\tau_1}}}}_{\ell_1} \\
        &+\underbrace{\log \frac{ \sum_{\mathbf{y}_{k} \sim \mathcal{Y}_{c_{\mathcal{G}}^*=2}} e^{\frac{s[\gamma^{(\mathbf{x}_{i})}, \gamma^{(\mathbf{y}_{k})}]}{\tau_2}}}{\sum_{\mathbf{y}_{t} \sim \mathcal{Y}_{c_{\mathcal{G}}^*=2}} e^{\frac{s[\gamma^{(\mathbf{x}_{i})}, \gamma^{(\mathbf{y}_{t})}]}{\tau_2}}+
        \sum_{\mathbf{y}_{t} \sim \mathcal{Y}_{c_{\mathcal{G}}^*=3}} e^{\frac{s[\gamma^{(\mathbf{x}_{i})}, \gamma^{(\mathbf{y}_{t})}]}{\tau_2}}+
        \sum_{\mathbf{y}_{t} \sim \mathcal{Y}_{c_{\mathcal{G}}^*=4}} e^{\frac{s[\gamma^{(\mathbf{x}_{i})}, \gamma^{(\mathbf{y}_{t})}]}{\tau_2}}}}_{\ell_2} \\
        &+\underbrace{\log \frac{ \sum_{\mathbf{y}_{k} \sim \mathcal{Y}_{c_{\mathcal{G}}^*=3}} e^{\frac{s[\gamma^{(\mathbf{x}_{i})}, \gamma^{(\mathbf{y}_{k})}]}{\tau_3}}}{\sum_{\mathbf{y}_{t} \sim \mathcal{Y}_{c_{\mathcal{G}}^*=3}} e^{\frac{s[\gamma^{(\mathbf{x}_{i})}, \gamma^{(\mathbf{y}_{t})}]}{\tau_3}}+
        \sum_{\mathbf{y}_{t} \sim \mathcal{Y}_{c_{\mathcal{G}}^*=4}} e^{\frac{s[\gamma^{(\mathbf{x}_{i})}, \gamma^{(\mathbf{y}_{t})}]}{\tau_3}}}}_{\ell_3} \Bigg ],
    \end{split}
    \label{eq:ctl-expansion}
\end{equation}
where $\tau_r$ represents the temperature term. As small temperature $\tau$ tends to be less tolerant to similar samples, and large $\tau$ tends to cluster similar samples together~\citep{Wang_2021_CVPR}, we empirically set $\tau_1 < \tau_2 < \tau_3 < \tau_4$, which remains the same as \citet{hoffmann2022ranking}.
\subsection{Theoretical Analysis}
In this section, we provide detailed proof for Eq. 14 and Eq. 15. Next, we prove the feasibility of our dual supervision. GMM classifier clusters sentence pairs in terms of Euclidean distance, while cross-lingual encoder minimizes the covariance of each semantic relation rank via cosine distance. Finally, we prove that these two metrics are actually equivalent to each other in the unit hypersphere of the embedding space.

\paragraph{Proof of Eq. 14.} We provide the derivation of Eq. 14. With the assumption that $P(\mathbf{x}_i, \mathbf{y}_k|c_{\mathcal{G}}^*=r,\Theta) \sim \mathcal{N}_r\big( 
\mathbf{x}_i - \mathbf{y}_k| \tilde{\mathbf{\mu}}_r, \tilde{\mathbf{\sigma}}_r\big)$, we have, 
\begin{equation}
    \footnotesize
    \begin{split}
         \sum_{\mathbf{x}_i \in \mathcal{X}} \sum_{\mathbf{y}_k \in \mathcal{Y}} \sum_{r=1}^{N} Q(r) \log  \frac{P(\mathbf{x}_i, \mathbf{y}_k,r|\Theta)}{Q(r)} 
        &\approx  \sum_{\mathbf{x}_i \in \mathcal{X}} \sum_{\mathbf{y}_k \in \mathcal{Y}} \sum_{r=1}^{N} \log P(\mathbf{x}_i, \mathbf{y}_k|c_{\mathcal{G}}^*=r,\Theta) \\
        &= \sum_{\mathbf{x}_i \in \mathcal{X}} \sum_{\mathbf{y}_k \in \mathcal{Y}} \sum_{r=1}^{N} \Big(\log \big(\frac{1}{(2\pi)^{(d/2)}|\tilde{\mathbf{\sigma}}_r|^{1/2}}\big) \\
        &+ \frac{1}{2}\big[(\mathbf{x}_i - \mathbf{y}_k) - \tilde{\mu}_r\big]^T\tilde{\mathbf{\sigma}}_r^{-1}\big[(\mathbf{x}_i - \mathbf{y}_k) - \tilde{\mu}_r\big] \Big) \\
        & \geq \sum_{r=1}^{N} \big[\sum_{\mathbf{x}_i \in \mathcal{X}} \sum_{\mathbf{y}_k \in \mathcal{Y}} (\mathbf{x}_i - \mathbf{y}_k)\big]^2 - 2 \Tilde{\mu}_r\sum_{\mathbf{x}_i \in \mathcal{X}} \sum_{\mathbf{y}_k \in \mathcal{Y}}(\mathbf{x}_i - \mathbf{y}_k) + n\tilde{\mu}_r^2 \\
        &= \sum_{r=1}^{N} n^2 \tilde{\mu}_r^2 - n\tilde{\mu}_r^2 \\
        &=  n (n - 1) \sum_{r=1}^{N}  \tilde{\mu}_r^2,
    \end{split}
    \label{eq:em_perspect_derv}
\end{equation}
with $n$ denoting the number of sentence pairs in semantic rank $r$. Here, we ignore the impact of $\tilde{\mathbf{\sigma}}_r$.

\paragraph{Proof of Eq. 15.} As we apply dual supervision, data in the contrastive label space also follows the distribution $\mathcal{N}_r\big( 
\mathbf{x}_i - \mathbf{y}_k| \tilde{\mathbf{\mu}}_r, \tilde{\mathbf{\sigma}}_r\big)$. Hence, under mild assumptions, we can get:
\begin{equation}
    \footnotesize
    \begin{split}
    \mathcal{L}^+_{\mathbf{CTL}}(\mathcal{X},\mathcal{Y};\Theta_{\mathcal{M}}) & = \mathbb{E}_{\mathbf{x}_{i} \sim \mathcal{X}} \sum_{r=1}^{N-1} \log \sum_{\mathbf{y}_{k} \sim \mathcal{Y}_{c_{\mathcal{G}}^*=r}} e^{s[\gamma^{(\mathbf{x}_{i})}, \gamma^{(\mathbf{y}_{k})}]} \\
    &=\sum_{\mathbf{x}_i \in \mathcal{X}} \sum_{\mathbf{y}_k \in \mathcal{Y}} \sum_{r=1}^{N-1} s(\mathbf{x}_i, \mathbf{y}_k) \\
    &=\sum_{\mathbf{x}_i \in \mathcal{X}} \sum_{\mathbf{y}_k \in \mathcal{Y}} \sum_{r=1}^{N-1} \frac{(\mathbf{x}_i - \mathbf{y}_k)^2 - 2}{2} \\
    &= n^2\sum_{r=1}^{N-1}\tilde{\mu}_r^2.
    \end{split}
    \label{eq:em_prove_derv}
\end{equation}

Based on the definition of semantic ranks, we have $\Tilde{\mu}_1<\Tilde{\mu}_2<...<\Tilde{\mu}_N$. Empirically, the number of sentence pairs in each rank $n$ is larger than the number of semantic ranks $N$. Hence, it can be derived that:
\begin{equation}
    \footnotesize
    \begin{split}
    \mathcal{L}^+_{\mathbf{CTL}}(\mathcal{X},\mathcal{Y};\Theta_{\mathcal{M}}) &= n^2\sum_{r=1}^{N-1}\tilde{\mu}_r^2 \\
    &< n^2\sum_{r=1}^{N - 1}\tilde{\mu}_r^2 + n^2\tilde{\mu}_N^2 - n \sum_{r=1}^{N} \tilde{\mu}_r^2 \\
    & = n (n - 1) \sum_{r=1}^{N}  \tilde{\mu}_r^2 \\
    & \leq  \sum_{\mathbf{x}_i \in \mathcal{X}} \sum_{\mathbf{y}_k \in \mathcal{Y}} \sum_{r=1}^{N} Q(r) \log  \frac{P(\mathbf{x}_i, \mathbf{y}_k,r|\Theta)}{Q(r)}. 
    \end{split}
    \label{eq:em_prove_final}
\end{equation}

Therefore, we prove that minimizing the positive terms $\mathcal{L}^+_{\mathbf{CTL}}(\mathcal{X},\mathcal{Y};\Theta_{\mathcal{M}})$ in contrastive learning is equivalent to maximizing a lower bound of the likelihood in Eq. 12.

According to the definition of semantic ranks, the approximated semantic rank $c_{\mathcal{G}}^*$ from GMM classifier should satisfy the following restriction, 
\begin{equation}
\footnotesize
    \mathbb{E}_{\mathbf{y}_k \sim \mathcal{Y}_{c_{\mathcal{G}}^*=1}}||\gamma^{(\mathbf{x}_i)} - \gamma^{(\mathbf{y}_k)}||<\mathbb{E}_{\mathbf{y}_k \sim \mathcal{Y}_{c_{\mathcal{G}}^*=2}}||\gamma^{(\mathbf{x}_i)} - \gamma^{(\mathbf{y}_k)}||<...<\mathbb{E}_{\mathbf{y}_k \sim \mathcal{Y}_{c_{\mathcal{G}}^*=N}}||\gamma^{(\mathbf{x}_i)} - \gamma^{(\mathbf{y}_k)}||.
    \label{eq:gmm-approximated-restriction}
\end{equation}

Similarly, the approximated semantic rank $c_{\mathcal{M}}^*$ from cross-lingual encoder should satisfy the following restriction, 

\begin{equation}
    \footnotesize
    \mathbb{E}_{\mathbf{y}_k \sim \mathcal{Y}_{c_{\mathcal{M}}^*=1}} s[\gamma^{(\mathbf{x}_i)},\gamma^{(\mathbf{y}_k)}] > \mathbb{E}_{\mathbf{y}_k \sim \mathcal{Y}_{c_{\mathcal{M}}^*=2}} s[\gamma^{(\mathbf{x}_i)},\gamma^{(\mathbf{y}_k)}] >...> \mathbb{E}_{\mathbf{y}_k \sim \mathcal{Y}_{c_{\mathcal{M}}^*=N}} s[\gamma^{(\mathbf{x}_i)},\gamma^{(\mathbf{y}_k)}].
    \label{eq:ctl-approximated-restriction}
\end{equation}

Next, we prove that these two restrictions are interchangeable with each other in a unit hypersphere. For simplicity, we consider only two ranks, but extending the explanation to more ranks is trivial. As the Euclidean distance is always larger than $0$, we have:
\begin{equation}
    \footnotesize
    \begin{split}
        \mathbb{E}_{\mathbf{y}_k \sim \mathcal{Y}_{c_{\mathcal{G}}^*=1}}||\gamma^{(\mathbf{x}_i)} - \gamma^{(\mathbf{y}_k)}||&<\mathbb{E}_{\mathbf{y} \sim \mathcal{Y}_{c_{\mathcal{G}}^*=2}}||\gamma^{(\mathbf{x}_i)} - \gamma^{(\mathbf{y}_k)}|| \\
        &\Leftrightarrow \mathbb{E}_{\mathbf{y}_k \sim \mathcal{Y}_{c_{\mathcal{G}}^*=1}}(\gamma^{(\mathbf{x}_i)} - \gamma^{(\mathbf{y}_k)})^2<\mathbb{E}_{\mathbf{y}_k \sim \mathcal{Y}_{c_{\mathcal{G}}^*=2}}(\gamma^{(\mathbf{x}_i)} - \gamma^{(\mathbf{y}_k)})^2\\
        &\Leftrightarrow \mathbb{E}_{\mathbf{y}_k \sim \mathcal{Y}_{c_{\mathcal{G}}^*=1}}(2 - 2\gamma^{(\mathbf{x}_i)}\gamma^{(\mathbf{y}_k)})<\mathbb{E}_{\mathbf{y}_k \sim \mathcal{Y}_{c_{\mathcal{G}}^*=2}}(2 - 2\gamma^{(\mathbf{x}_i)}\gamma^{(\mathbf{y}_k)}) \\
        &\Leftrightarrow \mathbb{E}_{\mathbf{y}_k \sim \mathcal{Y}_{c_{\mathcal{G}}^*=1}}s[\gamma^{(\mathbf{x}_i)},\gamma^{(\mathbf{y}_k)}]>\mathbb{E}_{\mathbf{y}_k \sim \mathcal{Y}_{c_{\mathcal{G}}^*=2}}s[\gamma^{(\mathbf{x}_i)},\gamma^{(\mathbf{y}_k)}] \\
        &\Leftrightarrow \mathbb{E}_{\mathbf{y}_k \sim \mathcal{Y}_{c_{\mathcal{M}}^*=1}}s[\gamma^{(\mathbf{x}_i)},\gamma^{(\mathbf{y}_k)}]>\mathbb{E}_{\mathbf{y}_k \sim \mathcal{Y}_{c_{\mathcal{M}}^*=2}}s[\gamma^{(\mathbf{x}_i)},\gamma^{(\mathbf{y}_k)}].
        \label{eq:gmm-to-ctl}
    \end{split}
\end{equation}

From the above analyses, we can tell that the approximated semantic rank from one module can provide a reasonable supervision signal to guide the training of the other module. Hence, all sentence pairs will be uniformly distributed according to a unified ranking semantic similarity in the embedding space.

\end{document}